%% file: arxiv.tex
\documentclass{article}

\usepackage{preprint_style}

\input{preamble}

\title{Don't Pick a Winner:\\Knowledge Conflicts Need a Prior, Not a Rule}

\author{%
  Minh Pham Van \qquad Cong Tran\thanks{Corresponding author.} \\
  Posts and Telecommunications Institute of Technology, Hanoi, Vietnam\\
  \texttt{MinhPV.B22KH078@stu.ptit.edu.vn}, \texttt{congtt@ptit.edu.vn}
}

\begin{document}
\maketitle

\input{sections/00_abstract}
\input{sections/01_introduction}
\input{sections/02_related}
\input{sections/03_theory}
\input{sections/04_grace}
\input{sections/05_diagnosis}

\input{sections/06_method}
\input{sections/07_results}
\input{sections/08_conclusion}

\bibliographystyle{plainnat}
\bibliography{references}

\clearpage
\appendix
\paragraph{Organisation of the appendix.} Appendices A--I follow. Five further sections (S1--S5: selector-internal ablations, where the method helps by task, full tables of the external tests with risk--coverage, the multiple-choice form of \grace{} with deployed GraphRAG-family systems, and selector v4 with earlier directions that failed) are in the supplementary material after Appendix I.

\paragraph{Names used in the appendix.} \emph{Qwen} and \emph{Llama} are Qwen3.5-35B-A3B and Llama-3.1-8B-Instruct. \emph{GRACE dev} is the 2,700 development records of \cref{sec:grace}. The \emph{plain reader} is the single-call reader of \cref{app:prompts}, also used as a baseline. \emph{Cautious mode} is the commit rule of \cref{prop:commit} (commit only above $1/2$); \emph{always answer} commits to the top candidate. \emph{Selector v3} is the selector of the paper, \emph{v2} the earlier version, \emph{v4} a later fix (\cref{supp:design}); ``ours'' without a version means v3. \emph{Protocol v3} is the scoring protocol of \cref{sec:grace} (\cref{app:grace}). \emph{Lockbox $n$} is the $n$-th preregistered external test (\cref{app:external}). Method families are named as in \cref{sec:diagnosis}: follow the context, decoding/trained control, implicit choice and memory first.

\input{appendix/A_proofs}

\input{appendix/B_grace}

\input{appendix/C_baselines}

\input{appendix/D_diagnosis}

\input{appendix/E_method}

\input{appendix/F_results}

\input{appendix/G_external}
\input{appendix/H_cost}

\input{appendix/I_lessons}

\clearpage
\begin{center}{\Large\bfseries Supplementary Material}\end{center}
\setcounter{section}{0}\setcounter{figure}{0}\setcounter{table}{0}
\renewcommand{\thesection}{S\arabic{section}}\renewcommand{\theHsection}{S\arabic{section}}
\renewcommand{\thefigure}{S\arabic{figure}}\renewcommand{\theHfigure}{S\arabic{figure}}
\renewcommand{\thetable}{S\arabic{table}}\renewcommand{\theHtable}{S\arabic{table}}

This part extends the appendix. Section, figure and table numbers with an S refer to this part.

\input{supplement/S1_selector}
\input{supplement/S2_where}
\input{supplement/S3_external}
\input{supplement/S4_mcq}
\input{supplement/S5_design}

\end{document}

%% file: preamble.tex
\usepackage[utf8]{inputenc}
\usepackage[T1]{fontenc}
\usepackage{hyperref}
\usepackage{url}
\usepackage{booktabs}
\usepackage{amsmath,amssymb,amsthm,amsfonts}
\usepackage{nicefrac}
\usepackage{microtype}
\usepackage{xcolor}
\usepackage{graphicx}
\usepackage{multirow}
\usepackage{makecell}
\usepackage{enumitem}
\usepackage{longtable,array,calc,capt-of}

\usepackage[capitalise,noabbrev]{cleveref}

\usepackage{newunicodechar}
\newunicodechar{−}{\ensuremath{-}}
\newunicodechar{π}{\ensuremath{\pi}}
\newunicodechar{α}{\ensuremath{\alpha}}
\newunicodechar{→}{\ensuremath{\rightarrow}}
\newunicodechar{Δ}{\ensuremath{\Delta}}
\newunicodechar{×}{\ensuremath{\times}}
\newunicodechar{≤}{\ensuremath{\le}}
\newunicodechar{≥}{\ensuremath{\ge}}
\newunicodechar{≈}{\ensuremath{\approx}}
\newunicodechar{č}{\v{c}}
\newunicodechar{ć}{\'{c}}
\graphicspath{{img/}}
\setlist{leftmargin=1.2em,itemsep=1pt,topsep=2pt}

\newtheorem{proposition}{Proposition}

\crefname{proposition}{Proposition}{Propositions}

\newcommand{\cw}{C$-$W}
\newcommand{\grace}{\textsc{GRACE}}
\newcommand{\logit}{\operatorname{logit}}
\newcommand{\1}{\mathbf{1}}
\newcommand{\E}{\mathbb{E}}
\newcommand{\ci}[2]{\,[#1,\,#2]}

%% file: sections/00_abstract.tex
\begin{abstract}
A retrieval-augmented language model answers from two sources, the retrieved context and its own knowledge, and the two disagree whenever the context is incomplete or wrong. Existing methods settle the disagreement with a rule fixed at design time: follow the context, amplify it in proportion to how far it moves the model, or trust the model's memory. Each rule is confidently wrong in some deployment. We show that this is not a tuning problem. In a decision-theoretic model of two sources, the score of any fixed rule is linear in the share of reliable contexts while the best attainable score is convex, so every fixed rule is beaten wherever it is not exactly optimal; and a symmetric divergence between the two sources, the signal adaptive decoding uses, carries no information about which one is right. Resolving a conflict therefore needs two inputs that no existing method keeps: evidence about each source separately, and a prior on how often the context is right. We test these claims on \grace{}, a benchmark that intervenes on the knowledge-graph context of a fixed question, so that what the model knows, what the context says and what is true are measured separately. On it, every baseline from four method families is strong on right contexts or on wrong ones, never on both, and AdaCAD's weight is the same whether the model is right or wrong while its losses grow with it. We then build the two inputs: the model states its knowledge as notes, a small policy voices each note with a calibrated confidence, a reader answers in two voices, and a selector commits or abstains under a prior $\pi$ applied as an exact prior shift. With $\pi$ set to the true share of right contexts, the system is above the best baseline at every mixture of \grace{} contexts, by up to 26 points and never significantly below. Its overall score is 45.3 and 35.0 (Qwen3.5-35B-A3B, Llama-3.1-8B), against 29.4 and 27.5 for the best baseline. On preregistered ClashEval it beats same-backbone CK-PLUG and TCR by 12 and 26 points. Where the model lacks the knowledge, the gains vanish, as the theory predicts.
\end{abstract}

%% file: sections/01_introduction.tex
\section{Introduction}
\label{sec:intro}

In May 2024 a deployed search engine advised its users to add glue to pizza sauce and to eat one rock a day \citep{reid2024aioverviews}. Both answers were faithful to retrieved pages, a joke in a forum thread and a satirical article. The language model that wrote them knew better, and nothing in the answers showed it. Retrieval-augmented generation \citep{lewis2020rag} is built on the opposite assumption, that the retrieved context is the more reliable source. It often is not: industry RAG systems answer only 63\% of CRAG questions without hallucination \citep{yang2024crag}, 57\% of web queries retrieve contradicting paragraphs \citep{yuan2026confrag}, and knowledge graphs miss and misstate facts at scale \citep{west2014kbcompletion,shenoy2022wikidataquality}. The model's own knowledge is no safer: it is reliable for popular entities and unreliable in the long tail \citep{mallen2023trust}.

\begin{figure}[t]
  \centering
  \includegraphics[width=0.96\linewidth]{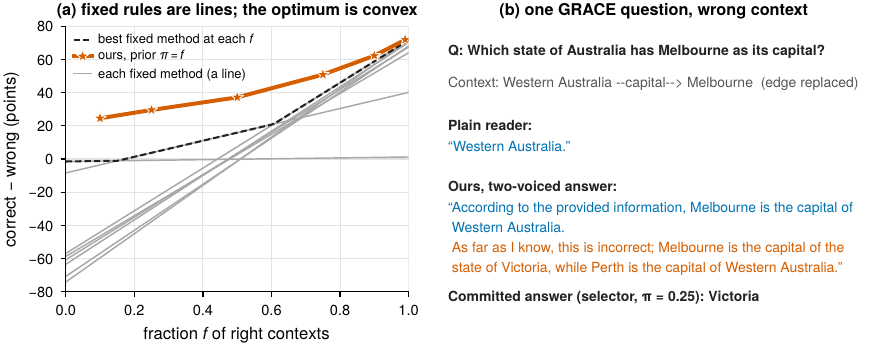}
  \caption{(a) \grace{} dev, Qwen: deployments with a share $f$ of right contexts (clean and wrong\_branch versions of the same questions, mixed). Every existing method is a line in $f$ (grey; dashed: the best at each $f$); ours with $\pi = f$ (orange) is above all of them. (b) A wrong context (real outputs): the plain reader repeats it; our answer states both sources and commits to the right one.}
  \label{fig:teaser}
\end{figure}

\paragraph{A rule chosen in advance.} The literature treats this situation as a \emph{knowledge conflict} to be resolved, and every method we know resolves it with a direction fixed before the question is seen. Context-following readers trust the context and use the model's knowledge only to fill silences or to locate conflicts \citep{zhang2025faithfulrag,liu2026truthfulrag,li2024selfroute}. Contrastive decoding amplifies the context against the model's prior \citep{shi2024cad,wang2025adacad}, and trained controllers learn a fixed preference \citep{bi2026ckplug,ye2026tcr}. Memory-first methods generate from the model's knowledge and lose what a right context says \citep{yu2023genread,wang2025astute}. Each choice is correct for some deployments and none for all: on a benchmark that varies only the context, each method is strong on right contexts or on wrong ones, not both (\cref{fig:teaser}a).

\paragraph{Why not adapt the weight to the disagreement?} The obvious repair is to let the data choose. AdaCAD \citep{wang2025adacad} sets its contrast weight to the Jensen--Shannon divergence between the model's next-token distributions with and without the context. The divergence rises when the context is wrong but does not tell \emph{which} source is wrong: on our benchmark it is the same whether the model is right or not, so a push toward the context is hardest exactly where the context is wrong. On the questions the model answers correctly without context, AdaCAD returns a wrong answer for 70\% of the wrong contexts, against 20\% for a plain reader.

\paragraph{What a decision needs.} These failures have one cause (\cref{sec:theory}). Whether to believe the context or the model is a Bayesian decision whose posterior odds are a prior, how often contexts are right in this deployment, plus evidence that distinguishes the two sources. A rule fixed at design time ignores the prior, so its score is a straight line in the share of right contexts, while the best achievable score is a convex curve above every such line (\cref{prop:envelope}). A symmetric divergence contributes no evidence about direction (\cref{prop:symmetric}). What remains is to supply the two missing quantities: evidence about the model's side, fact by fact, and a prior on the context that the user can set.

\paragraph{This paper.} We make the argument testable and then build what it asks for.
\begin{itemize}
  \item \textbf{A benchmark that intervenes on the context} (\cref{sec:grace}). \grace{} fixes a question and changes only its knowledge-graph context (right, missing, wrong, or both), so score changes are caused by the context alone.
  \item \textbf{A theory of two sources with predictions that can fail} (\cref{sec:theory,sec:diagnosis}): fixed rules are dominated, and a symmetric divergence cannot decide a conflict; both predictions hold on \grace{} for thirteen methods and two backbones.
  \item \textbf{A method that supplies the missing evidence and the prior} (\cref{sec:method}): calibrated knowledge notes, a two-voiced reader, and a selector under the context prior $\pi$, both controls being log-odds shifts with a stated meaning (\cref{prop:voice,prop:shift}).
  \item \textbf{Evidence, including where it fails} (\cref{sec:results}): above the best baseline at every mixture of \grace{} contexts, ahead of every same-backbone control method on a preregistered external test, and no gain where the model does not know, as predicted.
\end{itemize}

%% file: sections/02_related.tex
\section{Related Work}
\label{sec:related}

\paragraph{Knowledge conflicts.} \citet{longpre2021entity} formalised conflicts between a passage and a reader's memory by substituting answer entities. Large models followed coherent external evidence even against what they knew \citep{xie2024chameleon}, more so when their own prior was weak \citep{wu2024clasheval}; later benchmarks widened the evaluation \citep{su2024conflictbank,ming2025faitheval,xu2024conflictsurvey}. Analyses traced the behaviour to entity familiarity \citep{du2024context} and competing recall and copying mechanisms \citep{ortu2024competition}, and found both answers represented inside the model even when the output states one \citep{zhao2025reconciliation}. \citet{wang2024resolving} argued for one answer per source and found that models rarely give it. These benchmarks score one answer per question; none fixes the question and varies only the context, which is what separates what the model knows from what it reads.

\paragraph{Resolving conflicts with a fixed direction.} Context-following methods located conflicts with the model's own facts and then answered from the context \citep[FaithfulRAG;][]{zhang2025faithfulrag}, filtered conflicting triples inside the context \citep[TruthfulRAG;][]{liu2026truthfulrag}, or used memory only when the context was silent \citep[Self-Route;][]{li2024selfroute}. Contrastive decoding amplified the context against the model's prior with a fixed weight \citep[CAD;][]{shi2024cad} or a weight equal to the divergence of the two distributions \citep[AdaCAD;][]{wang2025adacad}; CK-PLUG \citep{bi2026ckplug} set the mixing weight by entropy, and TCR \citep{ye2026tcr} and Knowledgeable-R1 \citep{lin2026knowledgeabler1} trained the preference. Other methods let the model write rationales \citep[InstructRAG;][]{wei2025instructrag}, vote among samples \citep[USC;][]{chen2024usc}, or generate from memory \citep[GenRead, Astute RAG;][]{yu2023genread,wang2025astute}. All returned one answer, and each fixed either the direction or the procedure that chooses it.

\paragraph{Using both sources.} DisentQA \citep{neeman2023disentqa} trained a reader to output a contextual and a parametric answer side by side, without relating them, weighting them or choosing between them. Situated faithfulness \citep{huang2025situated} asked the model to trust the context according to its confidence in its own answer, self-reported or rule-based. Abstention and confidence estimation \citep{wen2025abstention,lin2022verbalized,tian2023justask,farquhar2024semanticentropy} judged when one answer can be trusted. We differ in attaching a learned confidence to each stated fact of the model, keeping both sources in the answer, and making the trust in the context an explicit prior (\cref{tab:positioning} in \cref{app:grace}).

\paragraph{Decision-theoretic foundations.} We build on the reject rule \citep{chow1970reject}, proper scoring rules \citep{brier1950,gneiting2007proper}, prior-shift correction \citep{saerens2002prior,lipton2018labelshift} and opinion pooling \citep{genest1986combining}, and contribute what they imply for knowledge conflicts.

%% file: sections/03_theory.tex
\section{Two Sources, One Decision}
\label{sec:theory}

\paragraph{Setting.} A question $x$ has a real-world answer $y^{*}$. The context yields an answer $a_c$ and the model's knowledge an answer $a_m$; either may be empty ($\varnothing$). Their reliabilities are $R_c = \1[a_c = y^{*}]$ and $R_m = \1[a_m = y^{*}]$. A system observes evidence $E$ (the context, the model's statements, any signal derived from them) and commits to an answer or abstains, scoring $s = +1$ if correct, $-1$ if wrong and $0$ if it abstains. Its expected score $\E[s]$ is the correct-minus-wrong rate (\cw) used throughout the paper. A deployment is characterised by $\pi = P(R_c = 1 \mid a_c \neq \varnothing)$, the probability that an answer the context gives is right.

\paragraph{When to commit.} Committing to $a$ has expected score $2q(a) - 1$ with $q(a) = P(a = y^{*} \mid E)$, and abstaining has $0$. Hence:

\begin{proposition}[commit rule]\label{prop:commit}
The rule that maximises $\E[s]$ commits to $\hat a = \arg\max_a q(a)$ if $q(\hat a) > 1/2$ and abstains otherwise.
\end{proposition}

This is Chow's reject rule \citep{chow1970reject} for these costs. It fixes what a system must estimate: a calibrated probability for every candidate answer, not only a ranking.

\paragraph{Fixed rules are lines.} Suppose deployments differ only in $\pi$: given whether the context is right, the distribution of evidence and answers is the same everywhere \citep[label shift;][]{lipton2018labelshift}. A \emph{fixed policy} $\phi$ maps $E$ to $a_c$, $a_m$ or an abstention without using $\pi$; every method of \cref{sec:related} is one. Conditioning on $R_c$ and averaging, its expected score on questions where the context gives an answer is
\begin{equation}\label{eq:line}
S_\phi(\pi) \;=\; \pi\, S_\phi(1) + (1-\pi)\, S_\phi(0),
\end{equation}
a straight line between its score on right contexts and its score on wrong ones.

\begin{proposition}[fixed rules are dominated]\label{prop:envelope}
Let $S^{*}(\pi) = \max_\phi S_\phi(\pi)$ be the best score achievable by a decision maker who knows $\pi$. Then (i) $S_\phi$ is affine in $\pi$, as in \cref{eq:line}; (ii) $S^{*}$ is convex and $S^{*}(\pi) \ge S_\phi(\pi)$ for every fixed $\phi$ and every $\pi$; (iii) if $\phi$ is optimal at $\pi_1$ but not at $\pi_2$, then $S^{*}(\pi_2) > S_\phi(\pi_2)$.
\end{proposition}

A method that is good on right contexts and bad on wrong ones is thus not a weak implementation of a good idea: it is one line, optimal at one end at best. Reaching $S^{*}$ takes what one fixed answer discards: both candidates with calibrated probabilities, and $\pi$.

\paragraph{A divergence detects a conflict but cannot resolve it.} Consider conflicts, $K = \{a_c \neq a_m,\ a_c \neq \varnothing,\ a_m \neq \varnothing\}$, where at most one source is right, and let $\rho_c = P(R_c = 1 \mid K)$ and $\rho_m = P(R_m = 1 \mid K)$. By Bayes' rule the posterior odds that the context and not the model is right are
\begin{equation}\label{eq:odds}
\log\frac{P(R_c = 1 \mid E, K)}{P(R_m = 1 \mid E, K)}
\;=\;
\underbrace{\log\frac{\rho_c}{\rho_m}}_{\text{prior}}
\;+\;
\underbrace{\log\frac{P(E \mid R_c = 1, K)}{P(E \mid R_m = 1, K)}}_{\text{evidence}} .
\end{equation}
The prior depends on the deployment, which no single input reveals. The evidence term in \cref{eq:odds} is zero for any statistic that is equally distributed under the two hypotheses. A symmetric divergence such as the Jensen--Shannon divergence, $\mathrm{JSD}(p,q) = \mathrm{JSD}(q,p)$ \citep{lin1991divergence}, is the natural candidate: the case ``context wrong, model right'' and its mirror ``context right, model wrong'' give it the same value (Lemma~A.1 in \cref{app:proofs}).

\begin{proposition}[a symmetric statistic carries no direction]\label{prop:symmetric}
Let $T$ be any statistic of the two next-token distributions, and let a rule output $a_c$, $a_m$ or an abstention with probabilities $g_c(T)$, $g_m(T)$ and $1 - g_c(T) - g_m(T)$. If $T$ is independent of $(R_c, R_m)$ given $K$, then
(i) the expected score on conflicts is $(2\rho_c - 1)\,\E[g_c(T)] + (2\rho_m - 1)\,\E[g_m(T)]$, so no rule that uses $T$ beats the best rule that ignores it;
(ii) that best rule is constant;
(iii) where $\rho_c < 1/2$ and $\rho_c \le \rho_m$, a rule that moves probability toward $a_c$ as $T$ grows scores no higher, and strictly lower wherever such probability moves, as $T$ grows.
\end{proposition}

Contrastive decoding with a weight that grows with the divergence is such a rule: by (iii), on wrong contexts it loses most where the divergence is largest. The divergence shows that a conflict has occurred, not how to resolve it \citep[cf.][]{genest1986combining}.

\paragraph{What the theory asks for.} A system on the envelope needs (a) a calibrated probability per candidate (\cref{prop:commit}), (b) asymmetric evidence about one source (\cref{prop:symmetric}), and (c) the deployment prior (\cref{prop:envelope}); \cref{sec:method} builds each. The propositions are short consequences of Bayes' rule whose value is that the data can contradict them (\cref{sec:diagnosis,sec:results}); proofs are in \cref{app:proofs}.

%% file: sections/04_grace.tex
\section{GRACE: Intervening on the Context}
\label{sec:grace}

To test these predictions the model's knowledge must stay fixed while the context changes, and the score must be measured against the real-world answer, not against the context. \grace{} does this for knowledge-graph question answering.

\paragraph{Construction.} Questions and SPARQL queries come from LC-QuAD 2.0 \citep{dubey2019lcquad2} and subgraphs from Wikidata \citep{vrandecic2014wikidata}. The \emph{gold edges} of a question are the edges on its path to the answer. Each question is given in four context states: \emph{clean} (the original subgraph, $R_c = 1$); \emph{broken} (one gold edge removed, so the context gives no answer, $a_c = \varnothing$); \emph{wrong\_branch} (the gold edges replaced by copies whose answer entity is wrong, $R_c = 0$); and \emph{competing} (the gold edges kept and the wrong copies added).

The question, the real-world answer and the model's closed-book answer are the same in all four states, so $R_m$ is fixed per question and every difference between states is caused by the context. Mixing the clean and wrong\_branch versions of the same questions in proportions $f$ and $1-f$ builds a deployment with $\pi = f$ in which label shift holds exactly.

\paragraph{Scoring.} Systems answer in open form. Each answer is correct, wrong or an abstention, and the main score is $\text{\cw} = 100\,(n_{\text{correct}} - n_{\text{wrong}})/n$. Answers longer than four words or containing a list are passed whole through one shared extractor that never sees the gold answer, and then matched strictly against the gold name and its Wikidata aliases; the same procedure applies to every method (\cref{app:grace}). Intervals are 95\% bootstrap intervals with all states of a question resampled together.

\paragraph{Data.} \grace{} has a train split of 15,683 questions and a held-out test split of 4,028, both with strictly validated subgraphs. All experiments use the train split: a training pool of 1,000 questions, three disjoint development (dev) slices of 225 questions (2,700 records), and \grace{}-QT, 1,000 questions of four unseen question types.

%% file: sections/05_diagnosis.tex
\section{Fixed Rules on GRACE: The Predictions Hold}
\label{sec:diagnosis}

We run thirteen methods from four families (follow the context, decoding/trained control, implicit choice, memory first) on two backbones, Qwen3.5-35B-A3B and Llama-3.1-8B-Instruct \citep[Qwen and Llama below;][]{meta2024llama31}, using official code wherever it exists (\cref{app:baselines}). CK-PLUG, TCR, CAD and AdaCAD need the model's logits or a released training recipe for Llama, so they are compared on Llama only.

\begin{figure}[t]
  \centering
  \includegraphics[width=0.64\linewidth]{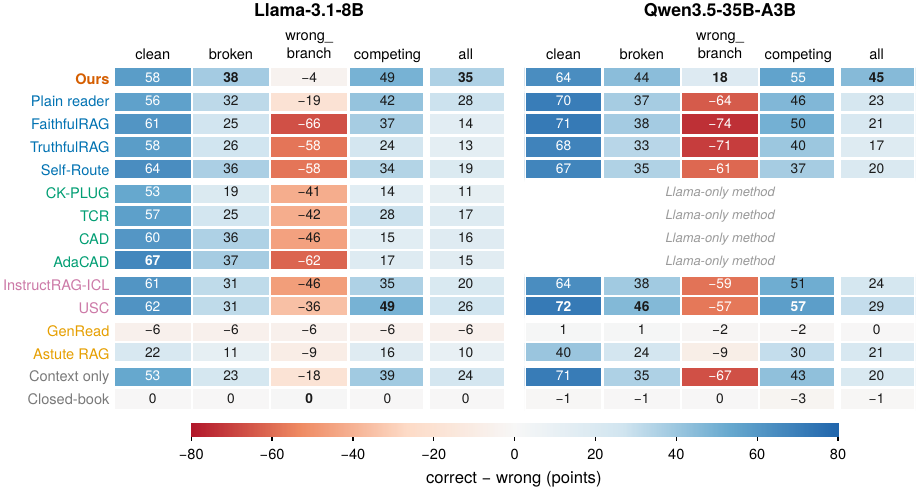}
  \caption{\cw{} on \grace{} development data by context state, for every method and both backbones. Every method that reads the context scores between $-36$ and $-74$ on wrong\_branch contexts; methods that ignore it lose the clean column. Ours (\cref{sec:method}) is the only row without a strongly negative cell. CK-PLUG, TCR, CAD and AdaCAD are run on Llama only.}
  \label{fig:states}
\end{figure}

\paragraph{Every fixed rule is strong at one end.} \Cref{fig:states} shows each method's score by context state. Methods that follow the context or choose implicitly reach 64 to 72 \cw{} on clean contexts with Qwen and $-57$ to $-74$ when the answer edge is wrong. Memory-first methods avoid that loss (Astute RAG $-8$, GenRead $-2$) and give up the clean case (40 and 1). Decoding/trained control methods fall below the plain reader they modify (Llama: CK-PLUG 11.4, TCR 17.0, CAD 16.3, AdaCAD 15.0, plain reader 27.5). No method is strong on both clean and wrong\_branch, as \cref{prop:envelope} predicts for lines. Neither source can simply be trusted: the closed-book model is right as often as it is wrong (Qwen: 43.4\% against 44.6\%).

\begin{figure}[t]
  \centering
  \includegraphics[width=\linewidth]{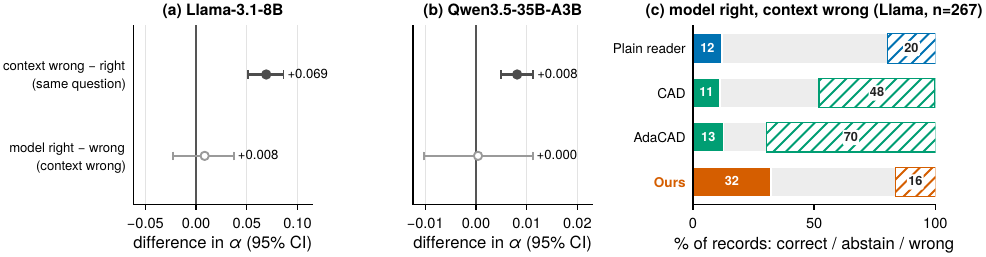}
  \caption{The divergence detects a wrong context but not which source is right. (a, b) AdaCAD's weight $\alpha$ at the first answer token: higher when the context of the same question is wrong (filled), no higher when the model's own answer is right (open). (c) Where the model is right and the context wrong, amplifying the context turns abstentions into wrong answers. Per-state values in \cref{app:divergence}.}
  \label{fig:divmain}
\end{figure}

\paragraph{AdaCAD's weight does not depend on who is right (\cref{fig:divmain}).} \grace{} asks the same question with a right and a wrong context and knows the model's closed-book answer, so it can test the premise of \cref{prop:symmetric} directly. We read AdaCAD's weight $\alpha$, the Jensen--Shannon divergence at the first generated token, from its Llama run. Because AdaCAD's prompt ends with ``Answer:'', that token is the first token of the answer whenever the output begins with a name; we report those records (52--59\% per state) and give the full set in \cref{app:diagnosis}.
With Llama, $\alpha$ is higher on the wrong\_branch than on the clean version of a question by 0.069\ci{0.051}{0.086} (68\% of 300 questions), but within wrong\_branch it is 0.375 when the model answers correctly without context and 0.367 when it does not (difference 0.008\ci{-0.022}{0.037}): the premise of \cref{prop:symmetric}. With Qwen, under AdaCAD's prompts,\footnote{The server returns the top 100 tokens of each distribution (97\% of the mass on average); the rest is one bucket, so the divergence is a lower bound.} the pattern repeats ($+0.008$\ci{0.005}{0.012}; knows against wrong 0.169 and 0.169, difference 0.000\ci{-0.010}{0.011}). As \cref{prop:symmetric}(iii) predicts, AdaCAD's wrong\_branch score falls across thirds of $\alpha$ ($-38.2$, $-67.1$, $-79.1$; fall 40.9\ci{29.6}{51.6}), while the plain reader shows no comparable trend ($-6.7$\ci{-16.3}{3.1}).

The mechanism shows on the 267 wrong\_branch records whose question the model answers correctly on its own (\cref{tab:knows}): the plain reader is wrong on 19.9\%, CAD on 47.9\% and AdaCAD on 69.7\%, while all three are right on only 11--13\%. Amplification turns abstentions into wrong answers and recovers none of the model's knowledge; ours is right on 32.2\% and wrong on 16.5\%. The same amplification gives AdaCAD the highest clean score of any method (66.8 against 55.7 for the plain reader): it is the best method on \grace{} when the context is right and among the worst when it is wrong, because it amplifies the context in proportion to a quantity that does not depend on whether the context is right (\cref{fig:divmain}; per-state values in \cref{fig:divergence}, \cref{app:divergence}).

%% file: sections/06_method.tex
\section{Supplying the Evidence and the Prior}
\label{sec:method}

\begin{figure}[t]
  \centering
  \includegraphics[width=\linewidth]{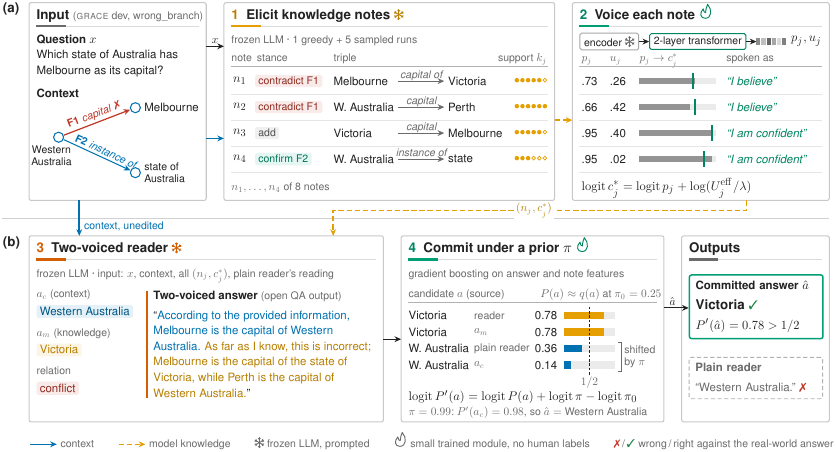}
  \caption{Method overview on one real \grace{} dev record with a wrong context (Qwen). (a) Six elicitation runs yield notes $n_j$ with support $k_j$; a frozen encoder and the learned two-layer policy give $p_j, u_j$ and the spoken confidence $c_j^{*}$ (\cref{eq:voice}). (b) The reader receives the question, the unedited context, all voiced notes and the plain reader's reading, and answers in two voices that keep $a_c$ and $a_m$. The selector scores four candidates, shifts the context-sourced ones from $\pi_0$ to the stated $\pi$ (\cref{eq:shift}), and commits if the best exceeds $1/2$.}
  \label{fig:overview}
\end{figure}

\Cref{sec:theory} asks for asymmetric evidence about the model's side, a calibrated probability per candidate, and the deployment prior. We build each with a frozen language model and two small learned components, keeping the context unedited (\cref{fig:overview}).

\paragraph{Evidence about the model's side: knowledge notes.} The evidence must speak about the model and be separable from the context, because a model that reads a context first tends to repeat it: on questions it answers correctly on its own, a plain reader allowed to use its knowledge is right on only 8.2\% (Qwen) and 12.0\% (Llama) of wrong contexts. We therefore elicit the model's knowledge before it answers, as \emph{notes}: triples tied to individual context facts with a stance (confirm, contradict, add, bridge). The elicitation runs once greedily and five times with sampling, and the \emph{support} $k_j \in \{1,\dots,6\}$ of a note counts the runs that produce it: what a model knows it states every time, what it guesses it states sometimes. With the notes, the same questions are answered correctly on 67.0\% (Qwen) and 32.2\% (Llama) of wrong contexts.

\paragraph{How confidently to say it: a calibrated note policy.} A note is evidence only to the extent that its reliability is known. A two-layer transformer over the question, the context facts and the notes, with embedding \citep{chen2024bgem3}, self-verification and support features, predicts for each note $p_j = P(\text{true})$ and $u_j = P(\text{true and useful})$, where useful means that it restores a missing edge, corrects a wrong fact or bridges to the answer. Labels come from the \grace{} construction and Wikidata, with no human annotation. Both heads are fitted with cross-entropy; the spoken confidence then maximises an asymmetric Brier reward \citep{brier1950}: voicing a note with confidence $c$ earns $U(1-(1-c)^2)$ if it is true and useful, $U_p(1-(1-c)^2)$ if true but peripheral, and costs $\lambda c^2$ if false ($U = 2$, $U_p = 0.25$, $\lambda = 1$). Maximising the expected reward gives the spoken confidence in closed form.

\begin{proposition}[spoken confidence]\label{prop:voice}
With $U_j^{\mathrm{eff}} = \eta_j U + (1-\eta_j) U_p$ and $\eta_j = u_j / p_j$, the expected reward has a unique maximiser $c_j^{*}$, and
\begin{equation}\label{eq:voice}
\logit c_j^{*} \;=\; \logit p_j \;+\; \log\frac{U_j^{\mathrm{eff}}}{\lambda}.
\end{equation}
Hence (i) $c_j^{*}$ is strictly increasing in $p_j$; (ii) $c_j^{*} = p_j$ exactly when $U_j^{\mathrm{eff}} = \lambda$; (iii) otherwise $c_j^{*}$ departs from $p_j$ by a constant shift in log-odds.
\end{proposition}

Case (ii) is the proper Brier score \citep{gneiting2007proper}; with our utilities a useful note with $p_j = 0.5$ is voiced at $2/3$, a peripheral one at $0.2$. The reader receives $c_j^{*}$ as a verbal hedge (\cref{fig:overview}, step 2).

\paragraph{Keeping both sources: a two-voiced reader.} One call to the frozen model receives the question, the unedited context and every note with its hedge, and returns what the context says, what the model knows, how they relate (agree, complement, conflict, context silent) and an answer in two voices (\cref{fig:overview}, step 3).

\paragraph{Committing under a prior: the selector.} When a task requires one answer, a gradient-boosting classifier \citep{friedman2001gbm} scores each candidate (the reader's conclusion, what the context says, what the model knows, and the plain reader's answer) with $P(a)$, its estimate of $q(a)$ in \cref{prop:commit}, from features of its source, its agreement with each source and the notes that support it. It is trained once, on data where one context in four is clean ($\pi_0 = 0.25$). At use, the user states $\pi$, and every context-sourced candidate is shifted:
\begin{equation}\label{eq:shift}
\logit P'(a) \;=\; \logit P(a) \;+\; \logit \pi \;-\; \logit \pi_0 .
\end{equation}
The selector then applies \cref{prop:commit} to the shifted probabilities: it commits to the most probable candidate if its probability exceeds $1/2$ and abstains otherwise (the \emph{cautious} rule).

\begin{proposition}[exact prior shift]\label{prop:shift}
If $P(a)$ is calibrated at $\pi_0$ and the deployment differs from training only in $\pi$ (label shift), then $P'(a)$ in \cref{eq:shift} is the exact posterior at $\pi$ \citep{saerens2002prior}, and the selector is the Bayes rule for that deployment.
\end{proposition}

The selector thus moves along the envelope $S^{*}$ as far as its calibration allows; without label shift \cref{eq:shift} is not exact but stays monotone in $\pi$, so the direction of the control survives. Both learned controls are thus the same operation, an additive log-odds shift whose size has a stated meaning: a ratio of utilities for the notes (\cref{eq:voice}), a ratio of priors for the context (\cref{eq:shift}).

%% file: sections/07_results.tex
\section{Results}
\label{sec:results}

\subsection{The system is on the envelope}
\label{sec:envelope}

\paragraph{Overall.} On \grace{} development data at the trained prior $\pi = 0.25$, the system scores 45.3 \cw{} with Qwen and 35.0 with Llama. Every difference in \cref{tab:main} excludes zero; the gain comes from the states where the context is not simply right (wrong\_branch: 17.9 and $-4.1$ against $-63.9$ and $-19.1$ for the plain reader).

\begin{table}[t]
  \centering\footnotesize
  \caption{\grace{} development data, \cw{}, $\pi = 0.25$. $\Delta$: ours minus method, 95\% paired interval. --: Llama only (official code or logit access; \cref{app:baselines}).}
  \label{tab:main}
  \begin{tabular}{@{}lrlrl@{}}
    \toprule
    & \multicolumn{2}{c}{Qwen3.5-35B-A3B} & \multicolumn{2}{c}{Llama-3.1-8B} \\
    \cmidrule(lr){2-3}\cmidrule(l){4-5}
    Method & \cw & $\Delta$ & \cw & $\Delta$ \\
    \midrule
    \textbf{Ours} & \textbf{45.3} & & \textbf{35.0} & \\
    USC \citep{chen2024usc} & 29.4 & $+15.9$\ci{12.6}{19.1} & 26.5 & $+8.5$\ci{4.8}{12.3} \\
    plain reader & 22.6 & $+22.7$\ci{19.8}{25.7} & 27.5 & $+7.5$\ci{4.4}{10.6} \\
    InstructRAG-ICL \citep{wei2025instructrag} & 23.7 & $+21.6$\ci{18.4}{24.8} & 20.1 & $+14.9$\ci{11.6}{18.2} \\
    Astute RAG \citep{wang2025astute} & 21.7 & $+24.0$\ci{19.8}{28.4} & 10.1 & $+24.9$\ci{20.4}{29.5} \\
    FaithfulRAG \citep{zhang2025faithfulrag} & 21.2 & $+24.1$\ci{20.7}{27.4} & 14.2 & $+20.8$\ci{17.1}{24.5} \\
    Self-Route \citep{li2024selfroute} & 19.6 & $+25.7$\ci{22.4}{29.0} & 19.1 & $+15.9$\ci{12.4}{19.4} \\
    TruthfulRAG \citep{liu2026truthfulrag} & 17.4 & $+27.9$\ci{24.3}{31.4} & 12.7 & $+22.3$\ci{18.6}{26.1} \\
    TCR \citep{ye2026tcr} & -- & & 17.0 & $+18.0$\ci{14.2}{21.9} \\
    CAD \citep{shi2024cad} & -- & & 16.3 & $+18.7$\ci{15.1}{22.4} \\
    AdaCAD \citep{wang2025adacad} & -- & & 15.0 & $+20.0$\ci{16.7}{23.2} \\
    CK-PLUG \citep{bi2026ckplug} & -- & & 11.4 & $+23.6$\ci{20.0}{27.4} \\
    \bottomrule
  \end{tabular}
\end{table}

\begin{figure}[t]
  \centering
  \includegraphics[width=\linewidth]{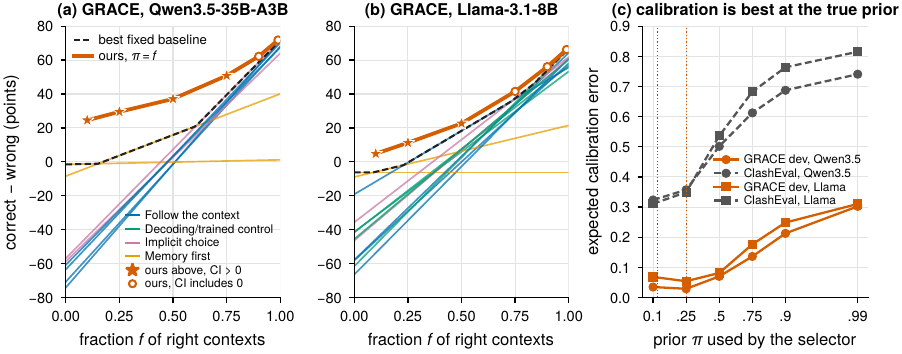}
  \caption{Testing the theory. (a, b) Deployments with a share $f$ of right contexts (\grace{} dev): every baseline is a line; dashed, the best at each $f$; orange, ours with $\pi = f$ (stars: significantly above). (c) Calibration error against the prior used: lowest at the true share on \grace{} (0.25) and ClashEval (0.114), but high on ClashEval, where label shift fails.}
  \label{fig:theory}
\end{figure}

\paragraph{The envelope (\cref{prop:envelope}).} A single score at one mixture hides the claim of the paper, which concerns every mixture. We build deployments with a share $f$ of right contexts from the clean and wrong\_branch versions of the same questions, run the system with $\pi = f$, and compare it with the best baseline at each $f$, chosen after seeing the scores, which favours the baselines (\cref{fig:theory}a,b; all values in \cref{tab:envelope}, \cref{app:pitable}).

As \cref{prop:envelope} predicts, the best fixed method changes with $f$ (memory first when contexts are mostly wrong, follow the context or decoding control when mostly right), and one system with one parameter is above all of them: by 25.7, 25.8, 21.3 (Qwen) and 10.7, 11.7, 4.4 (Llama) points at $f = 0.1, 0.25, 0.5$, significantly for $f \le 0.75$ (Qwen) and $f \le 0.5$ (Llama), and never significantly below. The margin vanishes as $f \to 1$, where following the context is already optimal.

\paragraph{Calibration is best at the true prior (\cref{prop:shift}).} On \grace{} (one context in four right) the selector's expected calibration error is 0.029\ci{0.021}{0.049} (Qwen) and 0.054\ci{0.043}{0.076} (Llama) at $\pi = 0.25$ and rises monotonically as $\pi$ moves away (\cref{fig:theory}c). \textbf{Committing only above $1/2$ never hurts (\cref{prop:commit}):} the cautious rule beats always answering at every $\pi$, on both backbones and both data sets, by 3.0 to 5.9 points on \grace{} and 1.3 to 34.6 on ClashEval.

\paragraph{Ablations.} Removing notes costs 21.7 (Qwen) and 14.4 (Llama) points, uniform confident voicing 7.1 and 9.3, sampling 3.3 to 3.7, the selector 8.9 to 9.7 (\cref{fig:ablation}).

\subsection{A preregistered external test}
\label{sec:external}

\grace{} is our development data, so its numbers are an upper bound on what to expect elsewhere. The frozen system and every baseline were tested once on ClashEval \citep[10,942 records, 89\% with a perturbed document;][]{wu2024clasheval} and Housing Statute QA \citep[1,000 yes/no questions;][]{zheng2025legalretrieval}, with hypotheses written to a hashed file before the run (\cref{app:external}).

\paragraph{ClashEval.} With Llama, where every control method can run, the system scores $-29.1$ at the registered $\pi = 0.25$ and beats CK-PLUG by 12.2 points\ci{9.3}{14.9}, TCR by 25.6\ci{22.9}{28.4}, the plain reader by 11.1, and CAD and AdaCAD (run after the preregistration) by 15.7 and 21.3. With Qwen it beats ten of eleven baselines by 13 to 33 points and ties Astute RAG ($-0.3$\ci{-3.0}{2.5}), which distrusts every document and falls behind once a quarter are unmodified.

\paragraph{Where it falls short, and why.} With Llama, GenRead scores higher ($-9.5$ against $-29.1$) by ignoring the document and abstaining on 62\% of questions, which pays when 89\% of documents are perturbed; on unmodified ones it is 53.9 points below ours. The selector, trained on graph facts, is miscalibrated on these passages (it states 0.49 on average for answers right 18\% of the time): the direction of the prior transfers (\cref{fig:theory}c), its calibration does not (\cref{app:lockbox2}).

\paragraph{When the model does not know.} On Housing Statute QA the model is wrong on 60\% of questions, so there is nothing to merge: with Qwen the system scores 2.0 against 11.4 for the plain reader, rising with $\pi$ toward it (8.8 at $\pi = 0.99$; Llama in Section~\ref{app:clasheval}). On \grace{}-QT (poor knowledge, right contexts) it ties the best baseline. It helps when the model knows and the context is unreliable.

\paragraph{Cost.}\label{sec:cost} Measured one request at a time on an idle GPU, the system takes 5.2\,s (Qwen) and 6.3\,s (Llama) per question, within the range of multi-call baselines (2.8 to 20.4\,s; \cref{app:cost}). Without the five sampled runs it takes 3.7 and 5.3\,s, still scoring 41.6 and 31.7, above every baseline.

%% file: sections/08_conclusion.tex
\section{Limitations and Conclusion}
\label{sec:conclusion}

\paragraph{Limitations.} The limit is the selector, a small learned classifier: even with good candidates and well-calibrated confidences, its calibration follows the question and context types it is trained on, so a new type may need it retrained (\cref{app:lockbox2}).

\paragraph{Conclusion.} A fixed rule is a line under a convex optimum, and a divergence shows conflict, not direction. Calibrated evidence about the model plus a prior on the context puts one system above every fixed method on \grace{} and ahead of same-backbone controls on a preregistered test; where the model does not know, there is nothing to merge.

%% file: appendix/A_proofs.tex
\section{Proofs}
\label{app:proofs}

Notation follows \cref{sec:theory}. All expectations are over questions of a deployment; $E$ is the evidence available to the system.

\textbf{\cref{prop:commit}.} Given $E$, committing to $a$ scores $+1$ with probability \(q(a)\) and $-1$ otherwise, so its expected score is \(q(a) - (1 - q(a)) = 2q(a) - 1\); abstaining scores $0$. The best action maximises \(\max\{\max_a (2q(a) - 1),\ 0\}\). The inner maximum is attained at \(\hat a = \arg\max_a q(a)\), and \(2q(\hat a) - 1 > 0\) if and only if \(q(\hat a) > 1/2\). Ties are indifferent. $\square$

\textbf{\cref{prop:envelope}.} Questions whose context gives no answer contribute a term that does not depend on $\pi$; write it $C_\phi$. For the others, let $D_1$ and $D_0$ be the distributions of \((E, y^{*})\) given $R_c = 1$ and $R_c = 0$. By the label-shift assumption these are the same in every deployment, and the deployment distribution is the mixture \(\pi D_1 + (1 - \pi) D_0\). A fixed policy $\phi$ has a fixed score function \(s_\phi(E, y^{*})\), so

\[
S_\phi(\pi) = C_\phi + \pi\,\mathbb{E}_{D_1}[s_\phi] + (1-\pi)\,\mathbb{E}_{D_0}[s_\phi],
\]

which is affine in $\pi$: (i). For each $\pi$, the Bayes rule of \cref{prop:commit} applied to the posterior under $\pi$ is a policy $\phi_\pi$ with \(S_{\phi_\pi}(\pi) = \max_\phi S_\phi(\pi) = S^{*}(\pi)\). A pointwise maximum of affine functions is convex, and it is at least each of them: (ii). If $\phi$ is not optimal at $\pi_2$, then \(S_\phi(\pi_2) < \max_{\phi'} S_{\phi'}(\pi_2) = S^{*}(\pi_2)\): (iii). $\square$

\phantomsection\label{lem:mirror}\textbf{Lemma A.1 (contrastive decoding follows the context\textquotesingle s relative peak).} Let $w$ maximise both $\ell_{\text{ctx}}$ and $\ell_{\text{ctx}} - \ell_{\text{no-ctx}}$ over the vocabulary. Then for every $\alpha \ge 0$, $w$ maximises \((1 + \alpha)\,\ell_{\text{ctx}} - \alpha\,\ell_{\text{no-ctx}} = \ell_{\text{ctx}} + \alpha\,(\ell_{\text{ctx}} - \ell_{\text{no-ctx}})\), a sum of two functions that are both maximised at $w$ with a non-negative weight on the second. Adding a constant to either logit vector (normalisation) shifts the sum by a constant and does not change the maximiser. $\square$

Together with the symmetry \(\mathrm{JSD}(p, q) = \mathrm{JSD}(q, p)\), which holds because \(M = (p+q)/2\) is symmetric and the two KL terms are exchanged, Lemma A.1 gives the mirror argument of \cref{sec:theory}: two inputs that differ only by exchanging the two distributions receive the same $\alpha$, and under the condition of the lemma in each, both are decoded to the context\textquotesingle s peak; exactly one of the two peaks is $y^{*}$.

\textbf{\cref{prop:symmetric}.} Under $K$ the two answers differ, so $R_c + R_m \le 1$. Given $T$, the rule\textquotesingle s expected score is

\[
\mathbb{E}[s \mid T, K] = g_c(T)\,\bigl(2\,P(R_c = 1 \mid T, K) - 1\bigr) + g_m(T)\,\bigl(2\,P(R_m = 1 \mid T, K) - 1\bigr).
\]

Independence of $T$ and \((R_c, R_m)\) given $K$ gives \(P(R_c = 1 \mid T, K) = \rho_c\) and \(P(R_m = 1 \mid T, K) = \rho_m\). Taking the expectation over $T$ gives (i). The score in (i) is linear in \((\mathbb{E}[g_c], \mathbb{E}[g_m])\), which ranges over the simplex $\{g_c, g_m \ge 0,\ g_c + g_m \le 1\}$; a linear function on a simplex is maximised at a vertex, \((1, 0)\), \((0, 1)\) or \((0, 0)\), with values $2\rho_c - 1$, $2\rho_m - 1$ and $0$: (ii). For (iii), fix $T = t$. Moving probability $\delta$ from abstention to $a_c$ changes the score by \(\delta\,(2\rho_c - 1)\), negative when $\rho_c < 1/2$; moving $\delta$ from $a_m$ to $a_c$ changes it by \(2\delta\,(\rho_c - \rho_m)\), non-positive when $\rho_c \le \rho_m$. A rule that moves probability in these directions as $t$ grows therefore has a score that does not increase in $t$, and decreases strictly wherever probability moves from abstention, or from $a_m$ when $\rho_c < \rho_m$. $\square$

\textbf{\cref{prop:voice}.} Write $U$ for $U_j^{\mathrm{eff}}$ and $p$ for $p_j$, with $U, \lambda > 0$ and \(p \in (0, 1)\). The expected reward \(G(c) = pU\,(1 - (1-c)^2) - (1-p)\,\lambda c^2\) has \(G'(c) = 2pU(1 - c) - 2(1-p)\lambda c\) and \(G''(c) = -2pU - 2(1-p)\lambda < 0\), so it is strictly concave and \(G'(c) = 0\) has the unique solution \(c^{*} = pU / (pU + (1-p)\lambda) \in (0, 1)\). Then \(c^{*}/(1 - c^{*}) = pU / ((1-p)\lambda)\), and taking logarithms gives \cref{eq:voice}. (i) and (iii) follow because the logit is strictly increasing and the shift \(\log(U/\lambda)\) does not depend on $p$; AUC depends only on the ranking. For (ii), with $U = \lambda$, \(G(c) = \lambda\,[\,p - p(1-c)^2 - (1-p)c^2\,]\), which is $\lambda p$ minus $\lambda$ times the expected Brier loss of forecast $c$; it is maximised at $c = p$, as the Brier score is strictly proper. $\square$

\textbf{\cref{prop:shift}.} By Bayes\textquotesingle{} rule, the posterior odds at prior $\pi$ are

\[
\frac{P_\pi(Z = 1 \mid X)}{P_\pi(Z = 0 \mid X)} = \frac{\pi}{1 - \pi}\cdot\frac{P(X \mid Z = 1)}{P(X \mid Z = 0)}.
\]

The likelihood ratio does not depend on the prior by assumption. Writing the same identity at $\pi_0$ and dividing the two eliminates it: \(\operatorname{logit} P_\pi(Z = 1 \mid X) = \operatorname{logit} P_{\pi_0}(Z = 1 \mid X) + \operatorname{logit}\pi - \operatorname{logit}\pi_0\), which is \cref{eq:shift}. The right-hand side is strictly increasing in $\pi$ for any value of $P_{\pi_0}$, with or without the assumption. $\square$

%% file: appendix/B_grace.tex
\section{GRACE: construction, scoring and checks}
\label{app:grace}

\paragraph{Construction and splits.}
\label{app:construction}

\begin{itemize}
\item
  Questions and SPARQL come from LC-QuAD 2.0 \citealp{dubey2019lcquad2}; subgraphs come from Wikidata \citealp{vrandecic2014wikidata}.
\item
  GRACE is released in two splits built by the same pipeline: a \textbf{test split of 4,028 questions} (from the 6,046 LC-QuAD 2.0 test questions) and a \textbf{train split of 15,683 questions} (from LC-QuAD 2.0 train). A question is kept only if its subgraph passes strict validation (zero structural or ground-truth errors; \texttt{dataset\_statistics.md} in each split).
\item
  Every experiment in the main text uses the train split only: the pool, the three dev slices and GRACE-QT (\cref{sec:grace}) are drawn from it, and no question of the test split was used to train, select or evaluate any component. The test split is used only for the diagnosis of deployed GraphRAG systems in Supplementary Section~\ref{app:mcq} and is otherwise released as a held-out set.
\item
  We release both splits, the exact four-state records of the pool and dev slices with their seeds, GRACE-QT, and the scoring code of protocol v3.
\item
  For each question the \emph{gold edges} are the edges on the path to the answer. The four states used in this paper are built from the clean subgraph (\texttt{overnight\_data.make\_states}):
\end{itemize}

{\footnotesize\setlength{\tabcolsep}{3pt}
\begin{longtable}[]{@{}>{\raggedright\arraybackslash}p{0.208\linewidth}>{\raggedright\arraybackslash}p{0.360\linewidth}>{\raggedright\arraybackslash}p{0.391\linewidth}@{}}
\toprule\noalign{}
& Definition & Measured by \\
\midrule\noalign{}
\endhead
\bottomrule\noalign{}
\endlastfoot
\textbf{Acc1} (knows) & correct without any context & closed-book arm \\
\textbf{Acc2} (reads) & answers what the context says, right or wrong & context-only arm, scored against the context\textquotesingle s answer \\
\textbf{Acc3} (is right) & answers the real-world answer & every method; the final objective \\
\end{longtable}}

Acc1 and Acc2 are diagnostic. Acc3 is what a user needs. All tables in this paper report Acc3-based scores.

\textbf{Scoring protocol.}

\begin{itemize}
\item
  \textbf{Open-form.} Systems generate an answer; no options are shown.
\item
  \textbf{Three-way.} Each answer is correct, an abstention, or wrong. An abstention is not penalised as an error.
\item
  \textbf{Main score:} \(\mathrm{C{-}W} = 100\,(n_{\text{correct}} - n_{\text{wrong}})/n\), percent correct minus percent wrong. A system that always answers and is right half the time scores 0.
\item
  \textbf{Protocol v3} (used for every number unless stated). Any answer that contains a list separator or has more than four words is passed, whole, through one shared extractor (an LLM prompt that sees only the question and the answer text, never the gold answer, options or context). The extracted short answer is matched strictly: equality after normalisation, or the gold contained in an answer at most three words longer, or the answer contained in a gold name. Gold names include Wikidata aliases. The same procedure is applied to every method, including ours.
\item
  \textbf{Statistics.} Paired bootstrap over questions (all states of a question are resampled together), 95\% intervals.
\end{itemize}

{\footnotesize\setlength{\tabcolsep}{3pt}
\begin{longtable}[]{@{}>{\raggedright\arraybackslash}p{0.122\linewidth}>{\raggedright\arraybackslash}p{0.563\linewidth}>{\raggedright\arraybackslash}p{0.275\linewidth}@{}}
\toprule\noalign{}
Split & Size & Use \\
\midrule\noalign{}
\endhead
\bottomrule\noalign{}
\endlastfoot
Pool & 1,000 LC-QuAD-train questions × 4 states & training the note policy and the selector \\
Dev & 3 disjoint slices of 225 questions × 4 states = 2,700 records & development and all GRACE tables \\
GRACE-QT & 1,000 LC-QuAD-train questions of types absent from the pool (rank, qualified statement, boolean, string); 500 train / 500 dev; real Wikidata 1-hop context, no perturbation & question-type generalisation \\
\end{longtable}}

Pool, dev and GRACE-QT are disjoint. External test sets are described in \cref{app:external}.

\subsection{Design properties of prior work}
\label{app:positioning}

\cref{tab:positioning} compares the lines of work above on the six properties that our contribution covers. Except for DisentQA, every prior method returns one answer, so the reader of that answer cannot tell which source it came from; DisentQA returns two answers without relating them, weighting them or choosing between them. Methods that use model knowledge explicitly (GenRead, Astute RAG, SCR and RCR) do not attach it to individual context facts, and their confidence is self-reported or rule-based, not learned. Our method is the only row that relates the two sources fact by fact, states a learned confidence for the model\textquotesingle s side, and exposes the trust in the context as a prior with a probabilistic meaning; it does so without training the reader. The table describes design properties, not accuracy: \cref{sec:external} shows that Astute RAG ties our method on one external test.

\begin{table}[ht]
\centering\caption{Design properties of prior work and of our method.}\label{tab:positioning}
\footnotesize\setlength{\tabcolsep}{3pt}
\begin{tabular}{@{}>{\raggedright\arraybackslash}p{0.145\linewidth}>{\raggedright\arraybackslash}p{0.13\linewidth}>{\raggedright\arraybackslash}p{0.14\linewidth}>{\raggedright\arraybackslash}p{0.135\linewidth}>{\raggedright\arraybackslash}p{0.12\linewidth}>{\raggedright\arraybackslash}p{0.085\linewidth}>{\raggedright\arraybackslash}p{0.135\linewidth}@{}}
\toprule
Method & Uses model knowledge explicitly & Knowledge tied to individual context facts & Calibrated confidence & Both sources visible in the answer & Abstains & Interpretable control \\
\midrule
FaithfulRAG, TruthfulRAG & partly (to find conflicts) & yes & no & no & no & no \\
Self-Route & only when context is silent & no & no & no & no & no \\
CK-PLUG, CAD, AdaCAD & implicitly (logits) & no & no & no & no & α (fixed, entropy- or divergence-based) \\
TCR, Knowledgeable-R1 & implicitly (trained) & no & no & no & no & no \\
InstructRAG, USC & implicitly & no & no & no & no & no \\
GenRead, Astute RAG & yes & no & self-reported & no & no & no \\
SCR, RCR (situated faithfulness) & yes (an answer and its confidence) & no & self-reported (SCR) or rule-based (RCR) & no & no & no \\
DisentQA & yes & no & no & yes (two separate answers) & no & no \\
\textbf{Ours} & yes & yes (confirm / contradict / add / bridge) & yes (learned, per note) & yes & yes & π (prior over context reliability) \\
\bottomrule
\end{tabular}
\end{table}

\subsection{Does the model use knowledge or benchmark cues?}
\label{app:cues}

\textbf{What the raw notes contain} (one dev slice, 900 records, greedy elicitation, before the policy). In the broken state, a note states the removed edge in 50.7\% of records, and in 37.8\% of the records whose question the model answers wrongly with no context. In wrong\_branch, a note marks the injected fact as wrong in 78.4\% of records and gives the correct value in 37.3\%; in competing, 79.7\% and 23.9\%. The model therefore often knows a piece of the answer path that a closed-book answer does not reveal, and it also states false notes, which is why confidence is calibrated instead of notes being filtered.

\textbf{Corrections concentrate where the model knows the answer.} On one dev slice (225 questions), the model answers 62.7\% of questions correctly with no context. Among records where a note states the removed edge (broken, 117 records), the share is 76.1\%; where a note contradicts the injected fact with the right value, it is 75.9\% (wrong\_branch, 79 records) and 76.5\% (competing, 51 records). A surface cue for injected facts would not depend on whether the model knows the answer.

\textbf{Reasoning traces.} The pipeline runs with thinking disabled. As a diagnostic, the elicitation prompt was run with thinking enabled on 120 records (30 per state). The traces refer to recalled knowledge (on average 7.7 phrases such as "known", "actually", "historically", "I recall" per trace). Example, wrong\_branch: \emph{"Fela Kuti\textquotesingle s children are Femi Kuti, Seun Kuti, etc. There is no prominent \textquotesingle George Washington\textquotesingle{} among them. This looks like \ldots{} erroneous data in the retrieved facts."} Words that could indicate benchmark awareness ("trick", "placeholder", "distractor") appear in 28\% of traces; read individually, they comment on awkward question wording, on how to fill a JSON field, or on irrelevant facts. Thinking costs about 4,500 generated tokens per elicitation against about 300 without, and 28 of 120 runs exceeded the 6,000-token limit; it does not change the score (\cref{sec:envelope}).

\textbf{The "George Washington" artefact.} The GRACE generator uses one default distractor entity: "George Washington" is the injected object in 101 of 675 wrong\_branch records (and in the 101 matching competing records). The note policy and the selector are trained on GRACE-derived data and could have learned this cue. C−W under protocol v3:

\begin{center}\footnotesize\setlength{\tabcolsep}{3.5pt}
\begin{tabular}{@{}lrrrr@{}}
\toprule
 & \multicolumn{2}{c}{Qwen} & \multicolumn{2}{c}{Llama} \\
\cmidrule(l{3pt}r{3pt}){2-3}\cmidrule(l{3pt}r{3pt}){4-5}
\makecell[r]{wrong\_\\branch} & artefact (101) & other (574) & artefact (101) & other (574) \\
\midrule
ours, π = 0.25 & 17.8 & 17.9 & 7.9 & −6.3 \\
plain reader & −73.3 & −61.7 & −15.8 & −19.7 \\
USC & −65.3 & −55.4 & −14.9 & −39.2 \\
GenRead (does not read the context) & −9.9 & 0.0 & −5.9 & −6.3 \\
Astute RAG & −4.0 & −9.2 & −7.9 & −9.1 \\
TCR & n/a & n/a & −40.6 & −41.6 \\
CK-PLUG & n/a & n/a & −53.5 & −39.2 \\
\bottomrule
\end{tabular}\end{center}

With Qwen our system scores the same on both groups. With Llama it scores 14.2 points higher on the artefact records; a baseline with no training on GRACE (USC) differs by 24.3 points in the same direction, so with 101 records the difference cannot be attributed to a learned cue. Removing all 202 artefact records changes the overall score from 45.3 to 46.1 (Qwen) and from 35.0 to 35.4 (Llama), and the margin over the plain reader from 22.7 to 20.3 and from 7.5 to 7.0. The external tests contain no GRACE artefact.

%% file: appendix/C_baselines.tex
\section{Baselines and prompts}
\label{app:baselines}

\begin{itemize}
\item
  \textbf{Qwen3.5-35B-A3B} (AWQ 4-bit), thinking disabled.
\item
  \textbf{Llama-3.1-8B-Instruct} (Meta, 2024).
\end{itemize}

The same backbone is used for every stage of our system and for every baseline in a comparison. TCR and CK-PLUG run only on Llama-3.1-8B in their official code, and CAD and AdaCAD need the output logits, so the direct comparison with these four is on Llama. Qwen results are reported against Qwen baselines only.

{\footnotesize\setlength{\tabcolsep}{3pt}
\begin{longtable}[]{@{}>{\raggedright\arraybackslash}p{0.324\linewidth}>{\raggedright\arraybackslash}p{0.197\linewidth}>{\raggedright\arraybackslash}p{0.438\linewidth}@{}}
\toprule\noalign{}
Baseline & Venue & Source used \\
\midrule\noalign{}
\endhead
\bottomrule\noalign{}
\endlastfoot
FaithfulRAG & ACL 2025 & official code \\
TruthfulRAG & AAAI 2026 & official code \\
InstructRAG-ICL & ICLR 2025 & official code \\
GenRead & ICLR 2023 & official code \\
TCR & AAAI 2026 & official code, retrained with the authors\textquotesingle{} recipe (three fixes were needed to run it) \\
CK-PLUG & ICLR 2026 & official code, adaptive α (one line changed: a hard-coded GPU index) \\
CAD & NAACL 2024 & official arithmetic in one process, α = 1 (the authors\textquotesingle{} knowledge-conflict setting) \\
AdaCAD & NAACL 2025 & official arithmetic in one process, α = Jensen--Shannon divergence per step \\
Astute RAG & ACL 2025 & prompts from the paper\textquotesingle s appendix \\
Self-Route & EMNLP 2024 & prompts from the paper \\
USC & 2024 & prompts from the paper \\
ClashEval token-probability correction & NeurIPS 2024 D\&B & authors\textquotesingle{} prompts \\
plain reader; context-only reader; closed-book & & ours \\
\end{longtable}}

Not run: Knowledgeable-R1 (needs eight GPUs and has no released checkpoint), SHIFT (preprint, needs a patched inference server).

{\footnotesize\setlength{\tabcolsep}{3pt}
\begin{longtable}[]{@{}>{\raggedright\arraybackslash}p{0.135\linewidth}>{\raggedright\arraybackslash}p{0.150\linewidth}>{\raggedright\arraybackslash}p{0.355\linewidth}>{\raggedright\arraybackslash}p{0.304\linewidth}@{}}
\toprule\noalign{}
Baseline & Code & What is unchanged & What had to be adapted \\
\midrule\noalign{}
\endhead
\bottomrule\noalign{}
\endlastfoot
FaithfulRAG & official, commit \texttt{9181b11} & the authors\textquotesingle{} pipeline (\texttt{get\_self\_facts}, \texttt{get\_topk\_chunks}, \texttt{get\_predictions}, normal CoT), no line changed & the OpenAI-compatible backend points at our server \\
InstructRAG-ICL & official, commit \texttt{8147bbc} & the authors\textquotesingle{} prompt builder and template; rationales generated on 200 GRACE-pool records; two random in-context examples as in the paper & chat tokens of the template changed to the backbone\textquotesingle s \\
GenRead & official, commit \texttt{49828ba} & the original prompts and \texttt{add\_prompt}; step 1 at most 300 tokens, step 2 at most 10, greedy & called through the chat template instead of the completion API. GenRead does not read the retrieved context \\
TruthfulRAG & official & the steps of \texttt{TruthfulRAG.run()}; MiniLM on CPU as in the README & backend points at our server \\
CK-PLUG & official & \texttt{eval\_rag.py}, mode \texttt{ck}, adaptive α, the "Q: \ldots{} A:" prompt & one line: a hard-coded GPU index; the first output line is taken as the answer \\
CAD, AdaCAD & official arithmetic, commits \texttt{1281d7f} (CAD) and \texttt{57508c3} (AdaCAD) & \texttt{get\_jsd} and \texttt{filter\_logits\_top\_p} taken verbatim from the AdaCAD file; per-step combination as in the code (CAD: weights 2 and −1, the authors\textquotesingle{} NQ-Swap setting; AdaCAD: α = JSD of the two distributions); greedy, 32 new tokens, the AdaCAD question-answering prompts, first output line as the answer (as in \texttt{eval\_qa.py}) & the official code holds one model copy per input in two processes on two GPUs and adds logits through \texttt{torch.distributed}; we compute both passes in one process with one copy (the arithmetic is the same) \\
TCR & official & all three training stages with the authors\textquotesingle{} scripts and default hyper-parameters; 20 soft tokens as in the paper & see below \\
Astute RAG & none released & prompts from the paper\textquotesingle s appendix, t = 1, m = 1; zero generated passages allowed when the model does not know, as in the paper & \\
Self-Route & none released & prompts from the paper & the route chosen is between RAG and no context (as in Astute RAG\textquotesingle s use of it), not RAG and long context \\
USC & none released & the consistency-selection prompt from the paper & \\
\end{longtable}}

\textbf{TCR.} The repository contains neither a checkpoint nor training data, so TCR is retrained with the authors\textquotesingle{} code on Llama-3.1-8B-Instruct (4-bit, as in the code). Stage-1 data (5,000 Wikidata conflict samples) is built from LC-QuAD train facts with paraphrases, contradictions and irrelevant variants generated by Qwen (the paper uses GPT-4o). Stage-2 data uses TCR\textquotesingle s own functions on 2,000 SQuAD questions. Stage-3 data uses the authors\textquotesingle{} script (1,000 records). Three faults in the released code had to be fixed to run it: an indentation error in \texttt{tcr\_e2e/prepare.py}; a mismatch between the parameter names saved in stage 2 and read in stage 3 (renamed in the checkpoint, code unchanged); and a default of 5 soft tokens against 20 in the paper (we use 20). TCR is not retrained on GRACE data: it is trained with its own method on its own data, and its stage-1 data comes from the same knowledge graph as GRACE.

\textbf{Failure counts.} With Llama, TruthfulRAG\textquotesingle s official pipeline fails on 183 of 2,700 GRACE dev records (6.8\%) after one re-run; they are scored as abstentions.

\textbf{Methods that return long text} (InstructRAG rationales, Astute RAG, the two-voiced answer) pass through the shared extractor of \cref{app:prompts}, as does every answer longer than four words from any method.

\subsection{Prompts}
\label{app:prompts}

All prompts are used verbatim for both backbones. Thinking is disabled. Decoding is greedy (temperature 0) except for the five sampled elicitation runs (temperature 0.7, top-p 0.95, one request with n = 5).

\textbf{A.1 Knowledge-note elicitation} (system message; at most 900 generated tokens).

{\scriptsize\begin{verbatim}
You are a knowledgeable assistant. You are given a question and facts retrieved from a
knowledge graph. Write down what YOU know from your own background knowledge that relates to the
question, to the entities mentioned, or to the given facts. Do not answer the question directly in
prose. Write knowledge as short factual notes:
- confirm: a given fact agrees with what you know (cite its id)
- contradict: a given fact disagrees with what you know; state what you believe instead (cite its id)
- add: another fact you know about an entity that matters for the question
- bridge: a fact you know that connects the question's entities toward what is asked
Each note is one triple (subject, relation, object) using plain entity names. Include only things you
believe; it is fine to state useful facts that the given facts do not contain. At most 8 notes.
Return JSON only: {"notes":[{"stance":"confirm|contradict|add|bridge","about":"F1 or null",
"subject":"...","relation":"...","object":"..."}]}
\end{verbatim}}

User message: \texttt{Question:\ \textless{}question\textgreater{}}, a blank line, \texttt{Retrieved\ facts:}, then one line \texttt{F\textless{}i\textgreater{}:\ \textless{}fact\ text\textgreater{}} per context fact.

\textbf{A.2 Self-verification of a note} (one call per note; the note is shown alone, with no question and no context). The score is \(P(\text{Yes}) / (P(\text{Yes}) + P(\text{No}))\) from the log-probabilities of the first generated token.

{\scriptsize\begin{verbatim}
You check whether a knowledge-graph fact is consistent with what you know about the world. Reply
with exactly one word: Yes if it is correct or very likely correct, No if it is wrong or implausible.
\end{verbatim}}

\textbf{A.3 Plain reader} (also the \emph{plain reader} baseline; at most 64 generated tokens). The first block is the evidence policy shared with the multiple-choice form of GRACE, which is why it mentions options; the answer format that follows makes the task open-form.

{\scriptsize\begin{verbatim}
--- BENCHMARK EVIDENCE POLICY ---

You are answering a multiple-choice question with two possible evidence sources:
the supplied context and your learned knowledge.

1. Read the question and every option carefully. Treat numeric entities as
   actual values and resolve ties according to the stated relation.
2. Consider relevant evidence in the supplied context and relevant knowledge
   already learned by the model. Neither source is automatically authoritative.
3. If the sources disagree or one is incomplete, assess the question's
   semantics and the reliability of the available evidence. Do not blindly
   follow either source, and do not invent facts.
4. Select the option that is best supported. Select the option for insufficient
   information only when no option can be supported reliably.
5. Do not expose private chain-of-thought or add explanations unless the
   runner explicitly requests them.

The supplied context is a set of knowledge-graph facts. If neither source supports an answer,
reply exactly: INSUFFICIENT. Give only the answer entity name(s), most likely first, separated
by ' | '. No explanation.
\end{verbatim}}

The closed-book arm uses "You answer questions from your own learned knowledge." followed by the same answer format. The context-only arm uses "You are a careful knowledge-graph question answering reader. Answer using only the graph facts supplied. Do not use outside memory or add unstated facts. If the facts do not contain the answer, reply exactly: INSUFFICIENT." followed by the same answer format.

\textbf{A.4 Two-voiced reader} (at most 450 generated tokens).

{\scriptsize\begin{verbatim}
You answer a question using two clearly separated sources: (1) the retrieved facts (the provided
information) and (2) background-knowledge notes stating what you yourself know, each with a confidence
wording. A separate careful reader has already read the retrieved facts; its reading is given as
"Context reading". Report each source honestly and separately, then conclude.
- context_says: the context reading (copy it; write NONE if it is INSUFFICIENT)
- knowledge_says: the answer entity your background knowledge supports, or NONE
- relation: agree | complement | conflict | context_silent
- final: the single entity you conclude is factually correct, or NONE if you cannot tell.
  The retrieved facts are usually right: only conclude against them when a note you are confident
  about directly contradicts them. A note worded "not sure" or "believe" is not enough to override.
- answer: one or two sentences: "According to the provided information, X. As far as I know, Y."
  Use hedges (I am confident / I believe / I am not sure but possibly) matching the notes' confidence.
Return JSON only: {"context_says":"...","knowledge_says":"...","relation":"...",
                   "final":"...","answer":"..."}
\end{verbatim}}

User message: the question; the context facts as \texttt{F\textless{}i\textgreater{}:\ \textless{}text\textgreater{}}; the notes, one per line as \texttt{-\ \textless{}stance\textgreater{},\ \textless{}hedge\textgreater{}\ (about\ F\textless{}i\textgreater{}):\ \textless{}subject\textgreater{}\ -\/-\textless{}relation\textgreater{}-\/-\textgreater{}\ \textless{}object\textgreater{}}; and the line \texttt{Context\ reading:\ \textless{}plain\ reader\textquotesingle{}s\ answer\textgreater{}}. The hedge is "I am confident" for $c_j^{*} \ge 0.8$, "I believe" for $0.5 \le c_j^{*} < 0.8$ and "I am not sure, but possibly" below 0.5. Bridge notes are shown with the stance word "add".

\textbf{A.5 Answer extractor} (shared by the selector\textquotesingle s candidate shortening and by the scoring protocol for every method; Qwen, at most 32 generated tokens; it sees the question and the response only).

{\scriptsize\begin{verbatim}
You read a question and a response to it. Reply with only the final answer entity that the response
commits to (a short name), or NONE if the response gives no answer or says it cannot answer.
\end{verbatim}}

%% file: appendix/D_diagnosis.tex
\section{Per-state results and the divergence analysis}
\label{app:diagnosis}

\subsection{Every method by context state}
\label{app:perstate}

All numbers: GRACE dev, protocol v3, C−W. Qwen = Qwen3.5-35B-A3B; Llama = Llama-3.1-8B-Instruct.

\textbf{Single-source reference arms}

\begin{center}\footnotesize\setlength{\tabcolsep}{3.5pt}
\begin{tabular}{@{}lrr@{}}
\toprule
 & \multicolumn{1}{c}{Qwen} & \multicolumn{1}{c}{Llama} \\
\cmidrule(l{3pt}r{3pt}){2-2}\cmidrule(l{3pt}r{3pt}){3-3}
 & \makecell[r]{correct / wrong\\/ abstain (\%)} & \makecell[r]{correct / wrong\\/ abstain (\%)} \\
\midrule
closed-book (Acc1) & 43.4 / 44.6 / 12.0 & 39.4 / 39.6 / 21.1 \\
context-only reader & 44.1 / 23.6 / 32.2 & 34.4 / 10.3 / 55.3 \\
\bottomrule
\end{tabular}\end{center}

Closed-book answering is right as often as it is wrong (C−W ≈ 0), so model knowledge cannot simply be trusted. The context-only reader scores 71.1 C−W on clean contexts (Qwen) but −66.7 on wrong\_branch, where it is led to the wrong entity.

\textbf{Per-state C−W, Qwen}

\begin{center}\footnotesize\setlength{\tabcolsep}{3.5pt}
\begin{tabular}{@{}llrrrrr@{}}
\toprule
Family & Method & All & clean & broken & \makecell[r]{wrong\_\\branch} & competing \\
\midrule
follow the context & context-only reader & 20.5 & 71.1 & 35.0 & −66.7 & 42.7 \\
follow the context & plain reader & 22.6 & 70.5 & 37.3 & −63.9 & 46.5 \\
follow the context & FaithfulRAG & 21.2 & 71.4 & 37.9 & −74.1 & 49.6 \\
follow the context & TruthfulRAG & 17.4 & 68.0 & 32.6 & −70.7 & 39.6 \\
follow the context & Self-Route & 19.6 & 67.4 & 34.8 & −60.7 & 36.9 \\
implicit choice & InstructRAG-ICL & 23.7 & 64.0 & 38.7 & −58.7 & 51.0 \\
implicit choice & USC & 29.4 & 71.6 & 45.9 & −56.9 & 57.2 \\
memory first & Astute RAG & 21.7 & 40.1 & 24.6 & −8.4 & 30.4 \\
memory first & GenRead & −0.2 & 1.2 & 1.0 & −1.5 & −1.5 \\
\textbf{ours (π = 0.25)} &  & \textbf{45.3} & 64.3 & 44.1 & \textbf{17.9} & 55.0 \\
\bottomrule
\end{tabular}\end{center}

\textbf{Per-state C−W, Llama}

\begin{center}\footnotesize\setlength{\tabcolsep}{3.5pt}
\begin{tabular}{@{}llrrrrr@{}}
\toprule
Family & Method & All & clean & broken & \makecell[r]{wrong\_\\branch} & competing \\
\midrule
follow the context & context-only reader & 24.2 & 53.3 & 23.3 & −18.5 & 38.7 \\
follow the context & plain reader & 27.5 & 55.7 & 31.7 & −19.1 & 41.8 \\
follow the context & FaithfulRAG & 14.2 & 60.9 & 25.2 & −66.4 & 37.2 \\
follow the context & TruthfulRAG & 12.7 & 58.1 & 26.5 & −58.1 & 24.3 \\
follow the context & Self-Route & 19.1 & 64.3 & 35.9 & −57.8 & 34.2 \\
decoding/trained control & CK-PLUG & 11.4 & 53.3 & 19.0 & −41.3 & 14.5 \\
decoding/trained control & TCR & 17.0 & 56.7 & 24.7 & −41.5 & 27.9 \\
decoding/trained control & CAD & 16.3 & 60.3 & 35.6 & −45.5 & 14.7 \\
decoding/trained control & AdaCAD & 15.0 & 66.8 & 37.3 & −61.5 & 17.3 \\
implicit choice & InstructRAG-ICL & 20.1 & 60.7 & 30.7 & −46.4 & 35.4 \\
implicit choice & USC & 26.5 & 61.8 & 30.7 & −35.6 & 49.2 \\
memory first & Astute RAG & 10.1 & 21.5 & 11.3 & −8.9 & 16.4 \\
memory first & GenRead & −6.2 & −6.2 & −6.2 & −6.2 & −6.2 \\
\textbf{ours (π = 0.25)} &  & \textbf{35.0} & 57.6 & 37.6 & \textbf{−4.1} & 48.9 \\
\bottomrule
\end{tabular}\end{center}

\textbf{Findings}

\begin{enumerate}
\def\labelenumi{\arabic{enumi}.}
\item
  \textbf{Follow-the-context methods collapse on wrong\_branch.} With Qwen every one of them scores between −61 and −74. They are built to trust the context, so a replaced answer edge is followed.
\item
  \textbf{Memory-first methods do not collapse on wrong\_branch but lose the clean case.} Astute RAG scores −8 on wrong\_branch and only 40 (Qwen) or 22 (Llama) on clean. GenRead\textquotesingle s score is near zero and identical across states: its answer does not depend on the context.
\item
  \textbf{Decoding/trained control methods fall below the plain reader.} On Llama, CK-PLUG, TCR, CAD and AdaCAD score 11.4, 17.0, 16.3 and 15.0 against 27.5. AdaCAD has the highest clean score of any method (66.8 against 55.7 for the plain reader) and loses three times as much on wrong\_branch contexts (−61.5 against −19.1); CAD shows the same pattern at a smaller scale (60.3 and −45.5). \cref{sec:diagnosis} traces this to what their weight measures.
\item
  \textbf{No baseline is strong on both clean and wrong\_branch.} This is the trade-off that \cref{sec:method} makes explicit and adjustable.
\item
  \textbf{The model has usable knowledge.} On one dev slice (\cref{app:cues}), on questions the model answers wrongly without context, 37.8\% of the time it still states the missing edge correctly: partial knowledge that a closed-book answer does not reveal.
\end{enumerate}

\cref{fig:states} summarises the diagnosis. The wrong\_branch column separates the families: follow-the-context methods and implicit-choice methods are confidently wrong there (−36 to −74), the four decoding/trained control methods lie between −41 and −62, and the memory-first methods avoid the loss only by giving up most of the clean column.

\subsection{The divergence weight, all records and all tokens}
\label{app:divergence}
\begin{table}[!htbp]
  \centering\footnotesize
  \caption{Records of the wrong\_branch state whose question the model answers correctly on its own (Llama, $n = 267$): the model is right and the context is wrong. Percent of records.}
  \label{tab:knows}
  \begin{tabular}{@{}lccc@{}}
    \toprule
    & correct & abstain & wrong \\
    \midrule
    plain reader & 12.0 & 68.2 & 19.9 \\
    CAD (fixed $\alpha = 1$) & 11.2 & 40.8 & 47.9 \\
    AdaCAD ($\alpha$ = divergence) & 12.7 & 17.6 & 69.7 \\
    ours (\cref{sec:method}) & \textbf{32.2} & 51.3 & \textbf{16.5} \\
    \bottomrule
  \end{tabular}
\end{table}

\begin{figure}[!htbp]
  \centering
  \includegraphics[width=0.95\linewidth]{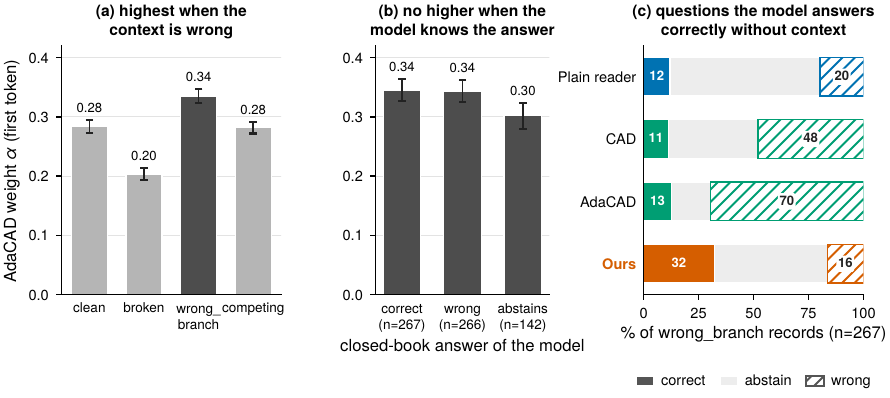}
  \caption{What AdaCAD's weight measures (Llama-3.1-8B, \grace{} development data, 95\% intervals). (a) $\alpha$ is highest when the context is wrong. (b) Within wrong contexts it is no higher when the model is right. (c) On wrong contexts that the model can answer, amplification converts abstentions into wrong answers.}
  \label{fig:divergence}
\end{figure}

CAD decodes from \((1+\alpha)\,\ell_{\text{ctx}} - \alpha\,\ell_{\text{no-ctx}}\), the logits with and without the context, with a fixed $\alpha = 1$. AdaCAD sets $\alpha$ at every step to the Jensen--Shannon divergence between the two next-token distributions, computed with the natural logarithm in the official code; it therefore lies between 0 (identical distributions) and $\ln 2 \approx 0.693$ (disjoint ones), and the weight on the no-context logits, $-\alpha$, lies between $-0.693$ and 0. The design rests on one assumption: when the two distributions diverge, the context carries information the model lacks, so the context should be amplified. GRACE tests the assumption directly, because each question is asked with a right and a wrong context and the model\textquotesingle s answer without any context is known. We read $\alpha$ at the first generated token from the saved AdaCAD run (Llama-3.1-8B, 675 questions, no new model call). Intervals are 95\% bootstrap intervals over questions.

\begin{center}\footnotesize\setlength{\tabcolsep}{3.5pt}
\begin{tabular}{@{}lrrrr@{}}
\toprule
 & clean & broken & \makecell[r]{wrong\_\\branch} & competing \\
\midrule
AdaCAD weight α, first token & 0.284\,{\scriptsize[0.273, 0.295]} & 0.203\,{\scriptsize[0.194, 0.213]} & \textbf{0.335}\,{\scriptsize[0.324, 0.347]} & 0.282\,{\scriptsize[0.272, 0.292]} \\
\bottomrule
\end{tabular}\end{center}

\textbf{The divergence responds to a wrong context.} α is highest on wrong\_branch contexts. Against the clean version of the same question it is higher by 0.051 ({[}0.041, 0.062{]}), and higher for 61.9\% of the questions. As a detector of a wrong context it is weak but real (AUC 0.60).

\textbf{It does not respond to which source is right.} Within wrong\_branch contexts, α is 0.345 {[}0.327, 0.364{]} on the 267 questions the model answers correctly without context and 0.343 {[}0.326, 0.362{]} on the 266 it answers wrongly; the difference is 0.002 ({[}−0.024, 0.028{]}). On the 142 questions it declines to answer, α is 0.302. The weight is as large when the model is about to be corrupted as when it is about to be corrected.

\textbf{The push is therefore strongest where it does the most harm.} We split the wrong\_branch records into thirds by α. AdaCAD scores −38.2, −67.1 and −79.1 C−W from the lowest third to the highest, a drop of 40.9 points ({[}−51.6, −29.6{]}). The plain reader, which uses no such weight, scores −18.2, −14.2 and −24.9 (difference −6.7 {[}−16.3, 3.1{]}). The table below takes the 267 wrong\_branch records whose question the model answers correctly without context: the cases in which the model\textquotesingle s knowledge is right and the context is wrong.

\begin{center}\footnotesize\setlength{\tabcolsep}{3.5pt}
\begin{tabular}{@{}lrrr@{}}
\toprule
\makecell[l]{wrong\_branch, model\\knows the answer (n =\\267)} & correct (\%) & abstain (\%) & wrong (\%) \\
\midrule
plain reader & 12.0 & 68.2 & 19.9 \\
CAD & 11.2 & 40.8 & 47.9 \\
AdaCAD & 12.7 & 17.6 & 69.7 \\
\textbf{ours, π = 0.25} & \textbf{32.2} & 51.3 & \textbf{16.5} \\
\bottomrule
\end{tabular}\end{center}

Faced with a wrong context, the plain reader mostly abstains. Contrastive decoding converts those abstentions into wrong answers (19.9\% → 47.9\% → 69.7\%) and recovers none of the model\textquotesingle s knowledge (12.0\% → 11.2\% → 12.7\% correct). Our system recovers the correct answer on 32.2\% of these records, 19.5 points more than AdaCAD ({[}13.1, 25.8{]}), with no more wrong answers than the plain reader (16.5\% against 19.9\%; difference 3.4 {[}−3.0, 9.4{]}).

The same amplification explains AdaCAD\textquotesingle s clean score. On these questions in the clean state, AdaCAD is correct on 92.8\%, against 81.1\% for the plain reader and 87.2\% for ours. On GRACE, AdaCAD is the strongest method in our comparison when the context is right and among the weakest when it is wrong, for one reason: it amplifies the context in proportion to a quantity that does not depend on whether the context is right.

\textbf{Scope of the analysis.} It covers one method\textquotesingle s weight at the first generated token; the score analysis uses Llama, and the premise is also checked with Qwen (end of this paragraph). Because the official prompt ends with "Answer:", that token is the first token of the answer whenever the generation starts with a name; this holds for 52\% to 59\% of records per state (first line of at most five words), the others starting with a sentence such as "There is no information \ldots". Restricted to those records, the two quantities above become sharper, not weaker: α is higher on the wrong\_branch than on the clean version of the same question by 0.069 ({[}0.051, 0.086{]}; higher for 68\% of 300 questions), and within wrong\_branch contexts it is 0.375 when the model knows the answer and 0.367 when it does not (difference 0.008 {[}−0.022, 0.037{]}; 172 and 151 questions). Averaged over all generated tokens, α is lower (0.10 to 0.17), the gap between wrong\_branch and clean shrinks to 0.010 ({[}0.004, 0.015{]}), and the gradient across thirds disappears (−1.8 {[}−13.7, 9.4{]}). Under that definition α is slightly lower when the model knows the answer (0.154 against 0.171). Under neither definition is it higher when the model is right. \textbf{Second backbone.} We repeated the first-token measurement with Qwen3.5-35B-A3B, computing the same divergence from the two next-token distributions under AdaCAD\textquotesingle s prompts (\texttt{divergence\_qwen.py}; the server returns the top 100 tokens of each distribution, which hold 97--98\% of the mass, and the rest is one bucket, so the value is a lower bound on the exact divergence). Qwen\textquotesingle s first generated token is the first token of a name for 674 of 675 question pairs, so the two definitions coincide. α is higher on the wrong\_branch than on the clean version of a question by 0.008 ({[}0.005, 0.012{]}; higher for 58\% of questions), and within wrong\_branch contexts it is 0.169 when the model knows the answer and 0.169 when it does not (difference 0.000 {[}−0.010, 0.011{]}; 294 and 296 questions). The premise of \cref{prop:symmetric} holds on both backbones. Qwen\textquotesingle s divergences are smaller than Llama\textquotesingle s, so on this backbone the signal is also a weaker detector of a wrong context. We did not run AdaCAD itself with Qwen, and we did not analyse the entropy-based weight of CK-PLUG.

\textbf{Scope of \cref{prop:symmetric}.} It concerns symmetric statistics such as the Jensen--Shannon divergence. Entropy-based weights, such as CK-PLUG\textquotesingle s, are not symmetric in the two sources, so the proposition does not apply to them; we did not test whether their weight depends on which source is right.

%% file: appendix/E_method.tex
\section{Method details}
\label{app:method}

\subsection{Three findings that fixed the design}
\label{app:findings}

Before the method below, we tried simpler ways to use model knowledge. Three negative results fixed the design (details in Supplementary Section~\ref{app:earlier}).

\begin{enumerate}
\def\labelenumi{\arabic{enumi}.}
\item
  \textbf{Do not edit the context.} We let the model verify each context fact and deleted the ones it rejected. The verification was accurate (AUC 0.875), but deletion removed the relation the reader needed, and the reader switched to "insufficient information"; the result was no better than random deletion. → The context is passed to the reader unedited.
\item
  \textbf{Do not attach confidence to context facts.} Annotating each context fact with the model\textquotesingle s belief lowered the score by 1 to 4 points: a low score mixes "the model does not know" with "the fact is false", so correct facts were distrusted. → Confidence is attached to the model\textquotesingle s own statements (notes), not to the context.
\item
  \textbf{Context silences knowledge.} Take the questions the model answers correctly with no context (GRACE dev, protocol v3; 265 to 296 questions per state and backbone). Given a broken context and a prompt that permits its own knowledge, the plain reader is correct on only 66.2\% (Qwen) and 68.3\% (Llama) of them and abstains on 31.4\% and 26.0\%. Given a wrong\_branch context it is correct on 8.2\% and 12.0\%: Qwen copies the wrong fact (72.4\% wrong) and Llama abstains (68.2\%). A prompt that merely permits using knowledge is not enough. → Knowledge is elicited as explicit notes in a separate step, before the context is read for an answer. With the notes, the same questions are answered correctly at 79.1\% and 74.7\% (broken) and 67.0\% and 32.2\% (wrong\_branch). We first saw this effect on CWQ validation data with an earlier scoring rule (Supplementary Section~\ref{app:earlier}).
\end{enumerate}

\subsection{Knowledge notes and their labels}
\label{app:notes}

One prompt receives the question and the numbered context facts (no options, no answer) and returns up to eight notes. Each note is a triple with a stance:

{\footnotesize\setlength{\tabcolsep}{3pt}
\begin{longtable}[]{@{}>{\raggedright\arraybackslash}p{0.243\linewidth}>{\raggedright\arraybackslash}p{0.732\linewidth}@{}}
\toprule\noalign{}
Stance & Meaning \\
\midrule\noalign{}
\endhead
\bottomrule\noalign{}
\endlastfoot
confirm & context fact Fi matches what the model knows \\
contradict & context fact Fi is wrong according to the model; the object is the value the model believes \\
add & additional knowledge about a related entity \\
bridge & knowledge that connects the question to the answer \\
\end{longtable}}

The prompt is run six times: once greedily and five times at temperature 0.7. Notes with the same normalised subject and object are merged. The \textbf{support} $k_j \in \{1,\dots,6\}$ of note $n_j$ is the number of runs, out of six, that produced it. The ten notes with the highest support are kept.

The idea is that what a model knows it states every time, and what it guesses it states only sometimes. Example (question about Dawkins\textquotesingle s adviser, broken state): "Dawkins → doctoral adviser → Tinbergen" has support $k_j = 6$; "Tinbergen → doctoral adviser → Karl von Frisch" has support $k_j = 1$ and is false.

\textbf{Labels.} No human annotation. From the GRACE construction: a note that matches the removed edge, or that correctly contradicts an injected fact, is true and useful; a note that confirms an injected fact or contradicts a correct one is false. Remaining notes are checked against Wikidata (all current values of the subject--relation pair, with aliases); Wikidata-derived "false" labels are down-weighted by 0.5. Unverifiable notes are dropped.

\textbf{Per-backbone calibration.} A policy trained on one model\textquotesingle s notes transfers poorly to another (AUC 0.66--0.69, ECE 0.10 for the Qwen policy on Llama notes; AUC 0.82--0.83, ECE 0.02 when retrained; \cref{app:policy}). Retraining takes minutes and needs no annotation.

\subsection{Note policy}
\label{app:policy}

\textbf{Input of a note token.} The bge-m3 embedding of the note text (1,024 dimensions) concatenated with 15 scalars: the self-verification score (\cref{app:prompts}); the stance (four indicators); whether the note cites a context fact; whether its subject and its object occur in the context and in the question (four indicators); whether a context fact has the same subject and relation with a different object, or with the same object (two indicators); the cosine similarity of the note to the question; the support $k_j$; and whether the note appears in the greedy run. The question and each context fact are tokens holding their bge-m3 embedding. No hidden state of the LLM is used.

\textbf{Training.} 1,000 LC-QuAD train questions (disjoint from every dev slice and from the GRACE test split) in four states give 4,000 records and about 24,000 notes. The loss is binary cross-entropy on both heads, with Wikidata-only negative labels weighted by 0.5; AdamW, learning rate $3 \times 10^{-4}$, weight decay 0.01, 15 epochs, and the checkpoint with the lowest loss on a 10\% split by question is kept. Training takes minutes on CPU. For selector training, note confidences on the pool are cross-fitted in five folds, so every note is scored by a policy that has not seen it.

\textbf{Calibration on GRACE dev} (three slices; about 2,850 verifiable notes per slice).

\begin{center}\footnotesize\setlength{\tabcolsep}{3.5pt}
\begin{tabular}{@{}>{\raggedright\arraybackslash}p{0.27\linewidth}rr@{}}
\toprule
 & Qwen & Llama \\
\midrule
AUC of $p_j$ (true against false) & 0.816 / 0.838 / 0.832 & 0.822 / 0.837 / 0.830 \\
ECE of $p_j$ & 0.025 / 0.013 / 0.053 & 0.041 / 0.026 / 0.032 \\
false notes voiced "I am confident" ($c_j^{*} \ge 0.8$) & 2.9 / 2.2 / 3.1\% & 2.9 / 2.6 / 2.7\% \\
\ldots{} if every note were voiced confidently & 20.7 / 20.4 / 25.4\% & 25.5--27.8\% \\
AUC of the self-verification score alone & 0.60--0.65 &  \\
AUC with labels from GRACE construction only (no Wikidata labels) & 0.61--0.67 &  \\
\bottomrule
\end{tabular}\end{center}

\textbf{What did not help.} (i) The LLM\textquotesingle s hidden state at the verification prompt: AUC 0.816 / 0.830 / 0.830 with it, 0.815 / 0.837 / 0.830 without. We also found that the server returned different hidden states for the same prompt depending on prefix caching (cosine 0.18--0.63), which a random prefix fixes; log-probabilities are not affected. (ii) A larger graph-structured policy with local and global attention over typed edges: AUC 0.78 / 0.82 / 0.79, below the plain transformer. With about 13 nodes per record and 3,600 training records, the scalar features already carry the structure.

\textbf{Transfer across backbones.} The Qwen policy applied to Llama notes without retraining gives AUC 0.66--0.69, ECE 0.10--0.11 and 10.1--10.6\% confident false notes, against 0.82--0.83, 0.02--0.03 and 2.0--2.3\% when retrained on Llama notes. Transfer still removes most confident errors (from about 27\% to about 10\%), but calibration is poor, so the policy is calibrated per backbone.

\subsection{Selector}
\label{app:selectordetails}

\textbf{Candidates.} The reader\textquotesingle s \texttt{final}, \texttt{context\_says} and \texttt{knowledge\_says}, and the plain reader\textquotesingle s answer. Any candidate longer than four words is shortened by the shared extractor, so every candidate is a short answer.

\textbf{Model.} A gradient-boosting classifier (Friedman, 2001; 300 trees, depth 3) predicts, for each candidate separately, the probability that it is correct. Features: which source the candidate comes from; the reader\textquotesingle s relation; embedding similarities between the candidate and each source and the question; and note statistics for the notes most similar to the candidate (calibrated confidence, support, stance). Similarity is by bge-m3 embedding with soft aggregation, with no string matching and no hand-set threshold. Labels use the strict matching rule of \cref{sec:grace}.

\textbf{Decision.} Let \(P(a)\) be the predicted probability that candidate $a$ is correct. The selector chooses \(\hat{a} = \arg\max_a P(a)\). In the \emph{cautious} mode it commits only if \(P(\hat{a}) > 0.5\), the break-even point when a correct answer scores $+1$ and a wrong one $-1$; otherwise it answers "I don\textquotesingle t know".

\textbf{Training data.} GRACE pool plus GRACE-QT train. Note objects are not candidates: on question types other than slot filling a note\textquotesingle s object is often an intermediate entity rather than the answer (\cref{app:lockbox2}). Under the final scoring protocol this choice costs 1.3 to 1.5 points on GRACE (Supplementary Section~\ref{app:selector}).

Three remarks on \cref{prop:shift}.

\begin{itemize}
\item
  \emph{Meaning of π.} The proposition treats $\pi$ as the probability that a context-sourced candidate is correct. This equals the share of reliable contexts when the reader reproduces the context\textquotesingle s answer whenever the context is right; reading errors make the two differ.
\item
  \emph{Robustness of the direction.} If label shift fails, for example on new question types, \(P'(a)\) is no longer the exact posterior, but \cref{eq:shift} remains strictly increasing in $\pi$ for every candidate and every data set. The direction of the control holds without the assumption; only its calibration needs it. This is why the analytic form replaced a learned π input, whose effect vanished outside the training distribution (Supplementary Section~\ref{app:earlier}).
\item
  \emph{What is not shifted.} In a conflict the context\textquotesingle s answer and the model\textquotesingle s answer cannot both be right, so the exact posterior of a knowledge-sourced candidate falls as $\pi$ rises. \cref{eq:shift} leaves it unchanged. The choice still moves toward context candidates as $\pi$ rises, but more slowly than under the exact posterior, and at a high $\pi$ a knowledge candidate can pass the threshold of \cref{prop:commit} more often than the exact posterior would allow. \cref{eq:shift} is therefore exact for context candidates and errs toward keeping knowledge candidates, in the direction the design prefers (never suppress model knowledge).
\end{itemize}

%% file: appendix/F_results.tex
\section{Further results on GRACE}
\label{app:results}

\subsection{Per-state scores at every prior}
\label{app:pitable}
\begin{table}[!htbp]
  \centering\small
  \setlength{\tabcolsep}{4pt}
  \caption{Ours with $\pi = f$ minus the best baseline at each share $f$ of right contexts (\grace{} development data, 95\% intervals). The best baseline is chosen after seeing the scores.}
  \label{tab:envelope}
  \begin{tabular}{@{}lcccccc@{}}
    \toprule
    $f = \pi$ & 0.1 & 0.25 & 0.5 & 0.75 & 0.9 & 0.99 \\
    \midrule
    Qwen: best baseline & GenRead & Astute RAG & Astute RAG & USC & USC & USC \\
    \quad ours $-$ best & $+25.7$ & $+25.8$ & $+21.3$ & $+11.4$ & $+3.5$ & $+1.5$ \\
    & \scriptsize[19.3, 32.1] & \scriptsize[20.2, 31.5] & \scriptsize[16.3, 26.4] & \scriptsize[7.7, 15.1] & \scriptsize[$-0.4$, 7.3] & \scriptsize[$-2.5$, 5.5] \\
    Llama: best baseline & Astute RAG & plain reader & plain reader & USC & AdaCAD & AdaCAD \\
    \quad ours $-$ best & $+10.7$ & $+11.7$ & $+4.4$ & $+4.1$ & $+2.1$ & $+0.7$ \\
    & \scriptsize[4.9, 16.9] & \scriptsize[7.2, 16.2] & \scriptsize[0.4, 8.4] & \scriptsize[$-0.6$, 8.8] & \scriptsize[$-1.8$, 6.1] & \scriptsize[$-3.6$, 4.9] \\
    \bottomrule
  \end{tabular}
\end{table}

Per-state scores at every π (GRACE dev, protocol v3, cautious mode; C−W, with correct / wrong in parentheses for All; comp.\ = competing).

{\footnotesize\setlength{\tabcolsep}{3pt}
\begin{longtable}[]{@{}l*{5}{l}*{5}{l}@{}}
\toprule\noalign{}
& \multicolumn{5}{c}{Qwen3.5-35B-A3B} & \multicolumn{5}{c}{Llama-3.1-8B} \\
\cmidrule(lr){2-6}\cmidrule(l){7-11}
π & All & clean & broken & wrong & comp. & All & clean & broken & wrong & comp. \\
\midrule\noalign{}
\endhead
\bottomrule\noalign{}
\endlastfoot
0.1 & 44.8 (56.9 / 12.0) & 61.2 & 43.0 & 20.4 & 54.7 & 33.3 (45.4 / 12.0) & 53.5 & 34.7 & −0.6 & 45.8 \\
0.25 & 45.3 (58.4 / 13.1) & 64.3 & 44.1 & 17.9 & 55.0 & 35.0 (48.5 / 13.6) & 57.5 & 37.6 & −4.1 & 48.9 \\
0.5 & 43.4 (59.7 / 16.3) & 70.2 & 44.9 & 4.0 & 54.4 & 34.1 (54.0 / 19.8) & 64.9 & 39.1 & −19.6 & 52.1 \\
0.75 & 38.8 (59.2 / 20.4) & 72.6 & 45.6 & −14.5 & 51.6 & 30.0 (55.1 / 25.1) & 66.8 & 38.2 & −34.4 & 49.5 \\
0.9 & 30.7 (55.9 / 25.2) & 73.3 & 45.9 & −37.9 & 41.5 & 26.4 (54.6 / 28.3) & 67.4 & 38.2 & −45.6 & 45.5 \\
0.99 & 21.9 (51.5 / 29.7) & 73.2 & 46.2 & −63.7 & 31.7 & 24.4 (54.1 / 29.7) & 67.4 & 38.4 & −50.7 & 42.5 \\
\end{longtable}}

Without the cautious rule (always answering with the top candidate), All at π = 0.25 is 39.4 (Qwen) and 30.3 (Llama).

GRACE dev, protocol v3, cautious mode, C−W. $\pi_0 = 0.25$ is the trained setting.

\begin{figure}[!htbp]
  \centering
  \includegraphics[width=0.95\linewidth]{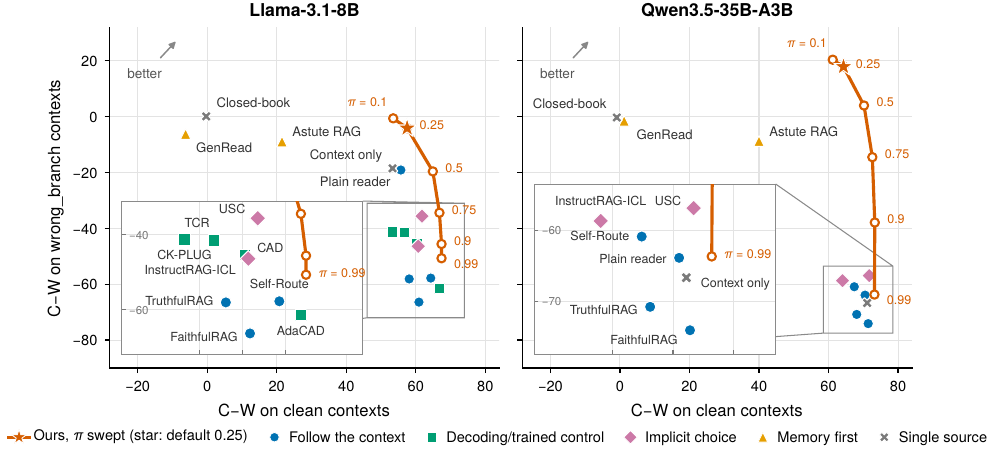}
  \caption{Clean against wrong\_branch score on GRACE dev (protocol v3). Each marker is one method; the line is our system as the context-trust prior π is swept from 0.1 to 0.99 (star: the trained setting, 0.25). Baselines of every family lie in the lower right (good on clean contexts, confidently wrong on wrong\_branch ones) or on the left (safe but uninformative). Our curve covers the region above them and reaches the context-following corner at π = 0.99. Inset: the dense region of the Qwen panel, enlarged.}
  \label{fig:pareto}
\end{figure}

\begin{itemize}
\item
  π moves the system along a clean-versus-wrong\_branch curve (\cref{fig:pareto}). No single setting is best for all states.
\item
  At high π the clean score exceeds the plain reader\textquotesingle s (Llama: 67.4 against 55.7; Qwen: 73.3 against 70.5). "Trust the context" does not mean "use only the context": when the plain reader finds no answer in a clean context, a knowledge candidate can still be chosen, and the two-voiced reader, which sees the notes, reads the context better than the plain reader does.
\item
  At π = 0.99 the Qwen system is the plain reader on wrong\_branch contexts (−63.7 against −63.9) and still better on clean and broken ones: the prior moves the system continuously from "merge" to "follow the context", and the context-following end is not worse than the context-following baseline.
\item
  The overall score is flat between π = 0.1 and π = 0.5 (Llama 33.3 to 35.0; Qwen 43.4 to 45.3), so the default does not need tuning on this mixture of states.
\end{itemize}

\subsection{Ablations}
\label{app:ablation}

GRACE dev, protocol v3, frozen selector v3, cautious mode, π = 0.25. Δ is the variant minus the full system, with a 95\% paired interval. For each variant the reader was re-run and the same frozen policy and selector were applied.

\begin{center}\footnotesize\setlength{\tabcolsep}{3.5pt}
\begin{tabular}{@{}>{\raggedright\arraybackslash}p{0.27\linewidth}rrrrrr@{}}
\toprule
 & \multicolumn{6}{c}{Qwen} \\
\cmidrule(l{3pt}r{3pt}){2-7}
Variant & All & Δ {[}CI{]} & clean & broken & \makecell[r]{wrong\_\\branch} & competing \\
\midrule
\textbf{full} & \textbf{45.3} &  & 64.3 & 44.1 & 17.9 & 55.0 \\
no notes (reader sees no model knowledge) & 23.6 & −21.7\,{\scriptsize[−24.6, −18.8]} & 68.6 & 38.4 & −47.6 & 35.0 \\
notes without confidence (all voiced "I am confident") & 38.3 & −7.1\,{\scriptsize[−9.7, −4.4]} & 61.5 & 36.7 & 7.7 & 47.1 \\
greedy elicitation only (no sampled runs) & 41.6 & −3.7\,{\scriptsize[−6.0, −1.4]} & 56.9 & 38.7 & 15.4 & 55.4 \\
no selector (reader\textquotesingle s own \texttt{final}) & 35.6 & −9.7\,{\scriptsize[−12.7, −6.9]} & 48.3 & 30.8 & 13.5 & 49.8 \\
\emph{plain reader} & \emph{22.6} & −22.7\,{\scriptsize[−25.7, −19.8]} & 70.5 & 37.3 & −63.9 & 46.5 \\
\bottomrule
\end{tabular}\end{center}

\begin{center}\footnotesize\setlength{\tabcolsep}{3.5pt}
\begin{tabular}{@{}lrrrrrr@{}}
\toprule
 & \multicolumn{6}{c}{Llama} \\
\cmidrule(l{3pt}r{3pt}){2-7}
Variant & All & Δ {[}CI{]} & clean & broken & \makecell[r]{wrong\_\\branch} & competing \\
\midrule
\textbf{full} & \textbf{35.0} &  & 57.6 & 37.6 & −4.1 & 48.9 \\
no notes & 20.6 & −14.4\,{\scriptsize[−17.7, −11.0]} & 46.7 & 29.2 & −25.8 & 32.1 \\
notes without confidence & 25.7 & −9.3\,{\scriptsize[−12.2, −6.5]} & 55.6 & 29.6 & −32.7 & 50.2 \\
greedy elicitation only & 31.7 & −3.3\,{\scriptsize[−5.6, −1.0]} & 57.0 & 32.4 & −9.3 & 46.8 \\
no selector (reader\textquotesingle s own \texttt{final}) & 26.1 & −8.9\,{\scriptsize[−12.0, −5.9]} & 47.3 & 24.4 & −13.2 & 45.8 \\
\emph{plain reader} & \emph{27.5} & −7.5\,{\scriptsize[−10.6, −4.4]} & 55.7 & 31.7 & −19.1 & 41.8 \\
\bottomrule
\end{tabular}\end{center}

\begin{figure}[!htbp]
  \centering
  \includegraphics[width=0.62\linewidth]{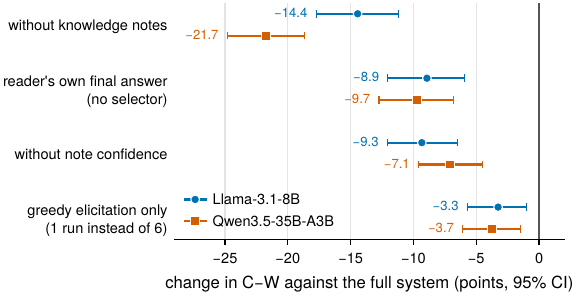}
  \caption{Change in C−W when one component is removed (GRACE dev, protocol v3, π = 0.25), with 95\% question-clustered bootstrap intervals. Every interval excludes zero on both backbones.}
  \label{fig:ablation}
\end{figure}

Reading (\cref{fig:ablation}):

\begin{itemize}
\item
  \textbf{Stating knowledge is the largest contribution} (−21.7 and −14.4 without notes). The loss is concentrated where the context is wrong or contradictory: wrong\_branch falls from 17.9 to −47.6 (Qwen) and competing from 55.0 to 35.0.
\item
  \textbf{Notes have a cost on clean contexts for the larger model} at π = 0.25: Qwen scores 68.6 on clean without notes and 64.3 with them. For Llama the notes help on clean as well (46.7 → 57.6). This cost is the one π trades off (\cref{sec:envelope}).
\item
  \textbf{Calibrated confidence is the second contribution} (−7.1 and −9.3). Without it, false notes voiced confidently mislead the reader: Llama\textquotesingle s wrong\_branch falls from −4.1 to −32.7.
\item
  \textbf{Sampling for support} adds 3.3 to 3.7 points for five extra elicitation runs.
\item
  \textbf{The selector} adds 8.9 to 9.7 points over the reader\textquotesingle s own conclusion, mainly on clean and broken.
\item
  Without notes, the pipeline is not better than a plain reader (Qwen 23.6 against 22.6; Llama 20.6 against 27.5): the gain comes from the knowledge, not from the extra calls.
\end{itemize}

\subsection{Are the spoken hedges calibrated?}
\label{app:hedge}

Accuracy of the "as far as I know" claim, grouped by the hedge the reader wrote (GRACE dev):

\begin{center}\footnotesize\setlength{\tabcolsep}{3.5pt}
\begin{tabular}{@{}lrr@{}}
\toprule
 & \multicolumn{1}{c}{Qwen} & \multicolumn{1}{c}{Llama} \\
\cmidrule(l{3pt}r{3pt}){2-2}\cmidrule(l{3pt}r{3pt}){3-3}
Reader wrote & n, accuracy & n, accuracy \\
\midrule
"I am confident / certainly" & 626, 79.9\% & 1,025, 70.9\% \\
"I believe / likely" & 120, 41.7\% & 484, 51.2\% \\
"not sure / possibly" & 99, 28.3\% & 237, 30.0\% \\
no hedge word & 1,758, 58.7\% & 712, 61.1\% \\
\bottomrule
\end{tabular}\end{center}

\begin{figure}[!htbp]
  \centering
  \includegraphics[width=0.5\linewidth]{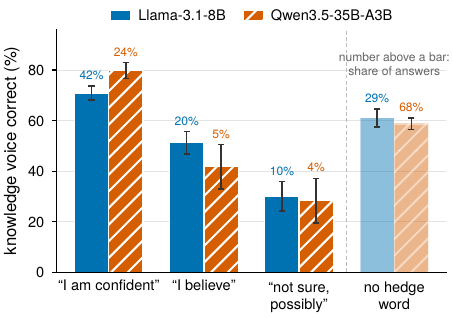}
  \caption{Accuracy of the knowledge part of the two-voiced answer by the hedge the reader wrote (GRACE dev; bars show 95\% binomial intervals; the number above a bar is the share of that backbone\textquotesingle s answers). Accuracy falls with the stated certainty on both backbones. Right of the dashed line: answers in which the reader wrote no hedge word.}
  \label{fig:hedge}
\end{figure}

Accuracy decreases with the stated certainty on both backbones. The weakness is coverage: the Qwen reader writes no hedge word in 67\% of answers (Llama: 29\%). The numeric note confidence itself has ECE of about 0.01--0.05 (\cref{app:policy}). The reader\textquotesingle s hedge words carry no numeric value, so we report the ordering of the three levels (\cref{fig:hedge}) and not an ECE for them.

\subsection{Questions that neither source answers alone}
\label{app:synergy}

On 37.9\% (Qwen) and 45.8\% (Llama) of dev records, both the closed-book arm and the context-only arm fail. On this subset:

\begin{center}\footnotesize\setlength{\tabcolsep}{3.5pt}
\begin{tabular}{@{}lrr@{}}
\toprule
 & \multicolumn{1}{c}{Qwen} & \multicolumn{1}{c}{Llama} \\
\cmidrule(l{3pt}r{3pt}){2-2}\cmidrule(l{3pt}r{3pt}){3-3}
 & \makecell[r]{correct / wrong\\(\% of all\\records)} & correct / wrong \\
\midrule
\textbf{Ours} & 8.1 / 9.9 & 11.0 / 9.6 \\
USC & 7.9 / 18.3 & 14.4 / 23.4 \\
TCR & --- & 14.7 / 26.6 \\
FaithfulRAG & 7.0 / 16.9 & 11.9 / 22.6 \\
TruthfulRAG & 6.7 / 17.3 & 12.4 / 21.8 \\
CK-PLUG & --- & 9.9 / 22.0 \\
plain reader & 2.1 / 14.2 & 5.3 / 10.7 \\
\bottomrule
\end{tabular}\end{center}

\begin{figure}[!htbp]
  \centering
  \includegraphics[width=0.85\linewidth]{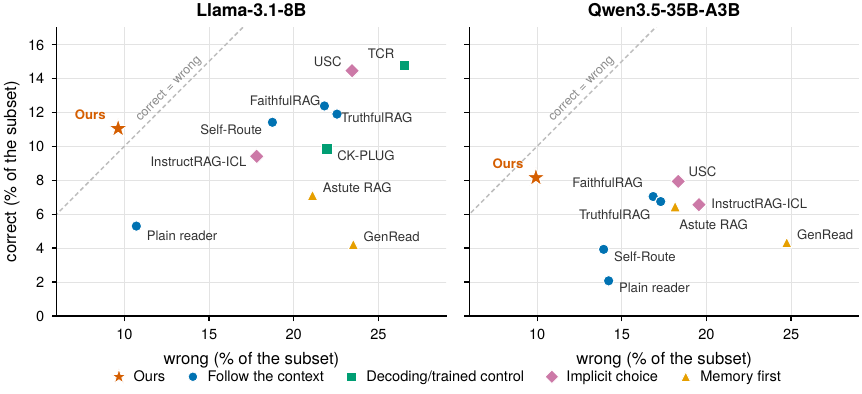}
  \caption{Records on which both the closed-book arm and the context-only arm fail (GRACE dev; percentages are of all records). Baselines recover a similar share of these records but are wrong on about twice as many. Ours is the only method on or above the line where correct equals wrong.}
  \label{fig:synergy}
\end{figure}

\begin{itemize}
\item
  \cref{fig:synergy} plots correct against wrong on this subset. Selection-type baselines also answer some of these questions correctly. A greedy closed-book failure does not mean the model lacks the knowledge; other prompts and sampling can surface it. So we do not claim that only our method combines the two sources.
\item
  What differs is the error rate. Ours is correct about as often as the best baselines on this subset and wrong less than half as often. Correct minus wrong on the subset: Llama ours +1.4, every baseline between −5 and −12; Qwen ours −1.8, baselines between −10 and −20.
\item
  On wrong\_branch with Qwen the difference is also in correct answers: ours 9.9\%, baselines at most 1.5\% except the memory-first ones (Astute 5.8\%, GenRead 7.1\%).
\item
  Of our correct answers, 14\% (Qwen) and 23\% (Llama) fall in this subset.
\end{itemize}

\subsection{Other question types (GRACE-QT)}
\label{app:qt}

GRACE-QT dev, non-boolean questions (n = 376), C−W, protocol v3:

\begin{center}\footnotesize\setlength{\tabcolsep}{3.5pt}
\begin{tabular}{@{}lrrrr@{}}
\toprule
 & \multicolumn{2}{c}{Qwen} & \multicolumn{2}{c}{Llama} \\
\cmidrule(l{3pt}r{3pt}){2-3}\cmidrule(l{3pt}r{3pt}){4-5}
 & C$-$W & ours $-$ method {\scriptsize[95\% CI]} & C$-$W & ours $-$ method {\scriptsize[95\% CI]} \\
\midrule
\textbf{Ours} & 2.7 &  & \textbf{1.3} &  \\
plain reader & \textbf{4.0} & −1.3\,{\scriptsize[−5.1, +2.4]} & −1.6 & +2.9\,{\scriptsize[−2.9, +8.8]} \\
USC & 2.4 & +0.3\,{\scriptsize[−5.9, +6.6]} & −15.4 & +16.8\,{\scriptsize[+9.0, +24.5]} \\
Self-Route & −2.4 & +5.1\,{\scriptsize[−1.1, +11.2]} & −2.1 & +3.5\,{\scriptsize[−2.9, +9.8]} \\
InstructRAG-ICL & −12.0 & +14.6\,{\scriptsize[+6.9, +22.1]} & −6.9 & +8.2\,{\scriptsize[+1.1, +15.4]} \\
Astute RAG & −14.1 & +16.8\,{\scriptsize[+9.3, +24.2]} & −33.5 & +34.8\,{\scriptsize[+26.6, +42.8]} \\
GenRead & −38.0 & +40.7\,{\scriptsize[+32.2, +49.2]} & −37.5 & +38.8\,{\scriptsize[+30.9, +46.8]} \\
ours with the earlier selector (v2) & −8.5 & +11.2\,{\scriptsize[+5.1, +17.3]} & −5.9 & +7.2\,{\scriptsize[+2.7, +11.7]} \\
closed-book & −32.4 & +35.1\,{\scriptsize[+26.1, +43.9]} & −34.8 & +36.2\,{\scriptsize[+27.9, +44.4]} \\
\bottomrule
\end{tabular}\end{center}

\begin{itemize}
\item
  Here the model\textquotesingle s knowledge is poor (closed-book about −33) and the contexts are unperturbed. The right behaviour is to follow the context or abstain.
\item
  The earlier selector trusted knowledge and lost. v3 is statistically indistinguishable from the best baseline on both backbones (plain reader: −1.3 {[}−5.1, +2.4{]} with Qwen, +2.9 {[}−2.9, +8.8{]} with Llama) and significantly better than its own earlier selector. Where knowledge does not help, the system does no harm; it does not help either.
\item
  Boolean questions are excluded from this table: the reader and the plain-reader prompts ask for an entity, so models answer "Switzerland" to "Does absinthe come from Switzerland?". This is a format limitation of our prompts.
\end{itemize}

\subsection{Does the answer warn the reader?}
\label{app:signal}

Every score above judges the short answer. The two-voiced answer also carries a signal that a short answer cannot: it states what the context says, what the model knows, and whether the two conflict. We measure that signal on GRACE dev from outputs that already exist (no new model call). A record is \emph{flagged} when the two-voiced reader labels the relation between the sources as a conflict. A wrong answer is \emph{unwarned} when the short answer shown to the user is wrong and no conflict is flagged. The plain reader returns a short answer only, so each of its wrong answers is unwarned.

\begin{center}
\footnotesize\setlength{\tabcolsep}{2.6pt}
\begin{tabular}{@{}l rrrr rrrr@{}}
\toprule
& \multicolumn{4}{c}{Qwen} & \multicolumn{4}{c}{Llama} \\
\cmidrule(lr){2-5}\cmidrule(l){6-9}
\% of records & clean & broken & \makecell[r]{wrong\_\\branch} & competing & clean & broken & \makecell[r]{wrong\_\\branch} & competing \\
\midrule
conflict flagged & 12.7 & 11.0 & 76.9 & 38.8 & 14.2 & 12.6 & 45.6 & 30.5 \\
\addlinespace
\multicolumn{9}{@{}l}{wrong answer, no warning}\\
\quad plain reader & 6.4 & 6.7 & 69.2 & 13.0 & 7.4 & 12.0 & 26.4 & 11.4 \\
\quad ours & 6.4 & 7.4 & 11.1 & 7.6 & 7.1 & 10.4 & 14.1 & 7.0 \\
\quad\quad reduction & 0.0 & $-$0.7 & 58.1 & 5.4 & 0.3 & 1.6 & 12.3 & 4.4 \\
\quad\quad 95\% CI & \scriptsize[$-$1.6, 1.5] & \scriptsize[$-$2.5, 1.0] & \scriptsize[53.9, 62.1] & \scriptsize[3.1, 8.0] & \scriptsize[$-$1.9, 2.4] & \scriptsize[$-$1.0, 4.4] & \scriptsize[8.4, 16.0] & \scriptsize[1.8, 7.4] \\
\addlinespace
\multicolumn{9}{@{}l}{correct answer visible in the output}\\
\quad plain reader & 78.5 & 44.4 & 3.4 & 75.7 & 79.3 & 51.4 & 9.6 & 74.1 \\
\quad ours & 91.7 & 78.8 & 63.6 & 89.3 & 90.5 & 72.9 & 42.7 & 86.4 \\
\bottomrule
\end{tabular}
\end{center}

\begin{itemize}
\item
  \textbf{The wrong\_branch state.} The plain reader gives a wrong answer with no warning on 69.2\% of records (Qwen) and 26.4\% (Llama). Ours does so on 11.1\% and 14.1\%. About half of our remaining wrong answers with Qwen (49.7\%) and 38.3\% with Llama are flagged as a conflict in the answer text.
\item
  \textbf{The correct answer reaches the reader far more often.} On wrong\_branch contexts a gold name appears somewhere in our output on 63.6\% of records (Qwen) and 42.7\% (Llama), against 3.4\% and 9.6\% for the plain reader. The gain is also large on broken contexts (78.8 against 44.4; 72.9 against 51.4).
\item
  \textbf{Cost: false alarms.} A conflict is flagged on 12.7\% (Qwen) and 14.2\% (Llama) of clean records, where the context is right. On those records the shown answer is wrong on 15.1\% and 14.6\%.
\item
  \textbf{A flag is a reason to check, not a verdict.} When a conflict is flagged on a wrong\_branch context, the knowledge voice names a gold answer on 63.6\% (Qwen) and 43.8\% (Llama) of records.
\item
  \textbf{Scope of this measurement.} It counts what the output contains. It does not show that people read the warning or act on it; that requires a user study, which we have not run (\cref{sec:conclusion}).
\end{itemize}

\begin{figure}[!htbp]
  \centering
  \includegraphics[width=0.85\linewidth]{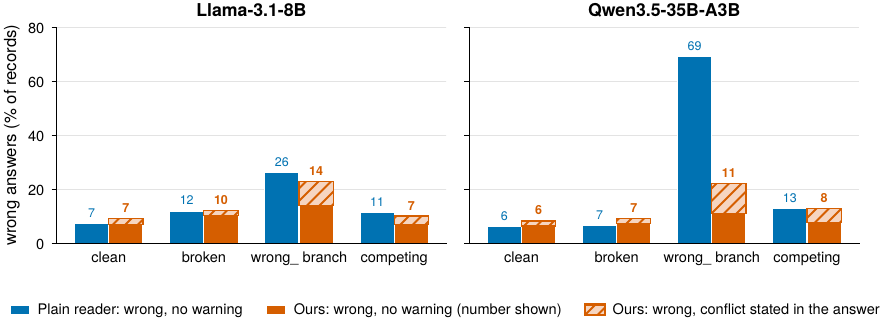}
  \caption{Wrong answers on GRACE dev, split by whether the answer shown to the user states a conflict between the context and the model\textquotesingle s knowledge. Solid bars are wrong answers that reach the user with no warning (number above each bar of ours); the hatched part is wrong answers whose text states the conflict. The plain reader returns a short answer, so all its wrong answers are unwarned.}
  \label{fig:signal}
\end{figure}

%% file: appendix/G_external.tex
\section{Preregistered external tests}
\label{app:external}

External benchmarks are used only for testing. For each, the system and every baseline were frozen, hypotheses and scoring were written to a hashed file, and the test was run once.

{\footnotesize\setlength{\tabcolsep}{3pt}
\begin{longtable}[]{@{}>{\raggedright\arraybackslash}p{0.09\linewidth}>{\raggedright\arraybackslash}p{0.372\linewidth}>{\raggedright\arraybackslash}p{0.239\linewidth}>{\raggedright\arraybackslash}p{0.244\linewidth}@{}}
\toprule\noalign{}
Lockbox & Data & System tested & Status \\
\midrule\noalign{}
\endhead
\bottomrule\noalign{}
\endlastfoot
1 & WebQSP test \citealp{yih2016webqsp}, CWQ test \citealp{talmor2018cwq} & first version (first selector, lenient matching rule) & mixed; scored with a rule later found lenient, so not used as evidence (\cref{app:prereg}) \\
2 & Mintaka test (\citealp{sen2022mintaka}; entity answers, 2,432 questions), QALD-10 test (\citealp{usbeck2024qald10}; 171) with Wikidata 1-hop context & v2 (earlier selector) & failed; reported in \cref{app:lockbox2} \\
3 & ClashEval (\citealp{wu2024clasheval}; 10,942 records: 1,247 original contexts, 9,695 perturbed, all perturbation levels); Housing Statute QA (\citealp{zheng2025legalretrieval}; 1,000 yes/no questions, BM25 retrieval within the state) & v2 as preregistered; v3 added by an amendment written before any output was read & \cref{sec:external} \\
\end{longtable}}

Preregistered hypotheses for lockbox 3: \textbf{H-v3}: v3 (cautious, π = 0.25) beats the plain reader; \textbf{H-v3b}: v3 is at least as good as the best baseline; \textbf{H-π}: the best π increases with the fraction of correct contexts.

Deviation, disclosed: after reading lockbox-3 scores we found that the preregistered scorer kept only the first segment of list-formatted answers (\cref{app:lessons}). We report the preregistered score and the corrected (whole-answer) score side by side and label the latter post hoc.

\subsection{Preregistration, frozen artefacts and deviations}
\label{app:prereg}

\textbf{Procedure.} For each external test, the system, every baseline command, the data files, the scorer and the hypotheses are written to a JSON file together with SHA-256 hashes before any model output on that test is produced. The test is then run once.

{\footnotesize\setlength{\tabcolsep}{3pt}
\begin{longtable}[]{@{}>{\raggedright\arraybackslash}p{0.169\linewidth}>{\raggedright\arraybackslash}p{0.494\linewidth}>{\raggedright\arraybackslash}p{0.297\linewidth}@{}}
\toprule\noalign{}
Lockbox & File (first 16 hex digits of its SHA-256) & Frozen at \\
\midrule\noalign{}
\endhead
\bottomrule\noalign{}
\endlastfoot
1 & \texttt{PREREGISTRATION\_LOCKBOX.json} (\texttt{5847cc03f7a51f05}) & 2026-10-06 \\
2 & \texttt{PREREGISTRATION\_LOCKBOX2\_SELECTION.json} (\texttt{22afe8ac578b90d5}), \texttt{PREREGISTRATION\_LOCKBOX2\_SYSTEM.json} (\texttt{4a9155d8dfb18bc4}) & 2026-10-06 (selection), 2026-10-07 (system) \\
3 & \texttt{PREREGISTRATION\_LOCKBOX3.json} (\texttt{c10d51bf9086bea1}) & 2026-10-08 01:28 \\
3, amendment 1 & \texttt{PREREGISTRATION\_LOCKBOX3\_AMENDMENT1.json} (\texttt{2c07407ff7150c2e}) & 2026-10-08 23:02 \\
\end{longtable}}

{\footnotesize\setlength{\tabcolsep}{3pt}
\begin{longtable}[]{@{}>{\raggedright\arraybackslash}p{0.447\linewidth}>{\raggedright\arraybackslash}p{0.528\linewidth}@{}}
\toprule\noalign{}
Frozen artefact & SHA-256 (first 16 hex digits) \\
\midrule\noalign{}
\endhead
\bottomrule\noalign{}
\endlastfoot
note policy, Qwen / Llama & \texttt{51e21f66c4754f5a} / \texttt{b96f56c4b6232bd4} \\
selector v2, Qwen / Llama & \texttt{53c8ccb97d90374d} / \texttt{cbf7f684c7b6fd68} \\
selector v3, Qwen / Llama & \texttt{60831c3e75732a59} / \texttt{c2121e95254bd208} \\
selector v4, Qwen / Llama & \texttt{3358115d6a4d9766} / \texttt{b9a0b3a7d5a4ecff} \\
\end{longtable}}

\textbf{Lockbox 1 (WebQSP and CWQ test).} The first version of the system (first selector, lenient matching rule of \cref{app:lessons}) was tested once on WebQSP and CWQ. Under that rule, H1 (cautious mode beats the plain reader in C−W) failed on WebQSP (−1.2 {[}−5.0, +2.4{]}; the context contains the answer for 87\% of questions) and held on CWQ (+6.8 {[}+4.2, +9.4{]}; 48\%). Because the matching rule was later found to be lenient, we do not use these numbers as evidence. The validation splits of both data sets had been used for early exploration (Supplementary Section~\ref{app:earlier}) and are marked as used.

\textbf{Amendment 1 of lockbox 3.} Written after the measurement flaw of \cref{app:lessons} was found on GRACE and lockbox 2, and before any lockbox-3 answer, score or output file was read; only build statistics and job progress had been inspected. It adds selector v3 (applied to the frozen notes, policy and reader outputs of the lockbox-3 run), adds scoring with the shared extractor and the ClashEval authors\textquotesingle{} comparison functions, keeps v2 with every original metric, and states H-v3, H-v3b and H-π.

\textbf{Deviations from the preregistration.}

\begin{enumerate}
\def\labelenumi{\arabic{enumi}.}
\item
  \emph{Whole-answer scoring is post hoc.} The preregistered scorer keeps the first \textquotesingle\textbar\textquotesingle-separated segment (\cref{app:lessons}). We report both scorers. For every method the two agree within 0.2 points on ClashEval with Llama; they differ for the plain reader (−40.2 whole-answer, −70.2 preregistered) and the closed-book arm (−38.3, −85.1), whose Llama answers are list-formatted.
\item
  \emph{Reader outputs that fail to parse are scored as abstentions and not re-run.} The two-voiced reader runs at temperature 0, so a failure is a deterministic property of the system. Counts: ClashEval 27 (Qwen) and 101 (Llama) of 10,942; Housing 40 (Qwen) and 30 (Llama) of 1,000.
\item
  \emph{Baseline failures.} TruthfulRAG\textquotesingle s official pipeline raises an exception when the model emits its intermediate graph as a list. After one re-run with the unmodified code, 88 of 10,942 ClashEval records and 68 of 1,000 Housing records still fail with Llama; they are scored as abstentions.
\item
  \emph{Baselines added after the lockbox-3 preregistration and before any output was read:} TruthfulRAG, CK-PLUG, and the ClashEval token-probability correction. SHIFT was planned and not run.
\item
  \emph{The ClashEval token-probability baseline with Llama} was first reported on a run with 1,992 failed rows (−46.3); the complete run gives −55.9, which is the number in \cref{sec:external}.
\end{enumerate}

\subsection{Task coverage of the trained selector: lockbox 2, selector v3 and ClashEval}
\label{app:lockbox2}

Every stage of the system except the selector is either frozen (the backbone, the encoder) or general by construction (the elicitation and reader prompts). The selector is the one component that learns, from labelled examples, which candidate to trust; it therefore covers the question types and context distributions it is trained on and no others. This subsection reports the two external tests where that coverage was missing, what failed, and how the selector was retrained for the task.

\paragraph{Lockbox 2: the selector trained on slot-filling questions only (v2).} The system frozen for lockbox 2 used selector v2, trained on \grace{} questions alone. Re-scored without re-running (protocol v2), its preregistered hypotheses failed:

\begin{center}\footnotesize\setlength{\tabcolsep}{3.5pt}
\begin{tabular}{@{}lrrrr@{}}
\toprule
C−W & Mintaka Qwen & Mintaka Llama & QALD-10 Qwen & QALD-10 Llama \\
\midrule
ours with selector v2, cautious & 21.8 & 6.7 & 43.9 & 36.8 \\
plain reader & 11.4 & 5.6 & \textbf{59.6} & 38.0 \\
GenRead & 45.4 & \textbf{37.4} & 38.6 & 31.0 \\
InstructRAG-ICL & \textbf{45.9} & 26.0 & 52.6 & \textbf{47.4} \\
USC & 36.4 & 23.1 & 56.1 & 43.9 \\
\bottomrule
\end{tabular}\end{center}

We decompose the system stage by stage on the same outputs (post hoc, no new model call; C−W under the scoring rule of the run, which differs from protocol v2 by a few points but not in ranking):

\begin{center}\footnotesize\setlength{\tabcolsep}{3.5pt}
\begin{tabular}{@{}lrrr@{}}
\toprule
Mintaka, Qwen (n = 2{,}432) & All & \makecell[r]{context has\\the answer (47\%)} & \makecell[r]{context\\lacks it} \\
\midrule
two-voiced reader, its own \texttt{final} & 44.2 & 66.9 & 24.0 \\
selector v2, cautious (registered) & 26.4 & 53.5 & 2.2 \\
selector v2, always answering & 23.9 & 58.4 & −6.8 \\
\quad same, without note-object candidates & \textbf{46.5} & \textbf{71.1} & 24.5 \\
GenRead & 45.5 & 62.1 & \textbf{30.7} \\
InstructRAG-ICL & 47.9 & 69.3 & 28.8 \\
\bottomrule
\end{tabular}\end{center}

\begin{enumerate}
\def\labelenumi{\arabic{enumi}.}
\item
  \emph{The knowledge reached the answer; the selector discarded it.} The reader\textquotesingle s own conclusion scores 44.2, level with GenRead (45.5) and InstructRAG-ICL (47.9). On questions whose context lacks the answer, where the model\textquotesingle s knowledge is the only source, the reader scores 24.0 and the registered system 2.2: the loss is made after the reader, at selection.
\item
  \emph{Cause: a candidate type that exists only for slot-filling questions.} Selector v2 also offered the object of each knowledge note as a candidate. \grace{} questions ask for one missing slot (``who advised X?''), so a note\textquotesingle s object is often the answer, and the selector learned to trust it. Mintaka asks comparison, counting, superlative and multi-hop questions, where a note\textquotesingle s object is an intermediate entity: the selector chose a note object on 493 Mintaka questions and was right on 14.6\% of them (it answered ``Mississippi River'' to a question about an author who piloted steamboats, and an actor\textquotesingle s name to a question asking for a film). Removing this one candidate type, with nothing retrained, lifts the system from 23.9 to 46.5, above GenRead.
\item
  \emph{A second cause: labels that rewarded long answers.} Its training labels came from a lenient matching rule (\cref{app:lessons}) under which an answer listing several entities counts as correct if one matches, so the selector preferred long, list-like candidates.
\item
  \emph{QALD-10 is too small to separate methods.} With 171 questions and the answer in 84\% of contexts, the system without note-object candidates (68.4 cautious, Qwen) is within the interval of the best baseline.
\end{enumerate}

\paragraph{One reader, configurations per deployment.} The system has two kinds of parts. The elicitation prompts, the voicing policy and the two-voiced reader are the same in every deployment and are never retrained. Three settings belong to the deployment: the prior $\pi$, the answer mode (cautious or always answering), and the selector, which learns from labelled examples of the task which candidate to trust. The tests differ in what these settings have to cover:

\begin{center}\footnotesize\setlength{\tabcolsep}{3pt}
\begin{tabular}{@{}>{\raggedright\arraybackslash}p{0.17\linewidth}>{\raggedright\arraybackslash}p{0.25\linewidth}>{\raggedright\arraybackslash}p{0.27\linewidth}>{\raggedright\arraybackslash}p{0.26\linewidth}@{}}
\toprule
Case & What differs & What the configuration must do & Result \\
\midrule
\grace{} (slot filling, graph facts) & one context in four right & selector trained on \grace{}; $\pi = 0.25$ & above every fixed method (\cref{sec:results}) \\
other question types (Mintaka, \grace{}-QT) & comparison, rank, qualifier, multi-hop; a note\textquotesingle s object is not the answer & candidates are answers the reader states; selector trained with these types (v3) & \grace{}-QT: v3 +11.2 (Qwen) and +7.2 (Llama) over v2, level with the best baseline \\
prose, mostly perturbed (ClashEval) & passages, not graph facts; $\pi \approx 0.11$ & low $\pi$; selector calibrated on passages & Qwen: ties the best baseline; Llama: calibration error 0.31, trails GenRead \\
model lacks the knowledge (Housing) & closed-book wrong on 60\% & high $\pi$: follow the context & rises with $\pi$ toward the plain reader (2.0 to 8.8 against 11.4, Qwen) \\
\bottomrule
\end{tabular}\end{center}

\textbf{Selector v3} is the configuration of the selector for a deployment that mixes question types. It keeps the architecture and the features of v2 and changes what it is trained on and what it may choose: (i) its training questions are the \grace{} pool plus \grace{}-QT train, 500 LC-QuAD train questions of types absent from the slot-filling pool (rank, qualified statement, boolean, string; \cref{app:grace}); (ii) its candidates are the answers the reader states and the plain reader\textquotesingle s answer, so every candidate is something a reader concluded rather than an entity mentioned along the way; (iii) every candidate is shortened to a short answer and labelled with the strict matching rule of \cref{sec:grace}. The prior enters through \cref{eq:shift} rather than as a learned input, so its direction does not depend on the training distribution (\cref{app:method}). v3 is evaluated on held-out \grace{}-QT development questions and on lockbox 3.

\begin{itemize}
\item
  \emph{On question types beyond slot filling (\grace{}-QT dev, \cref{app:qt}).} v3 scores 2.7 (Qwen) and 1.3 (Llama) against −8.5 and −5.9 for v2, gains of 11.2 {[}5.1, 17.3{]} and 7.2 {[}2.7, 11.7{]}, and is statistically level with the best baseline on both backbones. Where the model\textquotesingle s knowledge is weak (closed-book about −33), the selector follows the context or abstains instead of trusting notes.
\item
  \emph{On ClashEval (Supplementary Section~\ref{app:clasheval}).} v3 scores −22.5 (Qwen) and −29.1 (Llama) against −28.3 and −37.7 for v2 applied to the same frozen reader outputs, gains of 5.8 and 8.6 points.
\item
  \emph{Which change matters (Supplementary Section~\ref{app:selector}).} Short candidates (4.1 points) and strict labels (1.4) matter with Llama, whose answers are long and list-formatted, and are near zero with Qwen. Removing note-object candidates and adding \grace{}-QT train do not change the \grace{} score significantly; their effect is on question types \grace{} does not contain.
\end{itemize}

\paragraph{The reader alone.} Scored on its own conclusion (\texttt{final}), with no selector, the two-voiced reader carries over to new question types wherever the backbone can follow its output format. C−W on the lockbox-2 outputs (scoring rule of the run; in parentheses, questions whose context lacks the answer):

\begin{center}\footnotesize\setlength{\tabcolsep}{3.5pt}
\begin{tabular}{@{}lllll@{}}
\toprule
 & Mintaka Qwen & Mintaka Llama & QALD-10 Qwen & QALD-10 Llama \\
\midrule
two-voiced reader, \texttt{final} & 44.2 (24.0) & −1.2 (−28.2) & 58.5 (−14.3) & 37.4 (−53.6) \\
plain reader & 11.3 (−6.3) & 5.1 (−13.4) & 59.6 (−7.1) & 36.8 (−21.4) \\
best baseline & 47.9 (28.8) & 35.2 (21.1) & 69.6 (−14.3) & 57.3 (−32.1) \\
\bottomrule
\end{tabular}\end{center}

\begin{itemize}
\item
  \emph{With Qwen on Mintaka} the reader is 32.9 points above the plain reader {[}29.3, 36.7{]} and level with the best baseline (InstructRAG-ICL); on questions whose context lacks the answer it moves from −6.3 to 24.0, which is the model\textquotesingle s knowledge reaching the answer. The same holds on \grace{}: on wrong contexts a correct answer appears in the reader\textquotesingle s output on 63.6\% (Qwen) and 42.7\% (Llama) of records, against 3.4\% and 9.6\% for the plain reader, and about half of the remaining wrong answers (Qwen; 38\% with Llama) carry a conflict warning (\cref{app:signal}).
\item
  \emph{On QALD-10} the context contains the answer for 84\% of questions, so there is little for knowledge to add; the reader is level with the plain reader (−1.2 {[}−14.0, +12.3{]}, Qwen; +0.6, Llama), and with 171 questions no method is separated from the best baseline (USC) by its interval.
\item
  \emph{With Llama-3.1-8B on Mintaka} the reader falls below the plain reader: on comparison and multi-hop questions the 8B model often breaks the two-voiced output format (printing a raw triple such as \texttt{Chris Evans -{}-is older than-{}->}), so the limit there is the backbone\textquotesingle s instruction following, not the decision rule.
\end{itemize}

\paragraph{ClashEval with Llama: coverage of the context distribution.} v3 still trails GenRead on ClashEval with Llama (−29.1 against −9.5; \cref{sec:external}). The cause is again the training distribution, now of the contexts rather than the questions. \cref{prop:shift} gives the exact posterior only if the selector is calibrated and the deployment differs from training in $\pi$ alone. ClashEval differs in more than $\pi$: its documents are prose passages rather than knowledge-graph facts, its perturbations alter numbers, dates and names by graded amounts, and 89\% of records are perturbed. The selector, trained on \grace{} contexts, states a mean probability of 0.49 for answers that are right 18\% of the time; setting $\pi$ to the true share (0.114) lowers its calibration error only from 0.35 to 0.31, against 0.03 to 0.05 on \grace{}. The error is still lowest at the true prior (\cref{fig:theory}c), so the shift moves in the right direction, but it starts from miscalibrated probabilities, so the system commits where a calibrated selector would abstain. GenRead ignores the document and abstains on 62\% of records, which this mixture rewards; on unmodified documents it is 53.9 points below ours. With Qwen the same selector ties the strongest baseline (Astute RAG). Closing the Llama gap requires the step that closed the question-type gap: training or recalibrating the selector on contexts of the target kind, which uses labelled in-domain data and was outside the preregistered protocol.

\paragraph{What carries over and what does not.} The elicited notes and the two-voiced reader carry over without retraining: with a backbone that follows the format, the reader\textquotesingle s conclusion matches the strongest baselines on new question types, and the conflict warning and the correct answer surfaced on wrong contexts are properties of its output, not of the selector. The direction of the prior shift carries over by construction. What must be configured per deployment is the prior and the selector\textquotesingle s calibration, which is learned and holds on the question types and context kinds it is trained on.

%% file: appendix/H_cost.tex
\section{Latency and cost per stage}
\label{app:cost}
\begin{figure}[!htbp]
  \centering
  \includegraphics[width=0.95\linewidth]{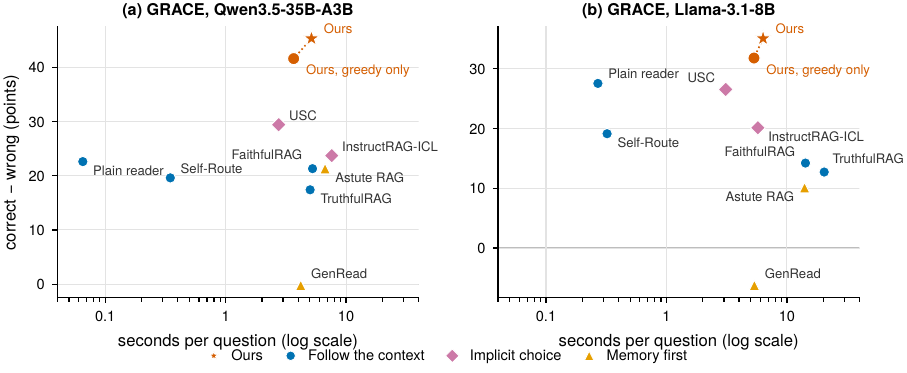}
  \caption{Accuracy against wall-clock time per question (\grace{} development data; time measured on 40 questions, one request at a time, idle GPU). Ours (star) is the most accurate method on both backbones at a cost comparable to the other methods that call the model several times; dropping the sampled elicitation runs (circle) trades 3 to 4 points for 1 to 1.5\,s.}
  \label{fig:latency}
\end{figure}

A counting proxy records, for every request of every method, the number of calls, prompt tokens, generated tokens and wall-clock time from the server\textquotesingle s own accounting. Each backbone ran alone on an otherwise idle GPU (sglang, one request at a time) on the same 40 GRACE dev questions, on 2026-10-22. Values are means per question. For a stage of our system, the time is the sum of its calls; for our total, it is the critical path when the calls of one question that do not depend on each other run concurrently (the plain reader and the elicitation runs; the self-verification calls). CAD, AdaCAD, CK-PLUG and TCR run in their own decoding code, not through a server, and are not timed; contrastive decoding needs two forward passes per generated token.

\begin{center}\footnotesize\setlength{\tabcolsep}{3.5pt}
\begin{tabular}{@{}>{\raggedright\arraybackslash}p{0.255\linewidth}rrrrrrrr@{}}
\toprule
 & \multicolumn{4}{c}{Qwen} & \multicolumn{4}{c}{Llama} \\
\cmidrule(l{3pt}r{3pt}){2-5}\cmidrule(l{3pt}r{3pt}){6-9}
Stage or method & calls & prompt tok & gen tok & s/question & calls & prompt tok & gen tok & s/question \\
\midrule
plain reader & 1.0 & 296 & 5 & 0.07 & 1.0 & 306 & 7 & 0.27 \\
elicitation, greedy run & 1.0 & 304 & 342 & 2.38 & 1.0 & 323 & 162 & 3.04 \\
elicitation, five sampled runs (one request) & 1.0 & 304 & 1,638 & 3.86 & 1.0 & 315 & 696 & 4.05 \\
self-verification of notes & 7.7 & 649 & 8 & 0.24 & 7.3 & 738 & 7 & 0.27 \\
two-voiced reader & 1.0 & 577 & 134 & 0.95 & 1.0 & 586 & 107 & 1.95 \\
extractor for long candidates & 0.8 & 90 & 3 & 0.29 & 0.8 & 94 & 3 & 0.31 \\
\textbf{ours, total} (critical path) & 12.5 & 2,221 & 2,129 & 5.16 & 12.1 & 2,363 & 983 & 6.34 \\
ours, greedy elicitation only & 11.5 & 1,917 & 492 & 3.67 & 11.1 & 2,047 & 286 & 5.33 \\
Astute RAG & 2.0 & 424 & 968 & 6.69 & 2.0 & 708 & 786 & 14.01 \\
FaithfulRAG & 4.0 & 1,626 & 748 & 5.26 & 4.0 & 1,728 & 789 & 14.26 \\
TruthfulRAG & 8.0 & 9,348 & 663 & 5.03 & 14.5 & 12,176 & 1,097 & 20.41 \\
InstructRAG-ICL & 1.0 & 1,110 & 477 & 7.60 & 1.0 & 681 & 235 & 5.74 \\
USC & 2.0 & 255 & 827 & 2.76 & 2.0 & 303 & 476 & 3.10 \\
GenRead & 2.0 & 381 & 302 & 4.20 & 2.0 & 392 & 264 & 5.37 \\
Self-Route & 1.1 & 165 & 47 & 0.35 & 1.0 & 171 & 17 & 0.32 \\
\bottomrule
\end{tabular}\end{center}

\begin{itemize}
\item
  Our system takes 5.2 s (Qwen) and 6.3 s (Llama) per question, against 0.07 s and 0.27 s for the plain reader, and is within the range of the multi-call baselines: faster than Astute RAG, FaithfulRAG and TruthfulRAG with Llama (14 to 20 s), and between USC (2.8 s) and InstructRAG-ICL (7.6 s) with Qwen.
\item
  The five sampled elicitation runs are the largest single cost (3.9 to 4.1 s; 77\% of Qwen\textquotesingle s generated tokens). Without them the system takes 3.7 s and 5.3 s and scores 41.6 and 31.7, 3.7 and 3.3 points below the full system (\cref{sec:envelope}) and above every baseline.
\item
  The self-verification calls generate one token each and run in parallel, so 7 to 8 calls cost about 0.25 s.
\end{itemize}

%% file: appendix/I_lessons.tex
\section{Measurement lessons}
\label{app:lessons}

Three scoring flaws changed our conclusions during this work. We report them because they are easy to reproduce in any open-form RAG evaluation.

\textbf{A learned component exploited a lenient metric.}

Our first matching rule counted an answer correct if a gold name was a substring of it. The selector\textquotesingle s training labels used the same rule. The selector learned to prefer long, sentence-like candidates that list several entities. Under a strict rule the system\textquotesingle s GRACE score fell from 48.9 to 19.2 (Qwen) while baselines moved by a few points; 435 of our "correct" answers were longer than six words, against 32 for the plain reader. The fix has two parts: every method\textquotesingle s long answers go through one shared extractor before strict matching, and training labels use the strict rule.

\emph{Lesson:} a matching rule used to generate training labels must be at least as strict as the evaluation rule, and answer-length distributions should be compared across methods.

\textbf{Truncating list-formatted answers.}

Our plain-reader prompt asks for answer names "most likely first, separated by \textquotesingle{} \textbar{} \textquotesingle", and scorers kept the first segment. Llama-3.1-8B often writes "entity \textbar{} value" ("Piqray \textbar{} 300 MG") for value questions: 95.6\% of its answers on ClashEval and 77.6\% on GRACE dev contain the separator (Qwen: 1.2\% and 17.1\%). The raw Llama baseline was scored at 14.0 on GRACE dev under first-segment scoring and 27.5 under whole-answer scoring.

The same truncation existed inside our system: the plain reader\textquotesingle s candidate became the subject entity, so for value questions no candidate contained the answer. This explains why high π did not recover the original-context score with Llama on ClashEval. Selector v4 (Supplementary Section~\ref{app:v4}) extracts the plain reader\textquotesingle s candidate from the whole answer. v4 was developed on GRACE data only, but the defect was first observed on ClashEval, so ClashEval is not evidence for v4.

\textbf{Arm naming in the scoring call.}

Two raw output files carried no method field, and one overwrote the other in a scoring call. The affected numbers were never reported; normalising arm files before scoring removes the problem.

\textbf{Benchmark artefacts.}

In GRACE, one entity ("George Washington") is the injected wrong answer in 101 of 675 wrong\_branch records (15\%) and in the matching competing records. The note policy and the selector are trained on GRACE-derived data and could learn this cue. With Qwen our system scores the same on these records and on the others (17.8 and 17.9). With Llama it scores higher on them (7.9 against −6.3), as does a baseline that is not trained on GRACE (USC: −14.9 against −39.2). With all 202 records removed, the overall score moves from 45.3 to 46.1 (Qwen) and from 35.0 to 35.4 (Llama), and the margin over the plain reader from 22.7 to 20.3 and from 7.5 to 7.0 (\cref{app:cues}).

%% file: supplement/S1_selector.tex
\section{Inside the selector}
\label{supp:selector}

\label{app:selector}

\textbf{Selector-internal ablations.} The selector is retrained with one design choice changed and everything else as in the frozen recipe; no reader call is repeated. The row "full" reproduces the frozen selector\textquotesingle s output exactly (0 differing rows of 10,800 and 10,772). C−W at π = 0.25, cautious mode, protocol v3; Δ is the variant minus full with a 95\% paired question-clustered interval. GRACE-QT is the 376 non-boolean dev questions.

\begin{center}\footnotesize\setlength{\tabcolsep}{3pt}
\begin{tabular}{@{}lrrrr@{}}
\toprule
Variant & Qwen GRACE & Δ {[}CI{]} & Qwen GRACE-QT & Δ {[}CI{]} \\
\midrule
\textbf{full (frozen)} & \textbf{45.3} &  & 2.7 &  \\
no note-derived features & 39.8 & −5.5\,{\scriptsize[−7.5, −3.4]} & −4.3 & −6.9\,{\scriptsize[−11.2, −2.7]} \\
verbatim candidates (no shortening) & 44.9 & −0.4\,{\scriptsize[−1.7, +0.8]} & 2.4 & −0.3\,{\scriptsize[−2.7, +2.1]} \\
no cosine to sources or question & 44.5 & −0.8\,{\scriptsize[−2.3, +0.7]} & 1.1 & −1.6\,{\scriptsize[−5.3, +2.1]} \\
lenient training labels & 45.3 & 0.0\,{\scriptsize[−1.0, +1.1]} & 2.4 & −0.3\,{\scriptsize[−2.4, +1.9]} \\
note confidence fixed at 0.5 & 43.8 & −1.5\,{\scriptsize[−3.0, 0.0]} & 1.1 & −1.6\,{\scriptsize[−4.8, +1.3]} \\
note support fixed at 1 & 45.2 & −0.1\,{\scriptsize[−1.1, +0.8]} & 1.6 & −1.1\,{\scriptsize[−2.9, +0.5]} \\
no reader-stated relation & 45.3 & 0.0\,{\scriptsize[−1.0, +1.1]} & 4.8 & +2.1\,{\scriptsize[0.0, +4.5]} \\
no plain-reader answer & 44.7 & −0.6\,{\scriptsize[−1.9, +0.6]} & 4.0 & +1.3\,{\scriptsize[−1.9, +4.8]} \\
note objects added as candidates & 46.8 & +1.5\,{\scriptsize[0.0, +2.7]} & 2.1 & −0.5\,{\scriptsize[−3.7, +2.9]} \\
trained on the GRACE pool only & 45.8 & +0.5\,{\scriptsize[−0.5, +1.4]} & 2.1 & −0.5\,{\scriptsize[−3.2, +2.4]} \\
\bottomrule
\end{tabular}\end{center}

\begin{center}\footnotesize\setlength{\tabcolsep}{3.5pt}
\begin{tabular}{@{}lrrrr@{}}
\toprule
Variant & Llama GRACE & Δ {[}CI{]} & Llama GRACE-QT & Δ {[}CI{]} \\
\midrule
\textbf{full (frozen)} & \textbf{35.0} &  & 1.6 &  \\
no note-derived features & 30.8 & −4.2\,{\scriptsize[−6.2, −2.3]} & −1.3 & −2.9\,{\scriptsize[−7.2, +1.3]} \\
verbatim candidates (no shortening) & 30.9 & −4.1\,{\scriptsize[−6.0, −2.0]} & −1.3 & −2.9\,{\scriptsize[−7.2, +1.1]} \\
no cosine to sources or question & 32.2 & −2.8\,{\scriptsize[−4.7, −0.9]} & −0.8 & −2.4\,{\scriptsize[−6.1, +1.3]} \\
lenient training labels & 33.6 & −1.4\,{\scriptsize[−2.6, −0.2]} & 0.0 & −1.6\,{\scriptsize[−4.3, +1.1]} \\
note confidence fixed at 0.5 & 33.8 & −1.2\,{\scriptsize[−3.0, +0.6]} & −0.5 & −2.1\,{\scriptsize[−5.9, +1.6]} \\
note support fixed at 1 & 34.9 & −0.1\,{\scriptsize[−1.3, +1.0]} & 1.9 & +0.3\,{\scriptsize[−2.4, +2.9]} \\
no reader-stated relation & 35.1 & +0.1\,{\scriptsize[−1.0, +1.1]} & 2.9 & +1.3\,{\scriptsize[−1.3, +4.3]} \\
no plain-reader answer & 34.5 & −0.5\,{\scriptsize[−1.7, +0.7]} & −0.8 & −2.4\,{\scriptsize[−5.3, +0.8]} \\
note objects added as candidates & 36.3 & +1.3\,{\scriptsize[−0.3, +3.0]} & 1.3 & −0.3\,{\scriptsize[−3.5, +2.9]} \\
trained on the GRACE pool only & 36.2 & +1.2\,{\scriptsize[+0.1, +2.4]} & −0.8 & −2.4\,{\scriptsize[−5.9, +1.1]} \\
\bottomrule
\end{tabular}\end{center}

The Llama GRACE-QT score of the full system is 1.6 in this run and 1.3 in \cref{app:v4}; one record differs because the extractor returned a different string when called again.

Reading:

\begin{itemize}
\item
  \textbf{Note-derived features are the only component that matters on both backbones} (−5.5 and −4.2 on GRACE; −6.9 on GRACE-QT with Qwen). The selector needs to know which notes support a candidate.
\item
  \textbf{The two changes from selector v2 to v3 matter for Llama only.} Short candidates are worth 4.1 points and strict labels 1.4 points with Llama; both are near zero with Qwen. Llama writes longer, list-formatted answers (Appendix~\ref{app:lessons}), so verbatim candidates contain several entities.
\item
  \textbf{Calibrated confidence and support act through the reader, not through the selector.} Fixing them inside the selector costs 1.2 to 1.5 points (confidence; intervals touch zero) and 0.1 points (support). Removing confidence from the wording of the notes costs 7.1 to 9.3 points (Section~\ref{sec:envelope}).
\item
  \textbf{The reader-stated relation is not needed} (0.0 and +0.1 on GRACE). The plain-reader answer does not change the score at π = 0.25; it raises the clean score at π = 0.99 (73.2 against 69.0 with Qwen; 67.6 against 66.5 with Llama).
\item
  \textbf{Two choices of the frozen system are not supported by this ablation.} Adding note objects as candidates gains 1.5 {[}0.0, +2.7{]} (Qwen) and 1.3 {[}−0.3, +3.0{]} (Llama) on GRACE and changes GRACE-QT by −0.5 and −0.3 (intervals contain zero). Training on the GRACE pool alone gains 1.2 {[}+0.1, +2.4{]} on GRACE with Llama, and the GRACE-QT benefit of adding GRACE-QT train (2.4 points with Llama) has an interval that contains zero. Both choices target question types that GRACE does not contain, where a note\textquotesingle s object is often an intermediate entity (Appendix~\ref{app:lockbox2}); their benefit is not visible on GRACE dev, and on GRACE-QT dev its interval contains zero. They were fixed before protocol v3 existed, and the frozen system was not changed afterwards.
\end{itemize}

%% file: supplement/S2_where.tex
\section{Where the method helps and where it does not}
\label{supp:where}

\label{app:where}

The results across data sets follow one pattern.

{\footnotesize\setlength{\tabcolsep}{3pt}
\begin{longtable}[]{@{}>{\raggedright\arraybackslash}p{0.189\linewidth}>{\raggedright\arraybackslash}p{0.217\linewidth}>{\raggedright\arraybackslash}p{0.159\linewidth}>{\raggedright\arraybackslash}p{0.380\linewidth}@{}}
\toprule\noalign{}
Setting & Model knowledge (closed-book) & Context & Ours against the best alternative \\
\midrule\noalign{}
\endhead
\bottomrule\noalign{}
\endlastfoot
GRACE broken / wrong\_branch / competing & useful (about 40\% correct) & missing, wrong or contradictory & large gain \\
ClashEval names, years, locations, perturbed & useful & wrong & gain \\
GRACE clean; ClashEval original & useful & right & small cost at low π; parity at high π (Qwen) \\
GRACE-QT (rank, qualified, string) & poor (C−W about −33) & right, unperturbed & parity \\
Housing statutes & harmful (60\% wrong) & partly retrieved, right & Qwen: below the plain reader at the registered π, near it at high π. Llama: the frozen reader does not answer yes/no; ties the plain reader, 15--23 points below free-text methods \\
ClashEval drug dosages & poor (closed-book wrong on 56--58\% of answered questions) & wrong & reduced error rate with Qwen, no low-risk operating point \\
\end{longtable}}

\textbf{The method helps when the model has knowledge and the context is unreliable.} When the model does not know, there is nothing to merge; the correct setting is high π, which returns the system to context-following behaviour at a small cost.

\textbf{What the baselines\textquotesingle{} failures have in common.} Three of the four families fix the direction in which a conflict is resolved before they see the question. Follow-the-context methods and contrastive decoding resolve it toward the context, and collapse when the context is wrong; GenRead resolves it toward memory, and its score does not change when the context is right. The implicit-choice methods leave the direction to the model and follow a wrong context almost as often (−36 to −59 on wrong\_branch contexts). Section~\ref{sec:diagnosis} shows that a measure of how far the two sources diverge cannot repair this, because it does not indicate which source is right. Our system decides per question from two separate estimates, the reliability of the context (π) and the confidence of each note, and it shows both sources when it cannot decide.

The table above translates into deployment conditions. The method needs one precondition: the model already knows something about the domain. Given that, it is meant for retrieval pipelines whose context cannot be assumed correct.

{\footnotesize\setlength{\tabcolsep}{3pt}
\begin{longtable}[]{@{}>{\raggedright\arraybackslash}p{0.282\linewidth}>{\raggedright\arraybackslash}p{0.268\linewidth}>{\raggedright\arraybackslash}p{0.410\linewidth}@{}}
\toprule\noalign{}
Situation in a deployed system & What goes wrong today & Evidence in this paper \\
\midrule\noalign{}
\endhead
\bottomrule\noalign{}
\endlastfoot
Knowledge graph or index with missing facts (incomplete extraction, an entity not yet linked) & the reader abstains or answers from a partial path although the model knows the fact & GRACE broken: +6.9 (Qwen) and +5.9 (Llama) over the plain reader; the correct answer is visible in the output on 78.8\% and 72.9\% of records against 44.4\% and 51.4\% \\
Retrieved fact is wrong (extraction error, outdated or edited document, poisoned source) & the reader repeats the wrong fact with no sign of doubt & GRACE wrong\_branch: unwarned wrong answers fall from 69.2\% to 11.1\% (Qwen) and from 26.4\% to 14.1\% (Llama); ClashEval perturbed documents: +35.6 (Qwen) and +16.4 (Llama) over the plain reader \\
Sources disagree with each other (two documents, two graph branches) & the reader picks one silently & GRACE competing: +8.1 (Qwen) and +7.1 (Llama); the conflict is stated on 38.8\% and 30.5\% of records \\
Mixed traffic, reliability of the context unknown per query & a context-following system and a memory-first system each fail on one side & one system, one parameter: the total is flat for π between 0.1 and 0.5 on GRACE (Section~\ref{sec:envelope}), and π = 0.99 returns to context-following behaviour \\
\end{longtable}}

\textbf{Four concrete tasks inside this problem.} We group the GRACE dev questions by the Wikidata relation that carries the answer (a post hoc grouping; 337 of the 675 questions fall into these four families) and show, for each family, one real question and the difference between our system and the plain reader (C−W points, {[}95\% CI{]}).

{\footnotesize\setlength{\tabcolsep}{3pt}
\begin{longtable}[]{@{}>{\raggedright\arraybackslash}p{0.185\linewidth}>{\raggedright\arraybackslash}p{0.182\linewidth}>{\raggedright\arraybackslash}p{0.090\linewidth}>{\raggedright\arraybackslash}p{0.222\linewidth}>{\raggedright\arraybackslash}p{0.119\linewidth}>{\raggedright\arraybackslash}p{0.116\linewidth}@{}}
\toprule\noalign{}
Task & A real question, wrong fact in the context & Plain reader & Ours & Wrong fact in context: ours − plain reader (Qwen / Llama) & Missing fact: ours − plain reader (Qwen / Llama) \\
\midrule\noalign{}
\endhead
\bottomrule\noalign{}
\endlastfoot
\textbf{Biographical lookup} (father, education, award, adviser; 139 questions) & "What award did John Steinbeck receive?" The context says \emph{Alabama Hall of Fame}. & "Alabama Hall of Fame" & "According to the provided information, \ldots{} the Alabama Hall of Fame. As far as I know, John Steinbeck received the Nobel Prize in Literature" & +77.0 {[}62.6, 92.1{]} / +14.4 {[}3.6, 25.2{]} & +8.6 {[}0.0, 17.3{]} / +8.6 {[}−2.9, 20.1{]} \\
\textbf{Geographic and administrative facts} (capital, continent, located in, borders; 71) & "Which state of Australia has Melbourne as its capital?" The context says \emph{Western Australia}. & "Western Australia" & "\ldots{} As far as I know, this is incorrect; Melbourne is the capital of the state of Victoria" & +76.1 {[}53.5, 98.6{]} / +21.1 {[}5.6, 36.6{]} & +21.1 {[}9.9, 32.4{]} / +4.2 {[}−8.5, 18.3{]} \\
\textbf{Attribution of works and discoveries} (author, composer, inventor, developer; 71) & "Who is the writer of The Double Helix?" The context says \emph{Vladimir Lenin}. & "Vladimir Lenin" & "\ldots{} As far as I am confident, James Watson is the actual author of The Double Helix" & +95.8 {[}74.6, 116.9{]} / +21.1 {[}5.6, 36.6{]} & +5.6 {[}−7.0, 16.9{]} / −1.4 {[}−18.3, 15.5{]} \\
\textbf{Organisations: ownership, membership, leadership} (owner, founder, member of, chief executive; 56) & "Who is the business owner of Barbie the doll maker?" The context says \emph{United Parcel Service}. & "United Parcel Service" & "\ldots{} As far as I am confident, Mattel is the actual owner of Barbie, while United Parcel Service is a logistics company" & +87.5 {[}67.9, 107.1{]} / +10.7 {[}−5.4, 26.8{]} & −3.6 {[}−17.9, 8.9{]} / +17.9 {[}1.8, 35.7{]} \\
\end{longtable}}

\begin{itemize}
\item
  \textbf{A wrong fact in the context is where every task gains}, with Qwen by 76 to 96 points in each family. With Llama the gain is 11 to 21 points and its interval excludes zero in three of the four families.
\item
  \textbf{A missing fact helps most for geography with Qwen} (+21.1): borders, containment and capitals are facts the model knows well and graphs often lack.
\item
  \textbf{Facts that change over time need care.} In the organisation family (chief executive, chairman, head of government, owner) our Qwen system loses 16.1 points on clean contexts ({[}−32.1, −3.6{]}), the largest clean-context cost of the four; the people family loses 10.1 ({[}−18.7, −1.4{]}). Here the model\textquotesingle s knowledge can be out of date while the graph is current, and the system sometimes prefers the stale memory. For such questions a higher π is the right setting.
\item
  The families are small and the intervals wide; the table shows where the effect is, not a ranking of tasks.
\end{itemize}

\textbf{Deployed problems with the same structure.} Reports from several application areas describe the two conditions of this paper, a reader that has knowledge of its own and a context that cannot be assumed correct. We relate our results to them; we have experiments only for the first two.

{\footnotesize\setlength{\tabcolsep}{3pt}
\begin{longtable}[]{@{}>{\raggedright\arraybackslash}p{0.173\linewidth}>{\raggedright\arraybackslash}p{0.442\linewidth}>{\raggedright\arraybackslash}p{0.345\linewidth}@{}}
\toprule\noalign{}
Problem & What the literature reports & Relation to our results \\
\midrule\noalign{}
\endhead
\bottomrule\noalign{}
\endlastfoot
\textbf{Answers on search engines} (AI overviews, answer engines) & With a commercial search engine about 70\% of retrieved passages lack the answer, and in conflicts the model is right 47.4\% of the time (Astute RAG; Wang et al., ACL 2025). 57.2\% of real questions retrieve contradicting web paragraphs \citealp{yuan2026confrag}. A deployed system repeated a joke and a satire page as advice (Section~\ref{sec:intro}). & Directly tested: ClashEval is built from web documents with perturbed answers (Section~\ref{sec:external}); misinformation, insufficient context and conflicting sources are the wrong\_branch, broken and competing states of GRACE \\
\textbf{Assistants and question answering over knowledge graphs} & Knowledge graphs are incomplete and contain wrong statements: over 70\% of people in Freebase had no place of birth \citealp{west2014kbcompletion}; Wikidata has removed 76.5 million statements and deprecated 10 million \citealp{shenoy2022wikidataquality}. On CRAG, with mock web and knowledge-graph search, industry RAG systems answer 63\% of questions without hallucination \citealp{yang2024crag}. & Directly tested: GRACE is this setting (Sections 4 and 7) \\
\textbf{GraphRAG over graphs extracted by a language model} & Extraction loses triples and context (Huang et al., IJCAI 2025). & \cref{app:mcq} measures deployed GraphRAG-family systems on GRACE \\
\textbf{Corpora and agent memories that others can write to} & Five injected passages per question give about 90\% attack success \citealp{zou2025poisonedrag}; injecting records into an agent\textquotesingle s memory through queries alone succeeds for more than 95\% of injections (Dong et al., NeurIPS 2025). & Not a defence against a targeted attack; when the attacked fact is known to the model the conflict is stated (76.9\% of wrong\_branch records with Qwen) \\
\end{longtable}}

\textbf{Problems with the same two sources and the opposite direction.} In these, the context is usually the newer and authoritative side and the model\textquotesingle s memory is the stale one. Our evidence does not support using the model\textquotesingle s knowledge to correct the context there; π should be 0.9 or higher, and what remains useful is the format, which labels a remembered statement as memory.

\begin{itemize}
\item
  \emph{Agent and assistant memory.} A later observation can invalidate an earlier memory; LongMemEval (Wu et al., ICLR 2025) includes such knowledge updates among five memory abilities and finds that commercial assistants and long-context models lose about 30\% accuracy over sustained interactions.
\item
  \emph{Statutes, regulations, internal policies, prices, news.} Our Housing Statute QA result shows the failure: the model\textquotesingle s own answers are wrong on 60\% of questions and merging does not help (Section~\ref{sec:external}). Retrieval-based legal research tools still give wrong or unsupported answers on 17\% to 33\% of queries \citealp{magesh2025hallucinationfree}, so the need for caution is real.
\item
  \emph{Clinical guidance.} Guidelines change over time; on the ClashEval drug-dosage questions \citealp{wu2024clasheval} the model\textquotesingle s knowledge is poor and no operating point reaches a low error rate (Section~\ref{sec:external}).
\end{itemize}

\textbf{When not to use it.}

\begin{itemize}
\item
  \emph{The model has no knowledge of the domain, or its knowledge is wrong}: private or organisation-specific documents, recent events, statutes. There is nothing to merge; on Housing Statute QA the model\textquotesingle s own answers are wrong on 60\% of questions and our system does not beat the plain reader (Section~\ref{sec:external}). Use a context-following reader, or set π near 1.
\item
  \emph{The context is curated and trusted.} At π = 0.25 the method costs about 6 points on clean contexts with Qwen (64.3 against 70.5); at high π the cost disappears, and so does the benefit.
\item
  \emph{A wrong answer is costly and must be rare.} The selector\textquotesingle s probability does not provide a low-risk operating point once some contexts are perturbed (below).
\item
  \emph{The call budget is tight.} The system makes about a dozen short model calls per question (Section~\ref{sec:method}).
\end{itemize}

\textbf{Choosing π without labels.} π is the operator\textquotesingle s estimate of how often the context is right. Three settings cover the cases above: 0.25 when the context is unverified (the trained setting), 0.5 when it is mostly reliable, and 0.9 or more when it is curated. Estimating π from unlabelled traffic is future work (Section~\ref{sec:conclusion}).

\textbf{High-stakes use.} We tested whether the selector\textquotesingle s probability supports operation at a low error rate on ClashEval. It does not once any contexts are perturbed: with Qwen, coverage at risk ≤ 5\% is 93\% when all contexts are original and 16\% when 10\% are perturbed. In mixtures, our Qwen system lowers the error rate among answered questions relative to the plain reader and TCR (f = 0.75: 19\% against 25\% and 29\%) at slightly lower coverage. With Llama the error rate among answered questions is not lower than the plain reader\textquotesingle s (Section~\ref{sec:external}). We therefore claim \emph{harm reduction under unreliable sources in domains the model knows}, and we do not claim reliability sufficient for medical or legal decisions.

\textbf{Consistent wrong knowledge.} When a model recalls the same wrong fact in all six runs, support and self-verification are both high, and no internal signal catches the error (example: all six runs name John Maynard Smith as Dawkins\textquotesingle s adviser).

%% file: supplement/S3_external.tex
\section{ClashEval and Housing Statute QA: full tables, risk and coverage}
\label{supp:external}

\label{app:clasheval}

\paragraph{ClashEval, Llama-3.1-8B (direct comparison with TCR and CK-PLUG).}

All 10,942 records, whole-answer scoring; 89\% of them have a perturbed context, so every system has a negative total. The preregistered scorer gives the same values within 0.2 for every method row; it differs only for the plain reader (−70.2) and closed-book (−85.1), whose long outputs it truncates. The last column is ours at the registered point (π = 0.25) minus the method.

\begin{center}\footnotesize\setlength{\tabcolsep}{3.5pt}
\begin{tabular}{@{}>{\raggedright\arraybackslash}p{0.27\linewidth}rrrr@{}}
\toprule
 & All & original & perturbed & \makecell[r]{Ours − method\\{[}95\% CI{]}} \\
\midrule
GenRead & \textbf{−9.5} & −9.1 & \textbf{−9.5} & −19.7\,{\scriptsize[−23.6, −15.9]} \\
ours, π = 0.1 & −26.5 & 45.2 & −35.8 &  \\
\textbf{ours, π = 0.25} (registered primary point) & −29.1 & 44.8 & −38.6 &  \\
Astute RAG & −32.8 & 48.9 & −43.3 & +3.7\,{\scriptsize[−0.4, +7.5]} \\
ours with selector v2 (as originally registered) & −37.7 & 48.8 & −48.9 &  \\
closed-book & −38.3 & −39.1 & −38.2 &  \\
plain reader & −40.2 & 75.3 & −55.0 & +11.1\,{\scriptsize[+8.4, +13.6]} \\
CK-PLUG & −41.3 & 72.2 & −55.9 & +12.2\,{\scriptsize[+9.3, +14.9]} \\
CAD (added post hoc) & −44.9 & 71.1 & −59.8 & +15.7\,{\scriptsize[+13.3, +18.1]} \\
AdaCAD (added post hoc) & −50.4 & 78.9 & −67.0 & +21.3\,{\scriptsize[+18.7, +23.7]} \\
InstructRAG-ICL & −45.6 & \textbf{80.8} & −61.9 & +16.5\,{\scriptsize[+13.8, +19.1]} \\
USC & −51.4 & 75.6 & −67.7 & +22.2\,{\scriptsize[+19.6, +25.0]} \\
TCR & −54.7 & 76.2 & −71.6 & +25.6\,{\scriptsize[+22.9, +28.4]} \\
TruthfulRAG & −55.4 & 78.1 & −72.6 & +26.3\,{\scriptsize[+24.1, +28.6]} \\
FaithfulRAG & −55.5 & 74.6 & −72.2 & +26.4\,{\scriptsize[+23.7, +28.9]} \\
ClashEval token-probability correction & −55.9 & 76.6 & −72.9 & +26.7\,{\scriptsize[+23.5, +29.8]} \\
Self-Route & −57.6 & 72.0 & −74.3 & +28.5\,{\scriptsize[+25.9, +31.0]} \\
ours, π = 0.99 & −65.2 & 48.7 & −79.8 &  \\
\bottomrule
\end{tabular}\end{center}

TruthfulRAG\textquotesingle s official code fails on 88 records after the repair rounds, and our reader returns unparseable output on 101; both count as abstentions.

\textbf{H-v3 holds; H-v3b is not confirmed.} Ours beats the plain reader by 11.1 points and both same-backbone control methods, CK-PLUG by 12.2 and TCR by 25.6. Among the methods that read the provided document it beats nine by 11 to 29 points and ties Astute RAG (+3.7 {[}−0.4, +7.5{]}). The one method with a higher total is GenRead (by 19.7 points), which does not read the provided document: its score is the same on original and perturbed documents (−9.1 and −9.5), and on original documents it is 53.9 points below ours (−9.1 against 44.8). Because 89\% of the records have a perturbed document, ignoring the document is rewarded on this total. H-v3b was registered against the best baseline whatever its design, so we report it as not confirmed.

CAD and AdaCAD were not in the preregistration and were run afterwards with fixed settings from their papers (Appendix~\ref{app:baselines}). Both score below the plain reader, CAD by 4.7 points ({[}−7.1, −2.3{]}) and AdaCAD by 10.2 ({[}−12.5, −7.9{]}). On perturbed records CAD answers less often than the plain reader (75\% against 80\%) and AdaCAD more often (86\%), and both are wrong on about nine in ten of the records they answer (90\% and 89\%, against 84\%). The pattern of Section~\ref{sec:diagnosis} repeats on this second data set. AdaCAD has the second-highest score on original documents (78.9) and falls 12 points below the plain reader on perturbed ones (−67.0 against −55.0): amplifying the context by its divergence from the model\textquotesingle s prior pays when the document is right and costs when it is wrong, and the divergence does not say which kind of document is being read.

\begin{figure}[!htbp]
  \centering
  \includegraphics[width=0.9\linewidth]{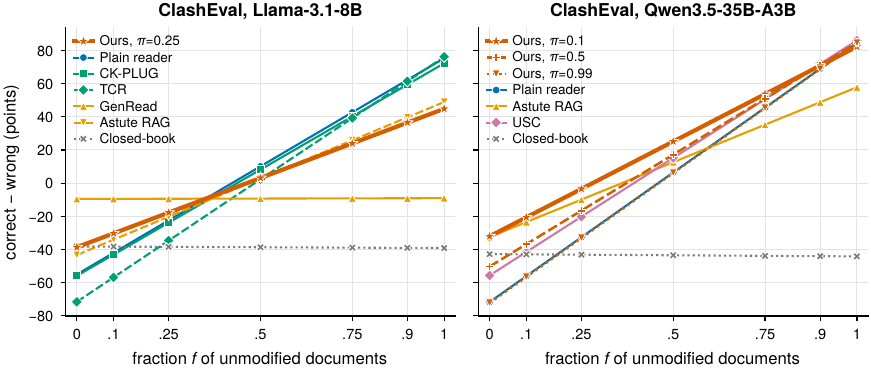}
  \caption{ClashEval (preregistered external test, whole-answer scoring): C−W as a function of the fraction f of documents left unmodified. Left, Llama: among the methods that read the document, ours is best when most documents are perturbed and falls behind from f ≈ 0.4, because its score on unmodified documents is capped by the candidate defect of Appendix~\ref{app:lessons}. GenRead ignores the document and abstains on 62\% of the questions, so its line is flat near −9: it is highest for f below about 0.35 and lowest above 0.6. Right, Qwen: lowering π helps when documents are unreliable, and at π = 0.99 the system coincides with the plain reader, so no setting of f makes it worse than the reader it is built on. Closed-book is flat by construction.}
  \label{fig:clasheval_mixture}
\end{figure}

\begin{itemize}
\item
  \textbf{GenRead leads by abstaining.} GenRead never reads the retrieved document, so its score is the same on original and perturbed records (−9.1 and −9.5). With Llama it abstains on 61.9\% of the records, is correct on 14.3\% and wrong on 23.8\%. Ours at the registered point abstains on 46.9\%, is correct on 12.0\% and wrong on 41.1\%. With Qwen the same method answers almost always and scores −35.9, below ours.
\item
  \textbf{The ranking depends on how many documents are wrong.} GenRead cannot use a correct document, so it stays near −9 at every mixture. Ours passes it once about a third of the documents are unmodified (\cref{fig:clasheval_mixture}).
\item
  \textbf{Why the prior does not close the gap.} Proposition~\ref{prop:shift} makes π exact only when the selector is calibrated on the deployment\textquotesingle s distribution. On ClashEval it is not: with Llama it states a mean probability of 0.49 for answers that are right 18\% of the time, and setting π to the true share lowers the calibration error only from 0.35 to 0.31 (Section~\ref{sec:envelope}, Figure~\ref{fig:theory}c). The selector therefore commits to wrong answers that a calibrated one would withhold, and GenRead, which abstains instead, scores higher on this mixture.
\end{itemize}

Mixtures of original and perturbed contexts (fraction f of original contexts):

\begin{center}\footnotesize\setlength{\tabcolsep}{3.5pt}
\begin{tabular}{@{}lrrrrr@{}}
\toprule
f & 0.1 & 0.25 & 0.5 & 0.75 & 0.9 \\
\midrule
ours, π = 0.1 & −27.7 & −15.5 & 4.7 & 25.0 & 37.1 \\
ours, π = 0.25 & −30.3 & −17.8 & 3.1 & 24.0 & 36.5 \\
GenRead & \textbf{−9.5} & \textbf{−9.4} & −9.3 & −9.2 & −9.1 \\
Astute RAG & −34.1 & −20.3 & 2.8 & 25.9 & 39.7 \\
CK-PLUG & −43.1 & −23.9 & 8.2 & 40.2 & 59.4 \\
TCR & −56.8 & −34.6 & 2.3 & 39.2 & 61.4 \\
plain reader & −42.0 & −22.5 & \textbf{10.1} & 42.7 & 62.3 \\
InstructRAG-ICL & −47.6 & −26.2 & 9.4 & \textbf{45.1} & \textbf{66.5} \\
\bottomrule
\end{tabular}\end{center}

With Llama, ours is ahead of every method that reads the document when most documents are wrong, and behind the plain readers when at least half are right. A high π does not recover the original-context score (45 to 51 at every π against 75.3 for raw), so H-π holds only weakly. The cause is a format defect in candidate construction, analysed in Appendix~\ref{app:lessons}: the plain reader\textquotesingle s candidate was truncated to the subject entity for value questions.

\paragraph{ClashEval, Qwen (supplementary: no same-backbone run of TCR or CK-PLUG).}

All 10,942 records, whole-answer scoring; the preregistered scorer gives the same values within 0.1 for every row except closed-book (−49.1). The last column is ours at the registered point (π = 0.25) minus the method, with a 95\% paired interval.

\begin{center}\footnotesize\setlength{\tabcolsep}{3.5pt}
\begin{tabular}{@{}>{\raggedright\arraybackslash}p{0.27\linewidth}rrrr@{}}
\toprule
 & All & original & perturbed & \makecell[r]{Ours − method\\{[}CI{]}} \\
\midrule
ours, π = 0.1 & \textbf{−19.0} & 82.3 & \textbf{−32.1} &  \\
Astute RAG & −22.2 & 57.8 & −32.5 & −0.3\,{\scriptsize[−3.0, +2.5]} \\
\textbf{ours, π = 0.25} (registered primary point) & −22.5 & 83.2 & −36.1 &  \\
ours with selector v2 (as originally registered) & −28.3 & 83.5 & −42.7 &  \\
GenRead & −35.9 & −38.5 & −35.5 & +13.3\,{\scriptsize[8.2, 18.7]} \\
USC & −39.5 & 86.0 & −55.7 & +17.0\,{\scriptsize[15.0, 19.0]} \\
InstructRAG-ICL & −40.0 & 84.8 & −56.0 & +17.5\,{\scriptsize[14.9, 20.0]} \\
closed-book & −43.0 & −44.2 & −42.8 &  \\
ClashEval token-probability correction & −47.7 & \textbf{87.8} & −65.1 & +25.2\,{\scriptsize[22.7, 28.0]} \\
Self-Route & −52.8 & 84.5 & −70.5 & +30.3\,{\scriptsize[27.7, 32.9]} \\
TruthfulRAG & −53.0 & 86.0 & −70.9 & +30.5\,{\scriptsize[28.1, 33.1]} \\
plain reader & −53.9 & 84.6 & −71.7 & +31.4\,{\scriptsize[29.0, 33.8]} \\
ours, π = 0.99 & −54.4 & 84.5 & −72.2 &  \\
FaithfulRAG & −55.3 & 86.0 & −73.5 & +32.8\,{\scriptsize[30.1, 35.6]} \\
\bottomrule
\end{tabular}\end{center}

\begin{itemize}
\item
  \textbf{Against ten of the eleven baselines the registered point is better by 13 to 33 points.} Against Astute RAG it is a tie on the total (−0.3 {[}−3.0, +2.5{]}). H-v3b ("at least as good as the best baseline") therefore holds only as "not worse": we do not claim a win over Astute RAG on this total.
\item
  \textbf{The two tied systems get there differently.} Astute RAG gives up the unmodified documents (57.8 against 83.2 for ours): it distrusts the context everywhere. Ours keeps the unmodified documents at the plain reader\textquotesingle s level and is 3.6 points behind Astute RAG on perturbed ones. Since 89\% of ClashEval records are perturbed, the total rewards blanket distrust. For any mixture with at least a quarter of unmodified documents ours is ahead: f = 0.25: −6.3 against −9.9; f = 0.5: 23.5 against 12.6; f = 0.75: 53.3 against 35.2; f = 0.9: 71.2 against 48.8 (\cref{fig:clasheval_mixture}, right).
\item
  At maximum π the system reproduces the plain reader (84.5 against 84.6 on original contexts; total difference {[}−1.1, +0.1{]}).
\item
  Ours is at least as good as the plain reader at every mixture (f = 0.1: −16.8 against −54.2; f = 0.9: 71.7 against 69.2), and the best π rises with f (0.1 → 0.5). H-π holds.
\item
  The Qwen plain reader almost never uses the list format (1.2\% of answers), so the candidate defect does not arise.
\end{itemize}

\cref{fig:clasheval_mixture} shows both panels. The system should be judged by where its line lies against the plain reader\textquotesingle s: above it for every f with Qwen, above it only for f below about 0.4 with Llama.

\paragraph{Housing Statute QA, Qwen.}

1,000 yes/no questions; the gold statute is retrieved for about 30\% of them.

\begin{center}\footnotesize\setlength{\tabcolsep}{3.5pt}
\begin{tabular}{@{}lrrr@{}}
\toprule
 & C−W & \makecell[r]{correct / wrong\\/ abstain (\%)} & \makecell[r]{Ours − plain\\reader {[}CI{]}} \\
\midrule
TruthfulRAG & \textbf{18.7} & 34.4 / 15.7 / 49.9 &  \\
InstructRAG-ICL & \textbf{18.7} &  &  \\
FaithfulRAG & 12.4 &  &  \\
plain reader & 11.4 & 29.0 / 17.6 / 53.4 &  \\
ours, π = 0.99 & 8.8 & 28.3 / 19.5 / 52.2 & −2.6\,{\scriptsize[−4.0, −1.3]} \\
ours, π = 0.5 & 5.2 &  & −6.2\,{\scriptsize[−8.4, −4.1]} \\
\textbf{ours, π = 0.25} (registered primary point) & 2.0 & 20.0 / 18.0 / 62.0 & −9.4\,{\scriptsize[−12.7, −6.1]} \\
closed-book & −27.4 & 33.1 / 60.5 / 6.4 &  \\
\bottomrule
\end{tabular}\end{center}

\textbf{H-v3 fails on Housing.} The model\textquotesingle s knowledge of housing law is harmful (closed-book wrong on 60.5\%). Our score rises monotonically with π, the direction H-π predicts, but maximum π only approaches the plain reader.

\paragraph{Housing Statute QA, Llama-3.1-8B (direct comparison with TCR and CK-PLUG).}

\begin{center}\footnotesize\setlength{\tabcolsep}{3.5pt}
\begin{tabular}{@{}lrrr@{}}
\toprule
 & C−W & \makecell[r]{correct / wrong\\/ abstain (\%)} & \makecell[r]{Ours (π = 0.25)\\− method {[}95\%\\CI{]}} \\
\midrule
GenRead & \textbf{11.6} & 54.8 / 43.2 / 2.0 & −22.5\,{\scriptsize[−29.1, −16.1]} \\
InstructRAG-ICL & 4.6 & 33.7 / 29.1 / 37.2 & −15.5\,{\scriptsize[−20.7, −10.3]} \\
ours, π = 0.1 & −7.9 & 0.5 / 8.4 / 91.1 &  \\
TruthfulRAG & −9.0 & 37.2 / 46.2 / 16.6 & −1.9\,{\scriptsize[−7.9, +4.0]} \\
CK-PLUG & −10.2 &  & −0.7\,{\scriptsize[−5.7, +4.4]} \\
\textbf{ours, π = 0.25} (registered primary point) & −10.9 & 1.2 / 12.1 / 86.7 &  \\
plain reader & −11.9 & 2.1 / 14.0 / 83.9 & +1.0\,{\scriptsize[−2.0, +3.9]} \\
TCR & −15.9 &  & +5.0\,{\scriptsize[−0.7, +10.3]} \\
FaithfulRAG & −19.8 &  & +8.9\,{\scriptsize[+3.1, +14.6]} \\
USC & −22.2 &  & +11.3\,{\scriptsize[+5.3, +17.0]} \\
Astute RAG & −23.9 &  & +13.0\,{\scriptsize[+7.8, +18.2]} \\
ours, π = 0.5 & −26.1 & 3.7 / 29.8 / 66.5 &  \\
closed-book & −27.1 & 5.3 / 32.4 / 62.3 &  \\
Self-Route & −28.5 &  & +17.6\,{\scriptsize[+13.2, +21.9]} \\
ours, π = 0.99 & −35.9 & 3.7 / 39.6 / 56.7 &  \\
\bottomrule
\end{tabular}\end{center}

The preregistered scorer agrees with these numbers within 1.2 points for every method row; it differs only for the plain reader (−33.1) and closed-book (−90.0), whose long outputs it truncates. TruthfulRAG\textquotesingle s official code fails on 68 of 1,000 questions after one retry; those count as abstentions. Our reader returns unparseable output on 30 questions; those also count as abstentions.

\textbf{H-v3 fails on Housing with Llama too, and the test is uninformative for our system.} With this backbone our frozen plain reader does not begin a single one of its 1,000 answers with yes or no: it mostly returns \texttt{INSUFFICIENT} or a list of entities, abstains on 83.9\% and is correct on 2.1\%. With Qwen the same prompt begins 442 answers with yes or no. Every system built on that reader inherits the failure. At the registered point our system abstains on 86.7\% and ties the plain reader, CK-PLUG and TruthfulRAG; the comparison with TCR is not significant. GenRead and InstructRAG-ICL, which write free text, are the only positive methods and beat ours by 15 to 23 points. The score falls as π rises, the opposite of Qwen, because a higher π makes the selector answer more often and the answers it then gives are wrong (39.6\% wrong at π = 0.99). We report this as a failure of the frozen reader prompt on yes/no legal questions, not as evidence about merging.

\paragraph{Risk and coverage on ClashEval.}

\begin{figure}[!htbp]
  \centering
  \includegraphics[width=0.9\linewidth]{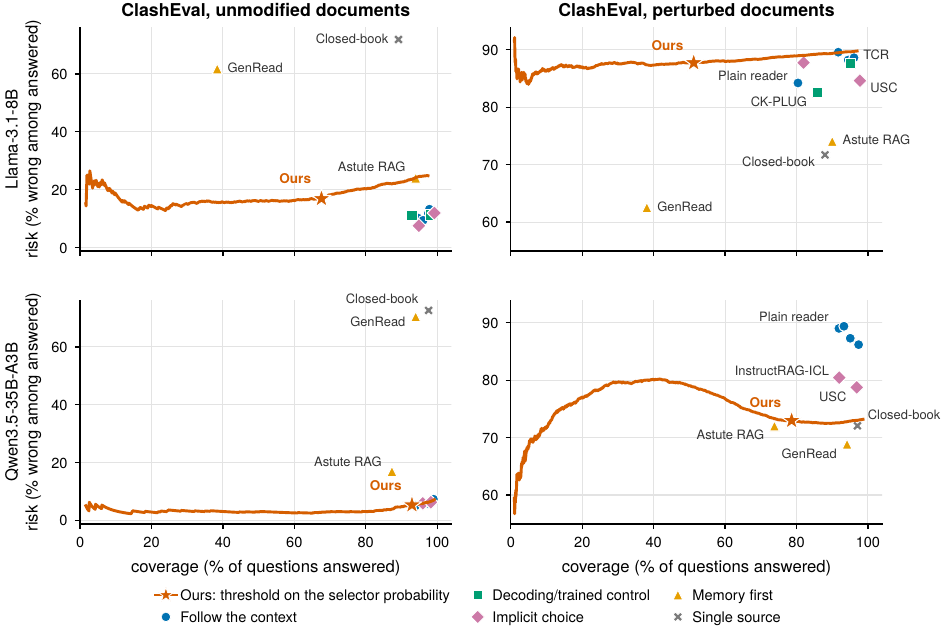}
  \caption{Risk against coverage on ClashEval (lockbox 3, whole-answer scoring), separately for unmodified and perturbed documents. Points are methods; the curve is our system at π = 0.25 when it answers only above a threshold on the selector probability, and the star is the frozen operating point (threshold 0.5). On perturbed documents the curve does not fall as the threshold rises: the selector probability does not rank answers by correctness there.}
  \label{fig:riskcov}
\end{figure}

C−W mixes two quantities: how often a system answers, and how often its answers are wrong. \cref{app:riskcov} (\cref{fig:riskcov}) separates them. With Qwen, on perturbed documents our system answers 79\% of the questions with 73\% of the answers wrong, against 92\% and 89\% for the plain reader; on unmodified documents the two are equal (93\% and 5\% against 94\% and 5\%). The gain over the plain reader therefore comes mostly from fewer wrong answers. With Llama this is not the case: on perturbed documents our system answers 51\% with 88\% wrong, against 80\% and 84\% for the plain reader, and on unmodified documents 68\% with 17\% wrong against 95\% and 10\%. Its 11.1-point gain in C−W comes from answering less often when the document is wrong, and its answers are not more often right. On perturbed documents the selector probability does not rank answers by correctness with either backbone, so raising the threshold does not buy a low-risk operating point (\cref{app:where}).

\subsection{Risk and coverage}
\label{app:riskcov}

Coverage is the share of questions answered; risk is the share of wrong answers among the answered.

\begin{center}\footnotesize\setlength{\tabcolsep}{3.5pt}
\begin{tabular}{@{}lrrrr@{}}
\toprule
Coverage / risk (\%) & \makecell[r]{Llama,\\unmodified} & Llama, perturbed & Qwen, unmodified & Qwen, perturbed \\
\midrule
ours, π = 0.25 & 68 / 17 & 51 / 88 & 93 / 5 & 79 / 73 \\
ours, π = 0.99 & 96 / 25 & 95 / 92 & 95 / 6 & 94 / 89 \\
plain reader & 95 / 10 & 80 / 84 & 94 / 5 & 92 / 89 \\
CK-PLUG & 93 / 11 & 86 / 83 & n/a & n/a \\
CAD (post hoc) & 90 / 11 & 75 / 90 & n/a & n/a \\
AdaCAD (post hoc) & 96 / 9 & 86 / 89 & n/a & n/a \\
TCR & 98 / 11 & 95 / 88 & n/a & n/a \\
InstructRAG-ICL & 95 / 7 & 82 / 88 & 96 / 6 & 92 / 80 \\
USC & 99 / 12 & 98 / 85 & 98 / 6 & 97 / 79 \\
Astute RAG & 94 / 24 & 90 / 74 & 87 / 17 & 74 / 72 \\
GenRead & 38 / 62 & 38 / 63 & 94 / 70 & 94 / 69 \\
closed-book & 89 / 72 & 88 / 72 & 98 / 73 & 97 / 72 \\
\bottomrule
\end{tabular}\end{center}

%% file: supplement/S4_mcq.tex
\section{Five-variant multiple-choice GRACE and deployed GraphRAG-family systems}
\label{supp:mcq}

\label{app:mcq}

\textbf{Setting.} The multiple-choice form of GRACE gives each question four options (the real-world answer, the injected answer of the swapping variant, a distractor or the model\textquotesingle s own pretraining answer, and "insufficient information") and five graph variants: clean, broken (the answer edge is removed), type-matching distractors (entities of the answer\textquotesingle s type are added), hub distractors (high-degree entities and their neighbourhoods are added near the answer) and swapping (the answer is replaced by an injected entity). Option letters are shuffled per question. Six retrieval-augmented systems answer with the same generator (Qwen3.5-35B-A3B). Five use the same evidence-neutral prompt, which presents the context and the model\textquotesingle s own knowledge as two sources of evidence and does not tell the model to prefer either: naive RAG, GraphRAG (community reports, local search, bge-m3 embeddings), LightRAG (custom-graph adaptation), RAPTOR and a FAICO-style Wikidata adaptation. The sixth is naive RAG with a routing classifier whose prompt sends each question either to the model\textquotesingle s own knowledge or to the context. Evaluation uses the 3,400 test questions for which every system has all five variants (17,000 observations). Acc1 is the no-context probe of the same generator (strict entity match, 6.6\%, 226 questions). Acc2 scores each variant against its own ground truth (for broken and swapping this is "insufficient" or the injected answer unless the model is known to hold the fact). Acc3 scores every variant against the real-world answer. Answers are parsed by the stored letter or an explicit A--D prefix; missing and unparsable answers count as wrong.

\begin{center}\footnotesize\setlength{\tabcolsep}{3.5pt}
\begin{tabular}{@{}lrrrrrrr@{}}
\toprule
 &  &  & \multicolumn{5}{c}{Acc3 by variant} \\
\cmidrule(l{3pt}r{3pt}){4-8}
System & Acc2 & \makecell[r]{Acc3\\(all)} & clean & broken & \makecell[r]{type-\\matching} & hub & swapping \\
\midrule
Naive RAG & 52.3\,{\scriptsize[51.2, 53.5]} & 51.7\,{\scriptsize[50.6, 52.9]} & 68.5 & 40.7 & 64.4 & 68.2 & 16.8 \\
GraphRAG & \textbf{56.2}\,{\scriptsize[55.0, 57.4]} & 53.6\,{\scriptsize[52.5, 54.7]} & \textbf{71.2} & 36.2 & \textbf{67.7} & \textbf{77.1} & 15.8 \\
LightRAG & 47.3\,{\scriptsize[46.2, 48.4]} & 45.8\,{\scriptsize[44.7, 46.8]} & 60.9 & 38.5 & 55.1 & 62.8 & 11.5 \\
RAPTOR & 50.3\,{\scriptsize[49.2, 51.4]} & 50.3\,{\scriptsize[49.1, 51.5]} & 68.0 & 44.3 & 58.1 & 59.4 & 21.7 \\
FAICO-style & 31.8\,{\scriptsize[30.7, 32.9]} & 19.1\,{\scriptsize[18.1, 20.0]} & 29.6 & 4.4 & 29.8 & 28.5 & 3.2 \\
Naive RAG + routing classifier & 55.1\,{\scriptsize[54.0, 56.2]} & \textbf{58.1}\,{\scriptsize[56.5, 59.7]} & 59.5 & \textbf{56.5} & 59.4 & 59.6 & \textbf{55.5} \\
\bottomrule
\end{tabular}\end{center}

\emph{Percent; 95\% CIs from a question-level bootstrap.}

\textbf{Distractors do little; conflicts do the damage.} Adding type-matched or high-degree entities changes Acc3 by at most 10 points for any context-reading system, and GraphRAG\textquotesingle s community structure even helps on the hub variant (77.1 against 71.2 on clean). The two variants that remove or contradict the answer are where every context-reading system collapses: Acc3 falls by 22 to 35 points on broken and by 46 to 55 points on swapping (by 26 for the FAICO-style system, which already answers "insufficient" on most clean graphs). These are the broken and wrong\_branch states of the open-form protocol, and the graph-based systems are not more robust to them than naive RAG.

\textbf{The systems override what the model knows.} On the 226 questions the generator answers correctly with no context:

\begin{center}\footnotesize\setlength{\tabcolsep}{3.5pt}
\begin{tabular}{@{}lrrrr@{}}
\toprule
\makecell[l]{System (226 questions\\the model knows)} & \makecell[r]{broken\\real answer} & \makecell[r]{broken\\"insufficient"} & \makecell[r]{swapping\\real answer} & \makecell[r]{swapping\\injected\\answer} \\
\midrule
Naive RAG & 88 & 12 & 54 & 23 \\
GraphRAG & 78 & 21 & 45 & 38 \\
LightRAG & 79 & 15 & 23 & 46 \\
RAPTOR & 91 & 9 & 63 & 18 \\
FAICO-style & 17 & 83 & 11 & 17 \\
Naive RAG + routing classifier & 99 & 0 & 99 & 0 \\
\bottomrule
\end{tabular}\end{center}

\emph{Percent of the 226 questions; the remaining share is the other distractor or an invalid answer.}

On clean graphs naive RAG, GraphRAG, LightRAG and RAPTOR pick the real answer for 89--97\% of these questions. When the graph is swapped, GraphRAG picks the injected answer for 38\% of questions the model knows, and LightRAG for 46\%; when the answer edge is removed, GraphRAG says "insufficient information" for 21\% of them. The evidence-neutral prompt allows the model to use its own knowledge, so the loss is not caused by an instruction to follow the context. It is the behaviour Section~\ref{sec:diagnosis} measures on the open form: retrieved context overrides correct knowledge.

\textbf{Routing fixes the known questions and pays elsewhere.} The routing classifier keeps 99\% of the known questions in every variant and has the best Acc3, but it does so by choosing one source per question. On the 3,174 questions the model does not know, it is 9 points below naive RAG on clean graphs (57 against 66), because questions routed to the model\textquotesingle s own knowledge no longer read a correct context. This is the selection trade-off of Section~\ref{sec:diagnosis} in a different format and with a different generator: a system that chooses one source per question gains on conflicts and loses where the other source was right. Our method merges both sources in the answer instead of choosing one (Section~\ref{sec:method}); it is evaluated on the open form, where the four states have a single real-world ground truth, and is not run on this multiple-choice form.

%% file: supplement/S5_design.tex
\section{Selector v4 and earlier directions that did not work}
\label{supp:design}

\label{app:v4}

Selector v4 is selector v3 with one change: the plain reader\textquotesingle s candidate is extracted from its whole answer, not from the first \textquotesingle\textbar\textquotesingle-separated segment (Appendix~\ref{app:lessons}). It is trained on the same data and frozen with hashes.

\begin{center}\footnotesize\setlength{\tabcolsep}{3.5pt}
\begin{tabular}{@{}lrr@{}}
\toprule
C−W, protocol v3 & v3 & v4 \\
\midrule
GRACE dev, Llama, π = 0.25 & 35.0 & 36.7 \\
GRACE dev, Llama, π = 0.99 & 24.4 & 30.3 \\
GRACE dev, Qwen, π = 0.25 & 45.3 & 45.7 \\
GRACE-QT dev, non-boolean, Llama & 1.3 & 0.3 \\
GRACE-QT dev, non-boolean, Qwen & 2.7 & 2.4 \\
\bottomrule
\end{tabular}\end{center}

The change matters for Llama at high π, where the context candidate decides the answer. v4 has no untouched test: the defect it fixes was first seen on ClashEval, so ClashEval cannot be evidence for it. All results in the main text use v3.

\subsection{Earlier directions that did not work}
\label{app:earlier}

These are reported because each one fixed a design decision of Section~\ref{sec:method}.

{\footnotesize\setlength{\tabcolsep}{3pt}
\begin{longtable}[]{@{}>{\raggedright\arraybackslash}p{0.188\linewidth}>{\raggedright\arraybackslash}p{0.245\linewidth}>{\raggedright\arraybackslash}p{0.197\linewidth}>{\raggedright\arraybackslash}p{0.315\linewidth}@{}}
\toprule\noalign{}
Direction & What was tried & Result & Lesson \\
\midrule\noalign{}
\endhead
\bottomrule\noalign{}
\endlastfoot
Graph-edit policy trained with reinforcement learning & a teacher LLM proposes edits to the context graph; a small policy accepts, rejects or stops & Acc3 50.3\% against 51.1\% for applying every edit & the reward gave nothing for a correct added edge that was not a gold edge, so stopping early was always safe; the policy discarded useful knowledge \\
Self-verification as a filter & each proposed triple is scored alone by the LLM (AUC 0.875 on 800 held-out facts); rejected triples are deleted & 50.4\%, equal to deleting the same number of triples at random (50.6\%) & the model knows which facts are false, but deleting a false fact with no replacement removes the relation and the reader answers "insufficient" (27 of 29 harmed wrong\_branch cases) \\
Verify, then repair & a rejected object is refilled from the model\textquotesingle s knowledge and re-verified; never delete without a replacement & +4.8 to +6.8 Acc3 over the plain reader on two fresh slices & useful, but the context is edited, and stale knowledge can overwrite a correct, newer fact \\
Confidence tags on context facts & each context fact annotated with the model\textquotesingle s belief (three wordings) & −1.4 to −4.4 Acc3 on every slice & 14\% of correct facts receive a low score; a low score mixes "unfamiliar" with "false", and the reader then ignores correct facts \\
Gates on a "does the model know" signal (WebQSP / CWQ validation) & answer from memory when a signal passes a threshold & C−W stayed near 27 & the best single signal reaches AUC 0.78 (agreement over five samples 0.73, minimum token probability 0.78, linear probe on the last hidden layer 0.76); a gate also hides the knowledge \\
Attention-based selector & candidates, notes and facts as tokens; softmax over candidates and "none" & C−W 37.6 against 48.2 for gradient boosting (Qwen, earlier scoring rule) & with 3,600 records the transformer does not learn which note supports which candidate; the softmax spreads probability so the cautious mode abstains too often \\
π as a learned input feature of the selector & training copies with random π and reweighted clean records & almost no effect outside the training distribution & replaced by the analytic prior shift of Eq.~(\ref{eq:shift}) \\
\end{longtable}}

The first observation of "context silences knowledge" (Appendix~\ref{app:findings}) came from CWQ validation data with the earlier scoring rule: on questions whose context lacks the answer, the model was correct for 35.4\% with no context and for 2.2\% to 8.7\% with context, depending on the prompt. Appendix~\ref{app:findings} reports the same effect on GRACE dev under protocol v3.

%% file: arxiv.bbl
\begin{thebibliography}{54}
\providecommand{\natexlab}[1]{#1}
\providecommand{\url}[1]{\texttt{#1}}
\expandafter\ifx\csname urlstyle\endcsname\relax
  \providecommand{\doi}[1]{doi: #1}\else
  \providecommand{\doi}{doi: \begingroup \urlstyle{rm}\Url}\fi

\bibitem[Bi et~al.(2026)Bi, Liu, Wang, Xu, Fang, Mei, and Cheng]{bi2026ckplug}
Baolong Bi, Shenghua Liu, Yiwei Wang, Yilong Xu, Junfeng Fang, Lingrui Mei, and
  Xueqi Cheng.
\newblock Parameters vs. context: Fine-grained control of knowledge reliance in
  language models.
\newblock In \emph{International Conference on Learning Representations}, 2026.
\newblock arXiv:2503.15888.

\bibitem[Brier(1950)]{brier1950}
Glenn~W. Brier.
\newblock Verification of forecasts expressed in terms of probability.
\newblock \emph{Monthly Weather Review}, 78\penalty0 (1):\penalty0 1--3, 1950.

\bibitem[Chen et~al.(2024{\natexlab{a}})Chen, Xiao, Zhang, Luo, Lian, and
  Liu]{chen2024bgem3}
Jianlv Chen, Shitao Xiao, Peitian Zhang, Kun Luo, Defu Lian, and Zheng Liu.
\newblock {BGE M3-Embedding}: Multi-lingual, multi-functionality,
  multi-granularity text embeddings through self-knowledge distillation.
\newblock \emph{arXiv preprint arXiv:2402.03216}, 2024{\natexlab{a}}.

\bibitem[Chen et~al.(2024{\natexlab{b}})Chen, Aksitov, Alon, Ren, Xiao, Yin,
  Prakash, Sutton, Wang, and Zhou]{chen2024usc}
Xinyun Chen, Renat Aksitov, Uri Alon, Jie Ren, Kefan Xiao, Pengcheng Yin,
  Sushant Prakash, Charles Sutton, Xuezhi Wang, and Denny Zhou.
\newblock Universal self-consistency for large language models.
\newblock In \emph{ICML 2024 Workshop on In-Context Learning},
  2024{\natexlab{b}}.
\newblock arXiv:2311.17311.

\bibitem[Chow(1970)]{chow1970reject}
C.~K. Chow.
\newblock On optimum recognition error and reject tradeoff.
\newblock \emph{IEEE Transactions on Information Theory}, 16\penalty0
  (1):\penalty0 41--46, 1970.

\bibitem[Du et~al.(2024)Du, Sn{\ae}bjarnarson, Stoehr, White, Schein, and
  Cotterell]{du2024context}
Kevin Du, V{\'e}steinn Sn{\ae}bjarnarson, Niklas Stoehr, Jennifer~C. White,
  Aaron Schein, and Ryan Cotterell.
\newblock Context versus prior knowledge in language models.
\newblock In \emph{Proceedings of the 62nd Annual Meeting of the Association
  for Computational Linguistics}, 2024.

\bibitem[Dubey et~al.(2019)Dubey, Banerjee, Abdelkawi, and
  Lehmann]{dubey2019lcquad2}
Mohnish Dubey, Debayan Banerjee, Abdelrahman Abdelkawi, and Jens Lehmann.
\newblock {LC-QuAD} 2.0: A large dataset for complex question answering over
  {Wikidata} and {DBpedia}.
\newblock In \emph{The Semantic Web -- ISWC 2019}, 2019.

\bibitem[Farquhar et~al.(2024)Farquhar, Kossen, Kuhn, and
  Gal]{farquhar2024semanticentropy}
Sebastian Farquhar, Jannik Kossen, Lorenz Kuhn, and Yarin Gal.
\newblock Detecting hallucinations in large language models using semantic
  entropy.
\newblock \emph{Nature}, 630:\penalty0 625--630, 2024.

\bibitem[Friedman(2001)]{friedman2001gbm}
Jerome~H. Friedman.
\newblock Greedy function approximation: A gradient boosting machine.
\newblock \emph{The Annals of Statistics}, 29\penalty0 (5):\penalty0
  1189--1232, 2001.

\bibitem[Genest and Zidek(1986)]{genest1986combining}
Christian Genest and James~V. Zidek.
\newblock Combining probability distributions: A critique and an annotated
  bibliography.
\newblock \emph{Statistical Science}, 1\penalty0 (1):\penalty0 114--135, 1986.

\bibitem[Gneiting and Raftery(2007)]{gneiting2007proper}
Tilmann Gneiting and Adrian~E. Raftery.
\newblock Strictly proper scoring rules, prediction, and estimation.
\newblock \emph{Journal of the American Statistical Association}, 102\penalty0
  (477):\penalty0 359--378, 2007.

\bibitem[Huang et~al.(2025)Huang, Chen, Cai, and Dhingra]{huang2025situated}
Yukun Huang, Sanxing Chen, Hongyi Cai, and Bhuwan Dhingra.
\newblock To trust or not to trust? enhancing large language models' situated
  faithfulness to external contexts.
\newblock In \emph{International Conference on Learning Representations}, 2025.

\bibitem[Lewis et~al.(2020)Lewis, Perez, Piktus, Petroni, Karpukhin, Goyal,
  K{\"u}ttler, Lewis, Yih, Rockt{\"a}schel, Riedel, and Kiela]{lewis2020rag}
Patrick Lewis, Ethan Perez, Aleksandra Piktus, Fabio Petroni, Vladimir
  Karpukhin, Naman Goyal, Heinrich K{\"u}ttler, Mike Lewis, Wen-tau Yih, Tim
  Rockt{\"a}schel, Sebastian Riedel, and Douwe Kiela.
\newblock Retrieval-augmented generation for knowledge-intensive {NLP} tasks.
\newblock In \emph{Advances in Neural Information Processing Systems}, 2020.

\bibitem[Li et~al.(2024)Li, Li, Zhang, Mei, and Bendersky]{li2024selfroute}
Zhuowan Li, Cheng Li, Mingyang Zhang, Qiaozhu Mei, and Michael Bendersky.
\newblock Retrieval augmented generation or long-context {LLM}s? a
  comprehensive study and hybrid approach.
\newblock In \emph{Proceedings of the 2024 Conference on Empirical Methods in
  Natural Language Processing: Industry Track}, 2024.

\bibitem[Lin et~al.(2026)Lin, Wen, Su, Tan, Sun, Chen, Bao, and
  Lv]{lin2026knowledgeabler1}
Chenyu Lin, Yilin Wen, Du~Su, Hexiang Tan, Fei Sun, Muhan Chen, Chenfu Bao, and
  Zhonghou Lv.
\newblock Resisting contextual interference in {RAG} via parametric-knowledge
  reinforcement.
\newblock In \emph{International Conference on Learning Representations}, 2026.
\newblock arXiv:2506.05154.

\bibitem[Lin(1991)]{lin1991divergence}
Jianhua Lin.
\newblock Divergence measures based on the {S}hannon entropy.
\newblock \emph{IEEE Transactions on Information Theory}, 37\penalty0
  (1):\penalty0 145--151, 1991.

\bibitem[Lin et~al.(2022)Lin, Hilton, and Evans]{lin2022verbalized}
Stephanie Lin, Jacob Hilton, and Owain Evans.
\newblock Teaching models to express their uncertainty in words.
\newblock \emph{Transactions on Machine Learning Research}, 2022.

\bibitem[Lipton et~al.(2018)Lipton, Wang, and Smola]{lipton2018labelshift}
Zachary~C. Lipton, Yu-Xiang Wang, and Alexander~J. Smola.
\newblock Detecting and correcting for label shift with black box predictors.
\newblock In \emph{Proceedings of the 35th International Conference on Machine
  Learning}, volume~80 of \emph{PMLR}, pages 3122--3130, 2018.

\bibitem[Liu et~al.(2026)Liu, Shang, and Zhang]{liu2026truthfulrag}
Shuyi Liu, Yuming Shang, and Xi~Zhang.
\newblock {TruthfulRAG}: Resolving factual-level conflicts in
  retrieval-augmented generation with knowledge graphs.
\newblock In \emph{Proceedings of the AAAI Conference on Artificial
  Intelligence}, volume~40, 2026.
\newblock URL \url{https://ojs.aaai.org/index.php/AAAI/article/view/40489}.

\bibitem[{Llama Team, AI @ Meta}(2024)]{meta2024llama31}
{Llama Team, AI @ Meta}.
\newblock The {Llama} 3 herd of models.
\newblock \emph{arXiv preprint arXiv:2407.21783}, 2024.

\bibitem[Longpre et~al.(2021)Longpre, Perisetla, Chen, Ramesh, DuBois, and
  Singh]{longpre2021entity}
Shayne Longpre, Kartik Perisetla, Anthony Chen, Nikhil Ramesh, Chris DuBois,
  and Sameer Singh.
\newblock Entity-based knowledge conflicts in question answering.
\newblock In \emph{Proceedings of the 2021 Conference on Empirical Methods in
  Natural Language Processing}, pages 7052--7063, 2021.

\bibitem[Magesh et~al.(2025)Magesh, Surani, Dahl, Suzgun, Manning, and
  Ho]{magesh2025hallucinationfree}
Varun Magesh, Faiz Surani, Matthew Dahl, Mirac Suzgun, Christopher~D. Manning,
  and Daniel~E. Ho.
\newblock Hallucination-free? assessing the reliability of leading {AI} legal
  research tools.
\newblock \emph{Journal of Empirical Legal Studies}, 2025.

\bibitem[Mallen et~al.(2023)Mallen, Asai, Zhong, Das, Khashabi, and
  Hajishirzi]{mallen2023trust}
Alex Mallen, Akari Asai, Victor Zhong, Rajarshi Das, Daniel Khashabi, and
  Hannaneh Hajishirzi.
\newblock When not to trust language models: Investigating effectiveness of
  parametric and non-parametric memories.
\newblock In \emph{Proceedings of the 61st Annual Meeting of the Association
  for Computational Linguistics}, 2023.

\bibitem[Ming et~al.(2025)Ming, Purushwalkam, Pandit, Ke, Nguyen, Xiong, and
  Joty]{ming2025faitheval}
Yifei Ming, Senthil Purushwalkam, Shrey Pandit, Zixuan Ke, Xuan-Phi Nguyen,
  Caiming Xiong, and Shafiq Joty.
\newblock {FaithEval}: Can your language model stay faithful to context, even
  if ``the moon is made of marshmallows''.
\newblock In \emph{International Conference on Learning Representations}, 2025.

\bibitem[Neeman et~al.(2023)Neeman, Aharoni, Honovich, Choshen, Szpektor, and
  Abend]{neeman2023disentqa}
Ella Neeman, Roee Aharoni, Or~Honovich, Leshem Choshen, Idan Szpektor, and Omri
  Abend.
\newblock {DisentQA}: Disentangling parametric and contextual knowledge with
  counterfactual question answering.
\newblock In \emph{Proceedings of the 61st Annual Meeting of the Association
  for Computational Linguistics}, pages 10056--10070, 2023.

\bibitem[Ortu et~al.(2024)Ortu, Jin, Doimo, Sachan, Cazzaniga, and
  Sch{\"o}lkopf]{ortu2024competition}
Francesco Ortu, Zhijing Jin, Diego Doimo, Mrinmaya Sachan, Alberto Cazzaniga,
  and Bernhard Sch{\"o}lkopf.
\newblock Competition of mechanisms: Tracing how language models handle facts
  and counterfactuals.
\newblock In \emph{Proceedings of the 62nd Annual Meeting of the Association
  for Computational Linguistics}, pages 8420--8436, 2024.

\bibitem[Reid(2024)]{reid2024aioverviews}
Elizabeth Reid.
\newblock {AI} overviews: About last week.
\newblock Google, The Keyword (blog), 2024.
\newblock URL
  \url{https://blog.google/products/search/ai-overviews-update-may-2024/}.
\newblock 30 May 2024.

\bibitem[Saerens et~al.(2002)Saerens, Latinne, and
  Decaestecker]{saerens2002prior}
Marco Saerens, Patrice Latinne, and Christine Decaestecker.
\newblock Adjusting the outputs of a classifier to new a priori probabilities:
  A simple procedure.
\newblock \emph{Neural Computation}, 14\penalty0 (1):\penalty0 21--41, 2002.

\bibitem[Sen et~al.(2022)Sen, Aji, and Saffari]{sen2022mintaka}
Priyanka Sen, Alham~Fikri Aji, and Amir Saffari.
\newblock Mintaka: A complex, natural, and multilingual dataset for end-to-end
  question answering.
\newblock In \emph{Proceedings of the 29th International Conference on
  Computational Linguistics}, 2022.

\bibitem[Shenoy et~al.(2022)Shenoy, Ilievski, Garijo, Schwabe, and
  Szekely]{shenoy2022wikidataquality}
Kartik Shenoy, Filip Ilievski, Daniel Garijo, Daniel Schwabe, and Pedro
  Szekely.
\newblock A study of the quality of {Wikidata}.
\newblock \emph{Journal of Web Semantics}, 2022.

\bibitem[Shi et~al.(2024)Shi, Han, Lewis, Tsvetkov, Zettlemoyer, and
  Yih]{shi2024cad}
Weijia Shi, Xiaochuang Han, Mike Lewis, Yulia Tsvetkov, Luke Zettlemoyer, and
  Wen-tau Yih.
\newblock Trusting your evidence: Hallucinate less with context-aware decoding.
\newblock In \emph{Proceedings of the 2024 Conference of the North American
  Chapter of the Association for Computational Linguistics: Human Language
  Technologies (Volume 2: Short Papers)}, pages 783--791, 2024.

\bibitem[Su et~al.(2024)Su, Zhang, Qu, Zhu, Li, Sun, Li, Zhang, and
  Cheng]{su2024conflictbank}
Zhaochen Su, Jun Zhang, Xiaoye Qu, Tong Zhu, Yanshu Li, Jiashuo Sun, Juntao Li,
  Min Zhang, and Yu~Cheng.
\newblock {ConflictBank}: A benchmark for evaluating the influence of knowledge
  conflicts in {LLM}s.
\newblock In \emph{Advances in Neural Information Processing Systems (Datasets
  and Benchmarks Track)}, 2024.

\bibitem[Talmor and Berant(2018)]{talmor2018cwq}
Alon Talmor and Jonathan Berant.
\newblock The web as a knowledge-base for answering complex questions.
\newblock In \emph{Proceedings of the 2018 Conference of the North American
  Chapter of the Association for Computational Linguistics}, 2018.

\bibitem[Tian et~al.(2023)Tian, Mitchell, Zhou, Sharma, Rafailov, Yao, Finn,
  and Manning]{tian2023justask}
Katherine Tian, Eric Mitchell, Allan Zhou, Archit Sharma, Rafael Rafailov,
  Huaxiu Yao, Chelsea Finn, and Christopher~D. Manning.
\newblock Just ask for calibration: Strategies for eliciting calibrated
  confidence scores from language models fine-tuned with human feedback.
\newblock In \emph{Proceedings of the 2023 Conference on Empirical Methods in
  Natural Language Processing}, 2023.

\bibitem[Usbeck et~al.(2024)Usbeck, Yan, Perevalov, Jiang, Schulz, Kraft,
  M{\"o}ller, Huang, Reineke, Ngonga~Ngomo, Saleem, and Both]{usbeck2024qald10}
Ricardo Usbeck, Xi~Yan, Aleksandr Perevalov, Longquan Jiang, Julius Schulz,
  Angelie Kraft, Cedric M{\"o}ller, Junbo Huang, Jan Reineke, Axel-Cyrille
  Ngonga~Ngomo, Muhammad Saleem, and Andreas Both.
\newblock {QALD-10} -- the 10th challenge on question answering over linked
  data.
\newblock \emph{Semantic Web}, 2024.
\newblock \doi{10.3233/SW-233471}.

\bibitem[Vrande{\v{c}}i{\'c} and Kr{\"o}tzsch(2014)]{vrandecic2014wikidata}
Denny Vrande{\v{c}}i{\'c} and Markus Kr{\"o}tzsch.
\newblock Wikidata: A free collaborative knowledgebase.
\newblock \emph{Communications of the ACM}, 57\penalty0 (10):\penalty0 78--85,
  2014.

\bibitem[Wang et~al.(2025{\natexlab{a}})Wang, Wan, Sun, Chen, and
  Arik]{wang2025astute}
Fei Wang, Xingchen Wan, Ruoxi Sun, Jiefeng Chen, and Sercan~O. Arik.
\newblock Astute {RAG}: Overcoming imperfect retrieval augmentation and
  knowledge conflicts for large language models.
\newblock In \emph{Proceedings of the 63rd Annual Meeting of the Association
  for Computational Linguistics (Volume 1: Long Papers)}, pages 30553--30571,
  2025{\natexlab{a}}.
\newblock URL \url{https://aclanthology.org/2025.acl-long.1476}.

\bibitem[Wang et~al.(2025{\natexlab{b}})Wang, Prasad, Stengel-Eskin, and
  Bansal]{wang2025adacad}
Han Wang, Archiki Prasad, Elias Stengel-Eskin, and Mohit Bansal.
\newblock {AdaCAD}: Adaptively decoding to balance conflicts between contextual
  and parametric knowledge.
\newblock In \emph{Proceedings of the 2025 Conference of the Nations of the
  Americas Chapter of the Association for Computational Linguistics: Human
  Language Technologies (Volume 1: Long Papers)}, pages 11636--11652,
  2025{\natexlab{b}}.

\bibitem[Wang et~al.(2024)Wang, Feng, Wang, Shi, Balachandran, He, and
  Tsvetkov]{wang2024resolving}
Yike Wang, Shangbin Feng, Heng Wang, Weijia Shi, Vidhisha Balachandran,
  Tianxing He, and Yulia Tsvetkov.
\newblock Resolving knowledge conflicts in large language models.
\newblock In \emph{Conference on Language Modeling}, 2024.

\bibitem[Wei et~al.(2025)Wei, Chen, and Meng]{wei2025instructrag}
Zhepei Wei, Wei-Lin Chen, and Yu~Meng.
\newblock {InstructRAG}: Instructing retrieval-augmented generation via
  self-synthesized rationales.
\newblock In \emph{International Conference on Learning Representations}, 2025.
\newblock URL \url{https://openreview.net/forum?id=P1qhkp8gQT}.

\bibitem[Wen et~al.(2025)Wen, Yao, Feng, Xu, Tsvetkov, Howe, and
  Wang]{wen2025abstention}
Bingbing Wen, Jihan Yao, Shangbin Feng, Chenjun Xu, Yulia Tsvetkov, Bill Howe,
  and Lucy~Lu Wang.
\newblock Know your limits: A survey of abstention in large language models.
\newblock \emph{Transactions of the Association for Computational Linguistics},
  13, 2025.

\bibitem[West et~al.(2014)West, Gabrilovich, Murphy, Sun, Gupta, and
  Lin]{west2014kbcompletion}
Robert West, Evgeniy Gabrilovich, Kevin Murphy, Shaohua Sun, Rahul Gupta, and
  Dekang Lin.
\newblock Knowledge base completion via search-based question answering.
\newblock In \emph{Proceedings of the 23rd International Conference on World
  Wide Web}, 2014.

\bibitem[Wu et~al.(2024)Wu, Wu, and Zou]{wu2024clasheval}
Kevin Wu, Eric Wu, and James Zou.
\newblock {ClashEval}: Quantifying the tug-of-war between an {LLM}'s internal
  prior and external evidence.
\newblock In \emph{Advances in Neural Information Processing Systems (Datasets
  and Benchmarks Track)}, 2024.

\bibitem[Xie et~al.(2024)Xie, Zhang, Chen, Lou, and Su]{xie2024chameleon}
Jian Xie, Kai Zhang, Jiangjie Chen, Renze Lou, and Yu~Su.
\newblock Adaptive chameleon or stubborn sloth: Revealing the behavior of large
  language models in knowledge conflicts.
\newblock In \emph{International Conference on Learning Representations}, 2024.

\bibitem[Xu et~al.(2024)Xu, Qi, Guo, Wang, Wang, Zhang, and
  Xu]{xu2024conflictsurvey}
Rongwu Xu, Zehan Qi, Zhijiang Guo, Cunxiang Wang, Hongru Wang, Yue Zhang, and
  Wei Xu.
\newblock Knowledge conflicts for {LLM}s: A survey.
\newblock In \emph{Proceedings of the 2024 Conference on Empirical Methods in
  Natural Language Processing}, 2024.

\bibitem[Yang et~al.(2024)Yang, Sun, Xin, Sun, Bhalla, Chen,
  et~al.]{yang2024crag}
Xiao Yang, Kai Sun, Hao Xin, Yushi Sun, Nikita Bhalla, Xiangsen Chen, et~al.
\newblock {CRAG} -- comprehensive {RAG} benchmark.
\newblock In \emph{Advances in Neural Information Processing Systems (Datasets
  and Benchmarks Track)}, 2024.

\bibitem[Ye et~al.(2026)Ye, Chen, Zhong, Xiao, Zhang, Wu, and Shen]{ye2026tcr}
Hua Ye, Siyuan Chen, Ziqi Zhong, Canran Xiao, Haoliang Zhang, Yuhan Wu, and Fei
  Shen.
\newblock Seeing through the conflict: Transparent knowledge conflict handling
  in retrieval-augmented generation.
\newblock In \emph{Proceedings of the AAAI Conference on Artificial
  Intelligence}, volume~40, 2026.
\newblock URL \url{https://ojs.aaai.org/index.php/AAAI/article/view/40740}.

\bibitem[Yih et~al.(2016)Yih, Richardson, Meek, Chang, and Suh]{yih2016webqsp}
Wen-tau Yih, Matthew Richardson, Christopher Meek, Ming-Wei Chang, and Jina
  Suh.
\newblock The value of semantic parse labeling for knowledge base question
  answering.
\newblock In \emph{Proceedings of the 54th Annual Meeting of the Association
  for Computational Linguistics (Short Papers)}, 2016.

\bibitem[Yu et~al.(2023)Yu, Iter, Wang, Xu, Ju, Sanyal, Zhu, Zeng, and
  Jiang]{yu2023genread}
Wenhao Yu, Dan Iter, Shuohang Wang, Yichong Xu, Mingxuan Ju, Soumya Sanyal,
  Chenguang Zhu, Michael Zeng, and Meng Jiang.
\newblock Generate rather than retrieve: Large language models are strong
  context generators.
\newblock In \emph{International Conference on Learning Representations}, 2023.

\bibitem[Yuan et~al.(2026)Yuan, Kong, Li, Li, and Liu]{yuan2026confrag}
Yizhen Yuan, Rui Kong, Dongze Li, Yuanchun Li, and Yunxin Liu.
\newblock Benchmarking {LLM}'s capability in reasoning over conflicting web
  references.
\newblock In \emph{Proceedings of the 64th Annual Meeting of the Association
  for Computational Linguistics}, 2026.

\bibitem[Zhang et~al.(2025)Zhang, Xiang, Xiao, Wang, Li, Wang, and
  Su]{zhang2025faithfulrag}
Qinggang Zhang, Zhishang Xiang, Yilin Xiao, Le~Wang, Junhui Li, Xinrun Wang,
  and Jinsong Su.
\newblock {FaithfulRAG}: Fact-level conflict modeling for context-faithful
  retrieval-augmented generation.
\newblock In \emph{Proceedings of the 63rd Annual Meeting of the Association
  for Computational Linguistics (Volume 1: Long Papers)}, 2025.
\newblock URL \url{https://aclanthology.org/2025.acl-long.1062}.

\bibitem[Zhao et~al.(2025)Zhao, Yang, Hu, Tong, Lu, Wu, Gui, Zhang, and
  Huang]{zhao2025reconciliation}
Jun Zhao, Yongzhuo Yang, Xiang Hu, Jingqi Tong, Yi~Lu, Wei Wu, Tao Gui,
  Qi~Zhang, and Xuanjing Huang.
\newblock Understanding parametric and contextual knowledge reconciliation
  within large language models.
\newblock In \emph{Advances in Neural Information Processing Systems}, 2025.

\bibitem[Zheng et~al.(2025)Zheng, Guha, Arifov, Zhang, Skreta, Manning,
  Henderson, and Ho]{zheng2025legalretrieval}
Lucia Zheng, Neel Guha, Javokhir Arifov, Sarah Zhang, Michal Skreta,
  Christopher~D. Manning, Peter Henderson, and Daniel~E. Ho.
\newblock A reasoning-focused legal retrieval benchmark.
\newblock In \emph{Proceedings of the 4th ACM Symposium on Computer Science and
  Law}, 2025.
\newblock arXiv:2505.03970.

\bibitem[Zou et~al.(2025)Zou, Geng, Wang, and Jia]{zou2025poisonedrag}
Wei Zou, Runpeng Geng, Binghui Wang, and Jinyuan Jia.
\newblock {PoisonedRAG}: Knowledge corruption attacks to retrieval-augmented
  generation of large language models.
\newblock In \emph{34th USENIX Security Symposium}, 2025.

\end{thebibliography}
